\documentclass[poms,final,nonblindrev]{poms1_V1} 

\OneAndAHalfSpacedXI 

\usepackage{natbib}

\bibpunct[, ]{(}{)}{,}{a}{}{,}%
\def\bibfont{\small}%
\renewcommand{\bar}{\overline}

\newcommand{\D}{ \mathcal{{D}} }
\newcommand{\R}{ \mathbb{R} }
\renewcommand{\P}{ \mathbb{P}}

\newcommand{\E}{ \mathbb{E} }

\usepackage{parskip} 
\DeclareMathAlphabet\mathbfcal{OMS}{cmsy}{b}{n}

\def\eq#1{Eq.~(#1)}

\def\M{\mathcal M}

\newcommand{\EG}{\mathrm{EG}}

\usepackage{braket}
\usepackage{algorithm}
\usepackage{algorithmicx}
\usepackage{algpseudocode}
\def\1{(\mathrm{\uppercase\expandafter{\romannumeral1}})}
\def\2{(\mathrm{\uppercase\expandafter{\romannumeral2}})}
\def\O{\mathcal O}
\newcommand{\mh}{\mathcal{H}}

\usepackage[backref = false, bookmarks, colorlinks = true, plainpages = false, citecolor = blue , urlcolor = blue, filecolor = blue, linkcolor = blue]{hyperref}
\usepackage{booktabs}
\usepackage{enumitem}
\DeclareMathAlphabet{\mathscr}{OT1}{pzc}{m}{it}
\def\p{\widetilde{\bm{p}}}
\def\k{\kappa}
\def\O{\mathcal O}
\def\wo{\mathcal{\widetilde O}}

\usepackage{bm}

\TheoremsNumberedThrough     
\ECRepeatTheorems

\EquationsNumberedThrough    

\usepackage{xcolor}
\definecolor{DSgray}{cmyk}{0,1,0,0}

\renewcommand{\hat}{\widehat}
\renewcommand{\tilde}{\widetilde}
\usepackage[normalem]{ulem}
\begin{document}


\RUNAUTHOR{Chen, Lyu, Wang and Zhou}

\RUNTITLE{NRM with Demand Learning and Fair Resource-Consumption Balancing}

\TITLE{ Network Revenue Management with Demand Learning and Fair Resource-Consumption Balancing}


\ARTICLEAUTHORS{%
 \AUTHOR{Xi Chen\footnotemark[1]}
 \AFF{Leonard N.~Stern School of Business, New York University, New York, NY 10012, USA, \EMAIL{xc13@stern.nyu.edu}} 
 \AUTHOR{Jiameng Lyu\footnotemark[1]\footnotemark[2]}
 \AFF{Department of Mathematical Sciences, Tsinghua University, Beijing 100084, China, \EMAIL{lvjm21@mails.tsinghua.edu.cn}}
 \AUTHOR{Yining Wang\footnotemark[1]}
 \AFF{Naveen Jindal School of Management, University of Texas at Dallas, Richardson, TX 75080, USA, \EMAIL{yining.wang@utdallas.edu}}
 \AUTHOR{Yuan Zhou\footnotemark[1]\footnotemark[2]}
 \AFF{Yau Mathematical Sciences Center \& Department of Mathematical Sciences, Tsinghua University, Beijing 100084, China, \EMAIL{yuan-zhou@tsinghua.edu.cn}}
 }  

\renewcommand{\thefootnote}{\fnsymbol{footnote}}
\footnotetext[1]{Author names listed in alphabetical order.}
\footnotetext[2]{Corresponding authors.}
\renewcommand{\thefootnote}{\arabic{footnote}}

\ABSTRACT{
In addition to maximizing the total revenue,  decision-makers in lots of industries would like to guarantee  {balanced} consumption across different resources.  {For instance, in the retailing industry, ensuring a balanced consumption of resources from different suppliers enhances fairness and helps maintain a healthy channel relationship; in the cloud computing industry, resource-consumption balance helps increase customer satisfaction and reduce operational costs.} Motivated by these practical needs, this paper studies the price-based network revenue management (NRM) problem with both demand learning and  {fair resource-consumption balancing.} 
We introduce the regularized revenue, i.e., the total revenue with a {balancing regularization}, as our objective to incorporate fair resource-consumption balancing  into the revenue maximization goal. We propose a primal-dual-type online policy with the Upper-Confidence-Bound (UCB) demand learning method to maximize the regularized revenue. We adopt several innovative techniques to make our algorithm a unified and computationally efficient framework for the continuous price set and a wide class of {balancing regularizers}. Our algorithm achieves a worst-case regret of $\wo(N^{5/2}\sqrt{T})$, where $N$ denotes the number of products and $T$ denotes the number of time periods. Numerical experiments in a few NRM examples demonstrate the effectiveness of our algorithm {in simultaneously achieving revenue maximization and fair resource-consumption balancing.}}


\maketitle

\section{Introduction}\label{sec:intro}

Network revenue management, as a fundamental and important model in revenue management, has been successfully applied in lots of industries, such as online retailing, airline, hotel \citep{talluri2004theory,klein2020review}. Classical research on NRM aims to maximize the total revenue over $T$ time periods under resource constraints assuming the demand function is known. See, for example, the seminal works of \cite{gallego1997multiproduct,jasin2014reoptimization,maglaras2006dynamic}.

In practice, there are two main challenges in adopting NRM models. First, the demand function in NRM is usually unknown, which needs to be learned on the fly. Second, in addition to maximizing the total revenue, many {decision-makers} would like to guarantee {balanced} consumption across different resources.

In various network revenue management applications, resource-consumption balancing may achieve multiple benefits such as fairness in channel relationships, customer satisfaction enhancement, and operational cost reduction. Let us provide two examples to illustrate the importance of resource-consumption balancing as follows.

\textbf{Retailing Industry:} In an NRM model for the retailing industry, a retailer decides prices for products and each product consumes several types of resources. Different resources are provided by different suppliers, and it is also possible that one supplier could provide more than one type of resource. If one supplier terminates the cooperation, certain products may no longer be able to be produced.   Therefore, to maximize long-term revenue and guarantee no product is out of stock, the retailer should keep long-term good cooperation with different resource suppliers. 
Extensive research in the fields of economics and marketing has consistently revealed the significant role of fairness in the development and sustenance of channel relationships between suppliers and retailers \citep{corsten2005suppliers,haitao2007fairness}.
To this end, it is essential for retailers to ensure profit fairness across different suppliers. Indeed, only paying attention to revenue maximization may cause an ``unfair" scenario in which some suppliers earn a lot but the shares of others are small, subsequently leading to a higher probability of these under-used resource suppliers to withdraw from cooperation. 
Suppose the profit expectations of different suppliers are the same,\footnote{To address the issue of different profit expectations among suppliers while maintaining profit fairness, the retailer could introduce a weighting mechanism that adjusts resource-consumption balancing based on each supplier's profit expectations or earnings per demand (please refer to the weighting mechanism in Section~\ref{sec:regularizer}  for details).} resource-consumption balancing could enhance profit fairness among suppliers in the channel relationship, and consequently help maintain a good collaboration between retailers and suppliers. Moreover, over-reliance on certain suppliers creates a precarious dependency, since if some of these suppliers were to withdraw from the partnership, the retailer would face severe revenue loss and supply disruptions.
 In this way, resource-consumption balancing ensures that the retailer is not overly reliant on a few suppliers, reducing the risk of severe revenue loss and supply disruptions due to supplier withdrawal.

 \textbf{Cloud Computing Industry: } 
In an NRM model for the cloud computing industry,  the firm needs to make decisions on the prices of products (e.g., Infrastructure as a Service (IaaS), Platform as a Service (PaaS), Software as a Service (SaaS), or specialized services like data storage, machine learning, or database management) subject to resource constraints (e.g., the capacity constraints of servers, virtual machines, data centers, storage systems, network bandwidth,  and other computing resources).  Designing an algorithm solely focused on profit maximization over a specific time period, without considering the balanced consumption of different resources, can have several potential consequences.
From the perspective of customer satisfaction, overutilization of certain resources can lead to saturation and bottlenecks, resulting in poor service quality, slower response times, inconsistent performance, and limited flexibility for customers. Even if the firm achieves short-term maximum profit, dissatisfied customers may seek alternative providers or express their dissatisfaction, leading to customer churn and potential long-term revenue loss.
From the perspective of operational cost and efficiency, neglecting resource consumption balance can lead to some resources being heavily utilized and quickly depleted, while others remain underutilized. Overloaded resources may require more frequent maintenance or replacement due to excessive usage, resulting in higher operational costs. Additionally, managing and maintaining imbalanced resource consumption simultaneously becomes more challenging, reducing overall operational efficiency. 
It seems that optimizing the capacity of different resources can be another way to solve the above issues. However, frequently scaling up or down to optimize capacities to match rapid shifts in demand and resource requirements might not be feasible and practical, since adjusting capacities in could computing industries can be a time-consuming and costly process and might lead to service disruptions. In contrast,
resource-consumption balancing could increase customer satisfaction and reduce operational costs,  which in turn contributes to long-term profitability and business growth, in a more convenient way.

As shown in the above examples, the resource-consumption balancing objective is often measured in the sense of the whole selling season and cannot be decomposed as an additive objective at each time period.
In this paper, we adopt several different metrics that are applied to the average resource consumption vector to measure its balancing level. For example, the balancing level on resource consumption could be measured by the minimum element of the average resource consumption vector, which is the famous max-min fairness metric extensively studied in economics and resource allocation literature, also included as a special case of the weighted max-min fairness metric studied in this paper. In some cases, it may also be desirable to consume the resources at similar rates, which can be incentivized by our range fairness metric. In addition to the above two examples, we propose several more practically useful balancing metrics. We further identify a wide class of balancing metrics (including all the above examples) with quite mild assumptions. All the balancing metrics in the class can be incorporated into our learning and doing framework.  Please refer to Section~\ref{sec:regularizer} for more details.

The main contribution of this paper is a dynamic pricing algorithm that simultaneously learns the unknown demand function and optimizes the composite objective concerning both the NRM revenue and the balancing metric. For any balancing metric included in the class mentioned above, our algorithm achieves a regret at most $\wo(N^{5/2} \sqrt{T})$ (where $N$ is the number of products and $T$ is the selling horizon, see Theorem~\ref{thm:main} for more details). 
Below we discuss the various technical challenges and our contributions that overcome them in detail.

\subsection{Main technical challenges}

There are several technical challenges to tackle for the NRM problem with resource-consumption balancing and demand learning. First, the balancing metrics are all applied to the average resource consumption vector, which is a global objective calculated across all time periods. There is also a global unreplenishable inventory constraint for the resources. The \emph{global objective and constraint} are usually difficult for sequential decision making problems since the decisions at each time step have to be well coordinated to jointly optimize the balancing metric and satisfy the inventory constraint. Second, the problem becomes even harder when the demand function is unknown to the retailer and the retailer has to balance the \emph{exploration vs.~exploitation} trade-off with only learned information or estimated demand. Here, ``exploration'' means that the retailer needs to explore different prices in order to learn the unknown demand function on the run, and ``exploitation'' means that the retailer needs to exploit the near-optimal price to simultaneously gain revenue and achieve the global balancing objective. Finally, we aim at designing a \emph{computationally efficient} algorithm (i.e., that runs in polynomial time) to minimize the regret (i.e., the cost of learning and sequential decision making) about the revenue and the balancing metric.

Compared with the work of \cite{balseiro2020regularized}, the fact that demand curves are unknown in our problem leads to several specific technical hurdles
that require novel solutions. In particular, we have the following challenges:
\begin{enumerate}
\item In \cite{balseiro2020regularized}, the dual problem is unconstrained, leading to complex dual space shapes and unbounded penalty vectors.
This causes two problems: first, with dual spaces having complex shapes, the dual update steps may not have closed forms or be solved efficiently for 
general balancing-induced penalty functions.
More importantly, with penalty vectors being unbounded the regret of the problem becomes unbounded as well because of the uncertainty exhibited
in the estimated demand functions, rendering bandit learning algorithms impractical.
\item With uncertainty quantified in the estimated demand functions, solving the primal update steps becomes computationally intractable as the objective functions
are not necessarily concave anymore. Novel algorithms and analysis are required to solve such non-concave problems efficiently and rigorously.
\end{enumerate}

\subsection{Our contributions}

We are able to address the above challenges with several technical innovations. At a higher level, our algorithm combines the primal-dual-type online policy proposed by \cite{balseiro2020regularized} with the Upper-Confidence-Bound (UCB) method, where the former solves the   regularized online allocation problem with the \emph{known} demand function and the latter is a widely adopted principle to balance the exploration and exploitation trade-off in many online learning algorithms. However, this combination is not black-box style and we make novel technical contributions in the design and analysis of our algorithm. Moreover, we also introduce new algorithmic ingredients to make sure that our algorithm is computationally efficient. In contrast, most of the online learning algorithms for linear demand models in the revenue management literature do not guarantee a polynomial time complexity, especially in the multi-product case. Below, we describe our main technical contributions in more detail.

First, instead of using the UCB of the objective function in the usual online learning and decision making algorithms, we make decisions according to the UCB of the specially designed \emph{adjusted revenue function}. While our adjusted revenue function involves the revenue, it does not directly include the balancing metric (as it is not obvious how to decompose such a global metric to each individual time period, as discussed previously). Instead, we include a carefully designed term in the adjusted revenue function to relate it to the dual variable. This term, together with our update rule of the dual variable, helps optimize the balancing metric in a global fashion and reflects the inventory constraint at the same time.

Second, to control the estimation error of the adjusted reward during learning, we need to design a bounded domain for the dual parameter (i.e., the dual space). In contrast, \cite{balseiro2020regularized} adopt an unbounded dual space which is unfriendly to the analysis of our learning process, 
{as illustrated in the first bullet point in the previous section.
By adopting a bounded domain for the dual parameter, we are able to upper bound the estimation errors as well, leading to a correct regret rate.}
An additional benefit of our new dual space is that due to its simpler shape, we are able to employ a closed-form dual update rule (Algorithm~\ref{alg:eg}, Section~\ref{sec:mirrordescent}) for \emph{any} balancing metric. In contrast, \cite{balseiro2020regularized} may only achieve this for selected balancing metrics and the dual update in our algorithm is much simplified.

Third, the regret analysis (especially the analysis related to the dual variable) greatly relies on the magnitude of our model parameter estimations. While the natural (regularized) least-squares estimator may not provide the desired bound, we employ an additional convex program $\mathcal{M}_t$ to compute a set of bounded model estimates. This $\mathcal{M}_t$ program is also helpful to guarantee the concavity of the estimated (adjusted) revenue function so that we may computationally efficiently find its maximum point which crucially connects to the decision we will make at each time step.

Finally, to achieve computational efficiency, we adopt an $\ell_\infty$-norm confidence radius instead of the usual $\ell_2$-norm confidence radius when computing the Upper Confidence Bounds so that we are able to maximize the UCB of the adjusted revenue function in polynomial time. Also, by reducing the $\mathcal{M}_t$ program to a linear program with infinitely many constraints, we design a polynomial-time separation oracle and invoke the Ellipsoid method to efficiently solve the $\mathcal{M}_t$ program. Both ingredients help our algorithm to achieve the polynomial time complexity that addresses the second bullet point of technical challenges
mentioned in the previous section.

For the first 3 technical contributions, we provide more concrete explanations at the end of Section~\ref{sec:alg}, after the introduction of notations and the algorithm description. For the last item, please refer to Section~\ref{sec:find} and Section~\ref{sec:optim} for more details.

\subsection{Related Works}\label{sec:relatework}

In this section, we introduce three streams of literature related to our paper:
network revenue management (NRM) with a known demand function, revenue management (RM) with demand learning, and fairness and resource-consumption balancing in operations management.
And we discuss how our paper is appropriately placed into contemporary literature by giving comparisons with closely-related existing works.

\textbf{NRM with known demand function.}
A large body of the price-based network revenue management literature focuses on the case in which the seller knows the underlying demand function in advance. And it is known that the optimal pricing policy of this case can be computed using dynamic programming (DP). However, the well-known curse of dimensionality of DP makes the optimal pricing policy computationally intractable. As a result, many works in the literature have investigated developing algorithms that are computationally efficient with a superior revenue performance. The seminal work by  \cite{gallego1994optimal,gallego1997multiproduct} proposed simple but powerful heuristics. Specifically, they solve the optimal price of the fluid approximation model which is a deterministic analog of the DP and choose a static price every time. And their approach achieves an $\O(\sqrt{T})$ regret. \cite{jasin2014reoptimization} introduced an improvement to the static pricing policy by resolving the static price periodically according to the remaining inventory, and attained $\O(\log T)$ regret bound. Recently, \cite{wang2022constant} proved that the resolving heuristics can achieve $\O(1)$ regret as compared to the optimal policy of the DP.

\textbf{RM with demand learning.}
There is a large body of literature focusing on the price-based revenue management with demand learning, which are either without inventory constraints (see, e.g., \cite{den2014dynamic,den2014simultaneously,keskin2014dynamic,keskin2017chasing,bu2022online} and references therein) or with inventory constraints (see, e.g., \cite{besbes2009dynamic,wang2014close,chen2014adaptive,ferreira2018online,miao2021general}). 
For dynamic pricing problems without inventory constraints, we refer the readers to \cite{den2015dynamic} for a detailed review.
For price-based revenue management problems with inventory constraints, there are two streams of literature, either considering the nonparametric demand model \citep{wang2014close,chen2019network,chen2019nonparametric,miao2021network} or the parametric demand model (see discussion below).
Since our paper considers a parametric demand function, we mainly investigate the literature on the revenue management problem with inventory constraints and the parametric model. There are three main approaches for tackling the learning-while-doing challenge.

The first approach is using the Explore-Then-Commit strategy, which separates the exploration phase and exploitation phases. This simple strategy has been widely used in online learning tasks, and \cite{besbes2009dynamic,besbes2012blind} and \cite{chen2014adaptive} applied this strategy to the NRM problem and \cite{chen2014adaptive} achieved $\O(\sqrt{T})$ regret assuming the strong concavity of the revenue function.

The second approach is using Thompson sampling to address the exploration-exploitation trade-off. 
\cite{ferreira2018online} introduced Thompson sampling into network revenue management and considered both the discrete price model and continuous price set with the linear demand model. They obtained a Bayesian regret $\wo(\sqrt{T})$ instead of the worst-case regret. The most important step in their algorithm for the continuous price set is to solve a quadratic program, which is not guaranteed to be a convex problem and is not clear how to be solved efficiently.

The third approach is incorporating the Optimism in the Face of Uncertain principle into the primal-dual optimization framework. 
This approach is closely related to the Bandit-with-Knapsack (BwK) model \citep{badanidiyuru2013bandits}, which introduces global resource constraints into the multi-armed bandit. \cite{agrawal2019bandits} further generalized BwK to bandit with global convex constraints and concave objective. The work by  \cite{agrawal2016linear}, which considered BwK in the linear bandit setting, can be applied to the NRM problem with the discrete price. However, in the continuous price setting, the regret and the running time will be exponentially dependent on the number of products due to the discretization procedure. \cite{miao2021general} considered the NRM problem with continuous price and generalized linear model. To tackle the high computational complexity due to the continuous price set, they designed a UCB solver to reduce the original optimization problem to the price optimization problem of an ordinary NRM problem by randomly sampling a vector on the unit sphere and using it to linearize the $\ell_2$-norm-based UCB term. However, the price optimization problem might still be non-convex and difficult to solve despite this reduction.

 Among the above three types of approaches, 
our work falls into the primal-dual category, which seems to be most proper to address (via the Lagrangian) both the global regularization and resource constraints in our problem. Our work is closely related to \cite{agrawal2016linear} and \cite{miao2021general}. However, there are several significant differences.
First, the primal-dual framework in \cite{agrawal2016linear} and \cite{miao2021general} do not consider the global regularization and their algorithms do not directly work in our setting.  To simplify the global regularization, we introduce an auxiliary variable. With the regularization and the auxiliary variable, our algorithm adopts very different primal and dual updates, which requires a different analysis. For example, on the primal side, we need to update the auxiliary variable according to a maximization problem of the regularizer in each period and carefully handle the term of the auxiliary variable in the regret analysis.  On the dual side, the update of the dual variable is also related to the auxiliary variable,  which makes the choice of the stationary benchmark dual variables more involved. Second, as compared to the random sampling method in \cite{miao2021general} which sacrifices an $\O({N})$ factor as shown in Lemma EC.2.1 of their paper, we introduce the $\ell_\infty$-norm-based UCB term (Section~\ref{sec:optim}) which is not only simpler to calculate, but also only sacrifices an $\O(\sqrt{N})$ factor in the regret. In contrast, even without the   {balancing regularization}, the regret of \cite{miao2021general} for the NRM problem is $\wo(N^{3.5}\sqrt{T})$, about $N$ times our regret (as discussed above there is $\O(\sqrt{N})$ term due to the random sampling method; the another $\O(\sqrt{N})$ term due to their suboptimal self-normalized concentration inequality lemma). Third, we introduce a feasibility program $\mathcal{M}_t$ (Section~\ref{sec:find}) to make sure the estimated revenue function is concave and computationally easy to optimize. We are able to combine the above new techniques to derive a computationally efficient low-regret learning-while-doing algorithm for the NRM problem with fair resource-consumption balancing.

\textbf{Fairness and resource-consumption balancing in operations.}
With the development of data-driven algorithms in operations, there is a growing concern about discrimination and unfairness.
As a result, the fairness issue has attracted a lot of attention in operations problem \citep{bonald2006queueing,ma2020group,Kallus:21,Kallus:21:fair,Zhang:22:routing,cohen2021dynamic,cohen2022price,chen2021fairness}


There is a vast body of literature considering the fairness in online allocation problem with inventory constraints (with \emph{known} demand models)  \citep{elzayn2019fair,ma2020group,balseiro2020regularized,chen2021fairer}, where the decision-maker must take an action upon each arriving request and generates a reward and the consumption of resources.  { \cite{balseiro2020regularized} and \cite{ma2022optimal}  considered revenue maximization and fair resource-consumption (or resource-allocation) balancing simultaneously in the online allocation problem by introducing a non-separable regularizer. \cite{balseiro2020regularized} emphasized the importance of resource-consumption balancing in online advertising \citep{miller2015algorithms} and cloud computing industries \citep{al2012survey}  to avoid saturation of certain resources and retain some free capacity to maintain flexibility.  \cite{pmlr-v202-zhang23at} 
studied the online allocation problem with two-side resource constraints, which can be seen as a novel way to achieve resource-consumption balancing. They also provided several real-life examples, such as the online orders assignment in e-commerce platforms and online advertising platforms \citep{zhang2020request}, where it is required to guarantee a certain amount of resource-consumption.}
Since our work is most related to \citep{balseiro2020regularized}, we have thoroughly discussed the technical differences in the introduction section above and we will present the comparison more concretely in Section~\ref{sec:alg}.
We also note that \citep{balseiro2020regularized} works for the regularized NRM problem in a \emph{quantity-based} setting, where the decision-maker must irrevocably accept or reject each arriving request given limited resources (a special case of the online allocation problem studied in their paper). In contrast, we study the NRM problem in the \emph{price-based} setting where the decision-maker has to decide the prices that influence the demand and the demand has always to be met (as long as permitted by the resource constraints). To the best of our knowledge, our work is the first to consider the fairness objective in the price-based NRM problem.


\subsection{Notations}
The vectors throughout this paper are all column vectors.
We denote the set $\{1,2,\dots,N\}$ by $[N]$ for any $N\in\mathbb N$.
For vectors $\bm a, \bm b \in \R^N$, we use $\bm a\leq \bm b$ ($\bm a \geq \bm b$ respectively) to denote $a_i\leq b_i$ ($a_i \geq b_i$ respectively) for all $i\in[N]$. 
We use $[\bm x_t]_i$ to denote  the $i$-th element of the vector $\bm x_t$.  
For $\bm x\in\R^N$ and $\bm{\Lambda}\in\R^{N\times N}$,  we define the following norms: $\|\bm x\|_1:=\sum_{i=1}^N|x_i|$,  $\|\bm x\|_2:=(\sum_{i=1}^Nx_i^2)^{1/2}$, $\|\bm x\|_\infty:=\max_{i\in[N]}x_i$, and $\|\bm x\|_{\bm{\Lambda}}:=\sqrt{\bm x^\top\bm{\Lambda} \bm x}$. 

We use $\mathbf{I}_{N}$ to denote the identity matrix of order $N$. For $\bm{A}\in\R^{M\times N}$, we define the following matrix norms: $\|\bm{A}\|_F:=\sum_{i,j}A_{ij}^2$ ($A_{ij}$ refers to the element in the $i$-th row and $j$-th column of matrix $\bm{A}$), $\|\bm{A}\|_2:=\mathrm{sup}_{\|\bm x\|_2=1}\|\bm{A}\bm x\|_2$,  $\|\bm{A}\|_\infty:=\mathrm{sup}_{\|\bm x\|_\infty=1}\|\bm{A}\bm x\|_\infty$ and it is easy to obtain $\|\bm{A}\|_\infty=\max_{i\in[M]}\sum_{j=1}^N|A_{ij}|$. For square matrix  $\bm{A} \in \mathbb{R}^{N \times N}$, we use $\lambda_{\mathrm{max}}(\bm{A})$  to denote the largest eigenvalue of $\bm{A}$. For a symmetric matrix $\bm{A} \in \mathbb{R}^{N \times N}$, we use $\bm{A}\preceq 0$  to represent that $\bm{A}$ is negative semi-definite. We use $[\bm B|\bm{\alpha}]$ to denote the augmented matrix by adding $\bm{\alpha}\in\R^{N}$ to the right of the matrix $\bm B\in\R^{N\times N}$ as a new column. 

We use the big-$\O$ notation $f(T)=\O(g(T))$  to denote that $\limsup_{T \to \infty}f(T)/g(T)\leq +\infty$.  We use $\wo(\cdot)$ to further omit the logarithmic dependency on $N$, $M$, and $T$.

\subsection{Organization}

The remainder of this paper is organized as follows. In Section~\ref{sec:model}, we formulate our problem by introducing the model assumptions and the performance measure; we also give plenty of examples of the balancing regularizers to illustrate the potential guidance our paper might bring to the practical scenarios. In Section~\ref{sec:alg}, we present our algorithm and discuss the high-level ideas of the algorithm design. 
Then we discuss reward and demand estimation (Section~\ref{sec:ucb}) and the design of the mirror descent solver (Section~\ref{sec:mirrordescent}) in detail, which are two key building blocks of our algorithm. In Section~\ref{sec:mainresults}, we present the main theorem that upper bounds the regret of our algorithm (the detailed proof of the main theorem is presented in Section~\ref{sec:regretanalysis} in E-Companion). To demonstrate the empirical performance of our policy, we conduct several numerical experiments and present the results in Section~\ref{sec:experiments}. In the end, we give a summary of our paper in Section~\ref{sec:conclusion}. The proofs of most technical lemmas and the additional experimental results are included in the supplementary materials.

\section{Model Description and Assumptions}\label{sec:model}

In an NRM model with $N$ types of products and $M$ types of resources, a retailer sells $N$ types of products during a selling season with $T$ time periods. Each product is defined as a combination of $M$ types of unreplenishable resources by the consumption matrix $\bm A\in\R^{M\times N}$, where $A_{ij}$ means that selling one unit of the type-$i$ product consumes $A_{ij}$ unit of the type-$j$ resource. At each time period $t$, the retailer must determine the prices for the $N$ products, i.e., the price vector $\bm{p}_t\in\R^N_+$. The retailer then observes the consumer's demand vector $\bm{d}_t\in\R^N_+$ which is realized from an unknown underlying demand function $\bm D(\bm{p}_t)$, and finally consumes the resources according to the consumption matrix $\bm{A}$. The retailer needs to choose the prices during the selling season to accomplish the following 3 goals:
\begin{enumerate}
\item to gradually learn the underline demand function $\bm D(\bm{p}_t)$ from the observed demands,
\item to maximize the total revenue based on the learned information and given the unreplenishable resource inventory, 
\item to balance the consumption of the different types of resources via maximizing the balancing regularizer $\phi(\cdot)$, which will be defined soon.
\end{enumerate}

More specifically, the initial inventory levels of the $M$ resources are $\bm{I}_0=(I_{0,1},\dots,I_{0,M})^\top\in \mathbb R^M_+$. At the end of time $t$, the inventory levels become $\bm{I}_t=\bm{I}_{t-1}-\bm{A}\bm{d}_t$ for $t = 1, 2, 3, \dots$. For convenience, we also define the normalized inventory level $\bm{\gamma} =(\bm{\gamma}_{1},\dots,\bm{\gamma}_{M})^\top  :=  \bm{I}_0/T$, which is the average amount of resources that can be used at a time period.

For simplicity, we assume that the price range for each product is $[\underline{p},\bar p]$  and the retailer has to choose $\bm{p}_t$ in the price set $\mathcal P$  at each time $t$. The price set $\mathcal P$ can either be $[\underline{p},\bar p]^N$ or a discrete subset in $[\underline{p},\bar p]^N$.
For brevity, we focus on the case $\mathcal P = [\underline{p},\bar p]^N$, which is much more challenging. One can easily adapt our algorithm and analysis to the discrete price set.

The realized demand $\bm{d}_t$ is a random variable centered at $\bm D(\bm{p}_t)$, i.e., 
\[
\bm{d}_t = \bm D(\bm{p}_t)+\bm{\varepsilon}_t
\]
where $\varepsilon_t$ is a zero-mean noise variable (see Assumption~\ref{assump:demand} for the more precise statement). We consider the linear demand function (which is the most commonly analyzed demand model in literature, e.g., \cite{keskin2014dynamic} and \cite{ferreira2018online})
\[
\bm D(\bm{p}_t)=\bm{\alpha}+\bm{B}\bm{p}_t,
\]
where $\bm{\alpha}\in\R^N$ and $\bm{B}\in\R^{N\times N}$ are the model parameters unknown to the retailer. For convenience, we also denote these unknown parameters by $\bm{\theta}=(\bm{\alpha}, \bm{B})\in\Theta \subseteq \R^{N^2+N}$, where $\Theta$ is the parameter space.

The revenue collected by the retailer at time $t$ is $r_t=\Braket{\bm{d}_t,\bm{p}_t}$. We also denote the corresponding expected revenue by 
\[
r(\bm{p}_t):= \E[r_t|\bm{p}_t]=\Braket{\bm{p}_t,\bm D(\bm{p}_t)}.
\]

The objective of the retailer is to design a policy  $\pi=(\pi_1,\dots,\pi_T)$ with $\pi_t:\mh_t\mapsto \bm{p}_t$ (where $\mh_t=\{\bm{p}_s,\bm{d}_s\}_{s<t}$ is the historical prices and demands before time $t$) to satisfy the inventory constraint $\bm{I}_t \geq 0$ for all $t \in \{1, 2, \dots, T\}$ and maximize the following expected total revenue plus the balancing  regularizer on resource consumption:
\begin{equation}\nonumber
\E\left[\sum_{t=1}^Tr(\bm{p}_t)+T\phi\left(\frac{1}{T}\sum_{t=1}^T\bm{A}\bm{d}_t\right)\right] .
\end{equation}
Below in Section~\ref{sec:regularizer} we will discuss more about the 
 regularizer $\phi(\cdot)$; in Section~\ref{sec:model-assump-performence-measure} we introduce some standard assumptions on the linear demand model and define the regret that our online policy aims to minimize.

\subsection{Balancing Regularizer $\phi(\cdot)$: Assumptions and Examples}\label{sec:regularizer}

The resource-consumption balance is measured by the regularizer  $\phi\left(\frac{1}{T}\sum_{t=1}^T\bm{A}\bm{d}_t\right)$ (i.e.,  the regularizer function $\phi(\cdot)$ applied to the average resource consumption vector). 

\begin{assumption}\label{assump:phi1}
Throughout this paper, we impose the following assumptions on  $\phi(\cdot)$.
\begin{enumerate}
\item $\phi(\bm{s})$ is $L$-Lipschitz continuous with respect to the $\|\cdot\|_{\infty}$-norm on its effective domain, i.e., $|\phi(\bm s_1) - \phi(\bm s_2)| \le L \| \bm s_1 - \bm s_2\|_{\infty}$ for any $\bm s_1,\bm s_2 \le \bm{\gamma}$.
\item There exists $\bar \phi$ such that $   0\leq\phi(\bm{s})\le \bar \phi$ for all $\bm{0}\leq \bm s\leq \bm{\gamma}$.
\item $\phi(\bm{s})$ is concave.
\end{enumerate}
\end{assumption}

In the following, we present several regularizers satisfying the above assumptions as examples (please refer to Section~\ref{sec:assumptions validations} in E-Companion for the detailed proof). We will use $\bm{s}=\frac{1}{T}\sum_{t=1}^T\bm{A}\bm{d}_t$ and $s_i$ refers to the average consumption of the type-$i$ resource. 

\noindent \textbf{Example 1: Weighted Max-min Fairness Regularizer.} The first example is rooted in the famous max-min fairness guarantee \citep{nash1950bargaining}, which has been well studied in the literature on static resources allocation \citep{bansal2006santa,bertsimas2011price}. The idea behind the max-min fairness guarantee is to promote fairness by maximizing the minimum resource allocation. In our paper, we consider the following weighted max-min fairness regularizer to promote fair resource-consumption balancing. It is worth noting that the max-min regularizer in \cite{balseiro2020regularized} can be seen as a special case of our weighted max-min regularizer by setting the parameters correspondingly.

Formally, we define the weighted max-min fairness regularizer as $\phi(\bm{s}):=\lambda\min_i(w_is_i)$, where $\lambda$ is the parameter to balance between the total revenue goal and the balancing objective,  and in the online retailing setting the parameter $w_i$ could be selected as the revenue of the resource supplier due to the consumption of one unit type $i$ resource and {$w_i$ could also denote different profit expectations among suppliers.}

\noindent \textbf{Example 2: Group Max-min Fairness Regularizer.}
We may divide the different types of resources into groups and only focus on promoting the minimum consumption of each resource group. In practice, each supplier may provide several types of resources (which naturally forms a group) and the group max-min fairness would be useful if we wish to guarantee fairness among the suppliers.

Formally, we define the group max-min fairness regularizer as  $\phi(\bm{s}):=\lambda\min_i( (\bm{U}\tilde{\bm{s}})_i)$,\footnote{In general, we may combine the grouping operation with any balancing  regularizer satisfying the Assumption~\ref{assump:phi1} to obtain a group version of the balancing  regularizer, but for simplicity, we only present the group version of the weighted max-min regularizer here.} where $\tilde{\bm{s}}=(w_1s_1,\cdots,w_ms_m)^\top$, $w_i$ is similarly defined as in Example 1, and 
$\bm{U}\in\R^{K\times M}$ is a $0$-$1$ matrix describing the grouping scheme. In particular, we require that in each column there is exactly $1$ non-zero element and in each row there is at least $1$ non-zero element, where a simple example of $\bm{U}$ is as follows,
\setlength{\arraycolsep}{8pt}
\renewcommand{\arraystretch}{1.5}
\[
    \bm{U} =
  \begin{bmatrix}
  1 & 0 & 1 & 0\\
     0 & 1 & 0& 1
  \end{bmatrix} .
\]
In this example, there are two resource suppliers.  The first supplier provides the type-$1$ and the type-$3$ resources and the second supplier provides the type-$2$ and the type-$4$ resources.

\noindent \textbf{Example 3: Range Fairness Regularizer.} Range is a fundamental statistical quantity that measures the difference between the highest and the lowest value of a population. The range fairness regularizer provides an incentive to minimize the range among the entries of the weighted average consumption vector $\tilde{\bm{s}}=(w_1s_1,\cdots,w_ms_m)^\top$.  

Formally, we define the range fairness regularizer as $\phi(\bm{s}):=\lambda(\min_i(w_is_i)-\max_i(w_is_i)+\max_i(w_i\gamma_i))$, where $- [\min_i(w_is_i)-\max_i(w_is_i)]$ is the range of $\tilde{\bm{s}}$ and $\max_i(w_i\gamma_i)$ is introduced to guarantee the positiveness of the regularizer. When $w_i$ is chosen to be per-unit revenue of the type-$i$ resource supplier, the range fairness regularizer can be applied to promote revenue fairness across different suppliers; when $w_i=1/\gamma_i$, this regularizer can evaluate the evenness of resource availability and help to avoid the pre-mature saturation of a few resource types.

\noindent \textbf{Example 4: Load Balancing Regularizer.} We finally present the load-balancing regularizer proposed in \cite{balseiro2020regularized}. The regularizer is defined as $\phi(\bm{s}):=\lambda(\min_i((\gamma_i-s_i)/\gamma_i)$, which measures the minimum relative resource availability, and also helps to make sure that no resource is too demanded.

\subsection{Model Assumptions and Performance Measure}\label{sec:model-assump-performence-measure}

\begin{assumption}\label{assump:demand}
Throughout this paper, we impose the following assumptions on the demand model $\bm{d}_t = \bm{\alpha}+\bm{B} \bm{p}_t+\bm{\varepsilon}_t$:
\begin{enumerate}
\item The noise $\{\bm{\varepsilon}_t\}_{t=1}^T$ is a martingale difference sequence adapted to the filtration $\{\mathcal F_t\}_{t=1}^T$ where $\mathcal F_t=\{\bm{p}_1,\bm{d}_1,\cdots,\bm{p}_t,\bm{d}_t,\bm{p}_{t+1}\}$, i.e., $\E[\bm{\varepsilon}_t|\mathcal F_{t-1}]= \bm{0}$. 
\item There exists $\bar d$ such that every entry of $\bm{d}_t$ is at most $\bar d$ almost surely for all $t \in \{1, 2, \dots, T\}$.
\item The underlying true parameter $\bm B$ in the linear demand model is negative definite; \footnote{$\bm{B}$ is is negative definite (not necessarily symmetric) if for any $\bm{z}\in\R^n$, it holds that $\bm{z}^\top \bm{B} \bm{z}<0$.} there exists $L_B$ (with $L_B\geq 1$) such that $\sqrt{\alpha_i^2+\| \bm B^\top \bm{e}_i \|^2_2} \leq L_B$ for every $i\in[N]$, where $\bm{e}_i$ is the $i$-th unit vector and $\bm{B}^\top \bm{e}_i$ is the $i$-th row of $\bm{B}$.
\end{enumerate}
\end{assumption}

All items in Assumption~\ref{assump:demand} are quite standard in the literature. The third item is usually seen in papers focusing on the linear demand model (see, e.g., \cite{keskin2014dynamic,ferreira2018online,bu2022online}).  By the definition $r(\bm{p}_t)=\Braket{\bm{d}_t,\bm p_t}$ and according to Assumption~\ref{assump:demand}, we may upper bound $r(\bm{p}_t)$ by $\bar r := N\bar p\bar d$. 

We now discuss the performance measure for the retailer's policy. We would like to compare the objective value achieved by the retailer with the optimal \emph{offline} policy, i.e., the one that knows all the model parameters $\theta$:
\begin{align}\label{eq:dp}
J_{\mathrm{opt}} := \max_{\pi = (\pi_1,\dots,\pi_T)} \E\left[\sum_{t=1}^Tr(\bm{p}_t)+T\phi\left(\frac{1}{T}\sum_{t=1}^T\bm{A}\bm{d}_t\right)\right] \qquad \text{s.t.~} \sum_{t=1}^T \bm{A}\bm{d}_t \leq T\bm{\gamma}~~a.s. 
\end{align}

$J_{\mathrm{opt}}$ upper bounds the objective value achieved by any \emph{online} policy (i.e., the one without access to $\theta$). In light of this, we define the regret of a policy $\pi$ up to time horizon $T$ as
\begin{align}\label{regretform1}
\mathcal{R}(T) &:=J_{\mathrm{opt}}- \E\left[\sum_{t=1}^Tr(\bm{p}_t)+T\phi\left(\frac{1}{T}\sum_{t=1}^T\bm{A}\bm{d}_t\right)\right].
\end{align}
 
Note that solving the exact value of $J_{\mathrm{opt}}$ is quite complicated due to the stochastic nature and adaptivity available to choose $\pi_1, \pi_2, \dots, \pi_T$ in sequence. We now introduce the following fluid model which is the deterministic and non-adaptive analog of $J_{\mathrm{opt}}$ and is easier to analyze.
\begin{align}\label{eq:fluid}
J_{\mathrm D} := \max_{\bm p_1,\dots,\bm p_T\in[\underline{p},\overline{p}]^N}  \left[\sum_{t=1}^T r(\bm{p}_t)+T\phi\left(\frac{1}{T}\sum_{t=1}^T \bm{A}\bm D(\bm{p}_t)\right)\right] \qquad \text{s.t.~} \sum_{t=1}^T \bm{A} \bm D(\bm{p}_t) \leq T\bm{\gamma}.
\end{align}
We assert that there exists an optimal solution $\{p^*_1, p_2^*, \dots, p^*_T\}$ to $J_{\mathrm D}$ such that $p^*_1 = p_2^* = \dots = p^*_T = p^*$, since otherwise, we can set $p_t = p' = \frac{1}{T}\sum_{t=1}^{T}p^*_t$ for every $t$, and the objective value of $\{p_1', p_2', \dots, p_T'\}$ becomes no smaller due to the concavity of $r(\bm{p})$ (since $r(\bm{p}) = \langle \bm{p}, \bm{\alpha} + Bp \rangle$ and $B$ is negative definite by Assumption~\ref{assump:demand}) and Jenson's inequality. Therefore we have the following equivalent definition of $J_{\mathrm D}$.
\begin{align}\label{eq:fluid3}
J_{\mathrm D} = \max\limits_{\bm{p}\in[\underline{p},\overline{p}]^N}\quad \left[Tr(\bm{p})+T\phi\left(\bm{A}\bm{\bm{D}}(\bm{p})\right) \right]
     \qquad \text{s.t.~~~~}  \bm{A}\bm{\bm{D}}(\bm{p}) \leq  \bm{\gamma} .
\end{align}
We denote $\bm p^*$ by the optimal solution to Eq.~\eqref{eq:fluid3}.

The following proposition shows that the fluid model $J_{\mathrm D}$ is an upper bound of $J_{\mathrm{opt}}$, and its proof is deferred to the supplementary materials.
\begin{proposition}\label{prop1}
$\displaystyle{J_{\mathrm{opt}}\leq J_{\mathrm D}}$.
\end{proposition}

By Proposition~\ref{prop1}, we upper bound the regret of any policy $\pi$ as follows, which will serve as the starting point of the analysis of our proposed policy.
\begin{align}
\mathcal{R}(T) \label{regretform2}
\leq T[r(\bm{p}^*)+\phi(\bm{A}\bm{\bm{D}}(\bm{p}^*))]-\E\left[\sum_{t=1}^Tr(\bm{p}_t)+T\phi\left(\frac{1}{T}\sum_{t=1}^T\bm{A}\bm{d}_t\right)\right].
\end{align}

\begin{algorithm}[ht]
\caption{Primal-dual + UCB for NRM with Demand Learning and Fair Resource-Consumption Balancing}
\label{algucb}
\begin{algorithmic}[1]
\State  \label{line:alg-main-initialization} Initialize the dual variable $\bm{\mu}_1=(0,\cdots,0)\in\R^M$.
\For{$t = 1,2,\dots,T$}
    \State \label{line:alg-main-RLS} Compute the regularized least-squares estimator 
\begin{equation}\label{ridge}
         (\hat{\bm{\alpha}}_t,\hat{\bm{B}}_t)\in \argmin_{(\bm{\alpha},\bm{B})} \left\{(N+1)(\|\bm{\alpha}\|_2^2+\|\bm{B}\|_F^2) + \sum_{s<t} \|\bm{d}_s-(\bm{\alpha}+\bm{B}\bm{p}_s)\|_2^2\right\}.
\end{equation}

    \State \label{line:alg-main-check-B} Find $(\check{\bm{\alpha}}_t,\check{\bm{B}}_t)$ around $(\hat{\bm{\alpha}}_t,\hat{\bm{B}}_t)$ such that $\check{\bm{B}}_t$ is negative semi-definite by solving the program
    \begin{align}\label{eq:program-check-B}
    (\check{\bm{\alpha}}_t,\check{\bm{B}}_t)\in \mathcal M_t := \left\{(\tilde{\bm{\alpha}},\tilde{\bm{B}}) : 
    \|(\tilde{\bm{\mathbfcal{B}}}-\hat{\bm{\mathbfcal{B}}}_{t})^\top \bm{e}_i\|_{\bm{\Lambda}_t}\le\kappa,   \|\tilde{{\mathbfcal{B}}}^\top \bm{e}_i\|_2 \leq 2L_B ~ \forall i\in[N] \text{~and~} \tilde{\bm{B}}+\tilde{\bm{B}}^\top  \preceq 0
    \right\},
    \end{align}
    {\small where we define the block matrices $\tilde{\bm{\mathbfcal{B}}}:=[\tilde{\bm{B}}|\tilde{\bm{\alpha}}]$ and $\hat{\bm{\mathbfcal{B}}}_t:=[\hat{\bm{B}}_t|\hat{\bm{\alpha}}_t]$ ($\bm{e}_i$ is the $i$-th canonical basis vector), $\bm{\Lambda}_t$ and $\kappa$ are defined in Eqs.~(\ref{eq:def-lambda},\ref{eq:kappa}) respectively.
    In the rare case when Eq.~\eqref{eq:program-check-B} is infeasible, we arbitrarily choose $(\check{\bm{\alpha}}_t,\check{\bm{B}}_t)$ as long as $\check{\bm{B}}_t$ is negative semi-definite and $\|\check{\bm{B}}_t^\top \bm{e}_i\|_2 \leq 2L_B ~ \forall i\in[N]$ (e.g., set $\check{\bm{B}}_t=0$).}

    \State \label{line:alg-main-UCB} Obtain  $\check{\bm{D}}_{t}(\bm{p})=\check{\bm{\alpha}}_t+\check{\bm{B}}_t \bm{p}$ and $\check{r}_{t}(\bm{p})=\Braket{\bm{p},\check{\bm{D}}_{t}(\bm{p})}$ to estimate $\bm{D}(\bm{p})$ and $r(\bm{p})$ respectively.
    \State \label{line:alg-main-primal-update} Update the primal variables
    \begin{align}
    \bm{p}_t \in \argmax _{\bm p \in [\underline{p},\bar p]^N}\left\{{\check{r}_{t}(\bm{p})-\bm{\mu}_t^\top \bm{A}\check{\bm{D}}_{t}(\bm{p})+2\Delta_t^f(\bm{p})}\right\}, \qquad \bm{s}_{t} \in \argmax_{  -\bm{\gamma}\leq\bm{s}\leq \bm{\gamma}}\left\{\phi(\bm{s})+\bm{\mu}_t^\top\bm{s}\right\} , \label{eq:primal-update}
    \end{align}
    {\small where $2\Delta_t^f(\bm{p})$ is the confidence radius of the \emph{adjusted reward} estimation $[\check{r}_{t}(\bm{p})-\bm{\mu}_t^\top \bm{A}\check{\bm{D}}_{t}(\bm{p})]$ which will be constructed later in Eq.~\eqref{eq:CI-f}. }
       
    \State Charge the price $\bm{p}_t$, observe the demand $\bm{d}_t$, consume resources $\bm{A}\bm{d}_t$ and update the inventory level $\bm{I}_t=\bm{I}_{t-1}-\bm{A}\bm{d}_t$ {\small (the algorithm stops whenever any resource is depleted)}.
      
    \State  \label{line:subgradient-g} Obtain an estimated subgradient of dual function $\mathfrak{q}(\cdot)$ at $ \bm{\mu}_t$: $\check{\bm{g}}_{t}=-\bm{A}\check{\bm{D}}_{t}(\bm{p}_t)+\bm{s}_{t}$.
      
    \State \label{line:alg-main-dual-update} Update the dual variable by invoking the mirror descent solver $\varsigma^{\textrm{D}}$ {\small (Definition~\ref{def:dualdescent})}:
    \begin{align}
    \bm{\mu}_{t+1}=\varsigma^{\textrm{D}}(\bm{\mu}_t,\check{\bm{g}}_{t} ;\D,\eta) , \label{eq:alg-dual-update}
    \end{align}
    where we set $\eta = \sqrt{\frac{C_1}{C_2T}}$ {\small ($C_1$ and $C_2$ are also defined in Definition~\ref{def:dualdescent})} and
    \begin{align}
    \D :=\{  \bm{\mu} \in \mathbb{R}^{M} \mid \|\bm{\mu}\|_1\leq C \}, \qquad C := L+((\bar r+\bar \phi)/\underline{\gamma}),  \qquad \underline{\gamma} :=\min_{i\in[M]}\gamma_i. \label{eq:dual-space}
    \end{align}
\EndFor
\end{algorithmic}
\end{algorithm} 
\section{Primal-dual Type Algorithm with Demand Learning}\label{sec:alg}
The pseudo-code of our main algorithm is given in Algorithm~\ref{algucb}.
 {In the following, we present an intuitive and high-level idea behind the design of the algorithm.}

To better introduce our algorithm design, we first imagine that the demand function was known and only explain the primal-dual framework. We then add the learning component for the demand function and address the additional challenges raised due to the unknown demand.

Given the demand function $\bm{D}(\cdot)$, thanks to Proposition~\ref{prop1}, we may use the fluid model $J_\mathrm{\bm{D}}$, which upper bounds $J_\mathrm{opt}$, to calculate an upper estimation of the regret of any online policy.\footnote{Indeed, this relaxation would not ruin our aimed $\widetilde{\mathcal{O}}(\sqrt{T})$ regret, as the difference between $J_\mathrm{\bm{D}}$ and $J_\mathrm{opt}$ is also $\widetilde{\mathcal{O}}(\sqrt{T})$ for the network revenue management problem either without \citep{gallego1997multiproduct} or with the global regularization (as we will see later in this paper).} We now focus on the primal formulation of the fluid model. 
 {To simplify the global regularization $\phi\left(\bm{A}\bm{\bm{D}}(\bm{p})\right)$, we introduce an auxiliary variable $\bm{s}$. With this new variable, the original optimization problem with one primal variable and an inequality constraint is transformed into one with two primal variables and an equality constraint:}
\begin{align}
    \mathfrak{p}^*&:= \max\limits_{\bm{p}\in[\underline{p},\overline{p}]^N}\{r(\bm{p})+\phi\left(\bm{A}\bm{\bm{D}}(\bm{p})\right) 
    \ \ \text{s.t.} \ \ \bm{A}\bm{\bm{D}}(\bm{p}) \leq  \bm{\gamma}\} \label{eq:primal-opt-def-1}\\
    &~=\max\limits_{\bm{p}\in[\underline{p},\overline{p}]^N, -\bm{\gamma}\leq\bm{s}\leq \bm{\gamma}}\{r(\bm{p})+\phi\left(\bm{s}\right) \ \
    \text{s.t.} \ \ \bm{A}\bm{\bm{D}}(\bm{p}) =\bm{s}\}. \nonumber
\end{align}
Note that we deliberately impose a lower bound constraint $\bm{s} \geq -\bm{\gamma}$ in the new formulation. This does not change the optimal value of the program since $\bm{A}\bm{\bm{D}}(\bm{p})$ is entry-wise non-negative for non-negative $A$ and $\bm{D}(\bm{p})$.

We then transform primal formulation into an unconstrained optimization problem using the Lagrangian dual method. By the well-known weak duality, we have that
\begin{align}
\mathfrak{p}^* &= \max_{\bm{p}\in[\underline{p},\overline{p}]^N,  -\bm{\gamma}\leq\bm{s}\leq \bm{\gamma}} \min_{\bm{\mu} \in \mathbb{R}^M} \left\{ r(\bm{p})+\phi\left(  \bm{s}\right)- \bm{\mu}^\top \bm{A}\bm{\bm{D}}(\bm{p}) + \bm{\mu}^\top   \bm{s} 
\right\} \nonumber \\
&\leq  \min_{\bm{\mu} \in \mathbb{R}^M} \max_{\bm{p}\in[\underline{p},\overline{p}]^N,  -\bm{\gamma}\leq\bm{s}\leq \bm{\gamma}}\left\{ r(\bm{p})+\phi\left(  \bm{s}\right)- \bm{\mu}^\top \bm{A}\bm{\bm{D}}(\bm{p})+ \bm{\mu}^\top   \bm{s}  \right\} . \label{eq:weak-duality}
\end{align}
In light of this, we define the dual function
\begin{align}\label{eq:def-dual-q}
\mathfrak{q}(  \bm{\mu}) :=\max_{\bm{p}\in[\underline{p},\overline{p}]^N,  -\bm{\gamma}\leq\bm{s}\leq \bm{\gamma}}\left\{ r(\bm{p})+\phi\left(  \bm{s}\right)- \bm{\mu}^\top \bm{A}\bm{\bm{D}}(\bm{p})+ \bm{\mu}^\top   \bm{s}
\right\}=r^\sharp(\bm{A}^\top   \bm{\mu})+ (-\phi)^*(\bm{\mu}),
\end{align}
where for every $\bm{\mu} \in \mathbb{R}^M$ we define
\begin{align}\label{eq:rsharp}
r^\sharp(\bm{A}^\top   \bm{\mu}) :=  \max_{\bm{p}\in[\underline{p},\overline{p}]^N} \left\{ r(\bm{p})- \bm{\mu}^\top \bm{A} \bm{D}(\bm{p})\right\} 
\end{align}
and the convex conjugate (following the convention, e.g., Chapter 3 in \cite{boyd2004convex})
\begin{align}\label{eq:dual-functions}
(-\phi)^*(\bm{\mu})  := \max _{  -\bm{\gamma}\leq\bm{s}\leq \bm{\gamma}}\left\{\phi(\bm{s})+\bm{\mu}^\top\bm{s}\right\}.
\end{align}
By Eq.~\eqref{eq:weak-duality}, for every $\bm{\mu} \in \mathbb{R}^M$, we have that 
\begin{align}\label{eq:weak-duality-2}
\mathfrak{p}^* \leq \mathfrak{q}(\bm{\mu}) .
\end{align}

To introduce the primal-dual framework, let us define the optimal dual solution $\bm{\mu}^*=\argmin_{\bm{\mu}\in\mathcal D}\mathfrak q(\bm{\mu})$. Suppose we had $\bm{\mu}^*$ in hand, it would be natural to make a good price decision according to 
\[
\argmax_{\bm{p}\in[\underline{p},\overline{p}]^N} \left\{ r(\bm{p})- [\bm{\mu}^*]^\top \bm{A} \bm{D}(\bm{p})\right\},
\]
which is indeed the optimal decision if strong duality holds. Since we do not know $\bm{\mu}^*$ a priori, in the primal-dual framework, we alternately optimize the primal variables $\bm{p}_t, \bm{s}_t$ (Line~\ref{line:alg-main-primal-update})\footnote{To illustrate the intuition, we assume the demand function is known. In Algorithm~\ref{algucb}, we need to deal with the upper confidence bounds of the estimated quantities when optimizing $\bm{p}_t$,  which will be explained soon.} based on the dual variable $\bm{\mu}_t$, 
\begin{align*}
    \bm{p}_t \in \argmax _{\bm p \in [\underline{p},\bar p]^N}\left\{{{r}(\bm{p})-\bm{\mu}_t^\top \bm{A}{\bm{D}}(\bm{p})}\right\}, \qquad \bm{s}_{t} \in \argmax_{  -\bm{\gamma}\leq\bm{s}\leq \bm{\gamma}}\left\{\phi(\bm{s})+\bm{\mu}_t^\top\bm{s}\right\}, 
\end{align*}
and optimize the dual variable $\bm{\mu}_t$ via calculating the subgradient of the dual function $\mathfrak{q}(\cdot)$ at $\bm{\mu}_t$ (Line~\ref{line:subgradient-g})\footnote{Again, here we assume the demand function is known.}:
\begin{align}\label{eq:subgradient-g-exact-def}
{\bm{g}}_{t}=-\bm{A}{\bm{D}}(\bm{p}_t)+\bm{s}_{t}.
\end{align}
Here $\bm s_t$ can be seen as the target resource-consumption vector related to the balancing regularization.

The optimal dual solution $\bm{\mu}^*$ is also known as the \emph{shadow price} in economics theory and revenue management optimization. It measures the change of the optimal revenue when given an additional unit of the constrained resource. In light of this, we refer to $[r(\bm{p})- \bm{\mu}^\top \bm{A} \bm{D}(\bm{p})]$ as the \emph{adjusted revenue function} with respect to $\bm p$ and $\bm \mu$. It represents the revenue that takes the economic value of the constrained resource into consideration. The subgradient $\bm{g}_t$ (Eq.~\eqref{eq:subgradient-g-exact-def}) also offers intuitive managerial insights on how the update of the dual variable $\bm{\mu}_t$ helps optimize the fair resource-consumption balancing objective: when the consumption of type-$i$ resource exceeds the resource-consumption target ($[\bm{A}{\bm{D}}(\bm{p}_t)]_i>[\bm s_t]_i$), $[\bm{g}_{t}]_i$ becomes negative, which in turn increases the shadow price $[\bm \mu_{t+1}]_i$ to discourage the consumption of type-$i$ resource; similar argument applies to the opposite case when $[\bm{A}{\bm{D}}(\bm{p}_t)]_i<[\bm s_t]_i$.

In Algorithm~\ref{algucb}, when updating the dual variable at Line~\ref{line:alg-main-dual-update}, we use the mirror descent method with the help of a mirror descent solver $\varsigma^{\text{D}}$ defined as follows.

\begin{definition}[Mirror Descent Solver]\label{def:dualdescent}
A mirror descent solver $\varsigma^{\textrm{D}}(\bm{\mu}_t,\check{\bm{g}}_{t}; \D,\eta)$ takes $\bm{\mu}_t \in \D$ and $\check{\bm{g}}_{t}$ as input and returns the updated dual variable $\bm{\mu}_{t+1} \in \D$ at each time $t$. For a sequence of input $\{\check{\bm{g}}_{t}\}$ and the initial dual variable $\bm{\mu}_1$, if we repeatedly apply $\varsigma^{\text{D}}$ and produce a sequence of dual variables $\{\bm{\mu}_t\}$. The solver makes sure that for all $\bm{\mu}\in\D$,
\begin{equation}\label{eq:ddsolver}
\sum_{t=1}^T\left\langle\bm{\mu}_{t}, \check{\bm{g}}_{t}\right\rangle\leq \sum_{t=1}^T\left\langle\bm{\mu}, \check{\bm{g}}_{t}\right\rangle+\frac{C_1}{\eta}+C_2\eta T,
\end{equation}
where $C_1$ and $C_2$ are constants that only depend on $\D$.
\end{definition} 
In other words, the mirror descent solver should generate a sequence $\{\bm{\mu}_t\}$ to minimize $\sum_{t=1}^T\left\langle\cdot, \check{\bm{g}}_{t}\right\rangle$ against any stationary benchmark with the regret at most $C_1/\eta+C_2\eta T$.

The above-described primal-dual framework is similar to (and inspired by) the algorithm proposed in \cite{balseiro2020regularized}. However, the key differences are two folds explained as follows.

\noindent \underline{\bf Demand Learning.} In contrast to the known demand function in \cite{balseiro2020regularized}, the demand function is not known to the decision-maker beforehand in our setting. Our algorithm learns the parameterized demand function from historical data via the regularized least-squares estimate (Line~\ref{line:alg-main-RLS}). We then solve another convex program (Line~\ref{line:alg-main-check-B}) to make sure the estimated parameters $(\check{\bm{\alpha}}, \check{\bm{B}})$ are bounded and $\check{\bm{B}}$ is negative semi-definite. Finally, we use the Upper Confidence Bound of the {adjusted revenue function} $[\check{r}_{t}(\bm{p})-\bm{\mu}_t^\top \bm{A}\check{\bm{D}}_{t}(\bm{p})]$ (Line~\ref{line:alg-main-primal-update}) to compute the primal update. We will explain this in more detail in Section~\ref{sec:ucb}.

An additional feature of our demand estimator is that the reward Upper Confidence Bound is defined based on the $\ell_\infty$-norm of $\bm{\Lambda}_t^{-1/2}\p_t$ where the $\ell_2$-norm is usually adopted in the linear bandit literature. Together with the convex program solved in Line~\ref{line:alg-main-check-B}, our definition of the reward Upper Confidence Bound renders the primal update a combination of a few convex optimization problems (Eq.~\eqref{eq:primal-update}) that can be efficiently solved, which will be further explained in Section~\ref{sec:optim}.\footnote{On the downside, we sacrifice an $\mathcal{O}(\sqrt{N})$ factor in the regret bound. However, we view this degradation as relatively small compared to the existing $\mathcal{O}(N^2)$ factor which seems necessary in the regret due to the $N^2$ parameters in $B$ to learn.} 

\noindent \underline{\bf A New Dual Space.} The dual space $\D$ is a crucial component in the design of the mirror descent solver $\varsigma^{\text{D}}$ and affects the regret analysis. \cite{balseiro2020regularized} adopt a dual space $\mathcal{{D}}^{\mathrm{Bal.}} = \{\bm{\mu} \in \mathbb{R}^M ~|~ \sup_{\bm{a} \leq \bm{\gamma}} {\phi(\bm{a}) + \bm{\mu}^\top \bm{a}}\}$ which might have different shapes for different balancing  regularizer $\phi(\cdot)$. The fundamental reason that we cannot directly adopt $\mathcal{{D}}^{\mathrm{Bal.}}$ in our problem, however, is the unboundedness of $\mathcal{{D}}^{\mathrm{Bal.}}$ which would lead to an unbounded regret due to the unbounded estimation error of the adjusted reward $[\check{r}_{t}(\bm{p})-\bm{\mu}_t^\top \bm{A}\check{\bm{D}}_{t}(\bm{p})]$ during the learning process (see Eq.~(\ref{eq:boundmu}) for more details). 
To deal with this issue, we construct a novel, simply and uniformly shaped, and bounded dual space (Eq.~\eqref{eq:dual-space}). We prove that our dual space encompasses all potential stationary benchmark dual variables $\overline{\bm{\mu}}$ which is necessary for the desired $\widetilde{\mathcal{O}}(\sqrt{T})$ regret. 

Thanks to the newly constructed dual space, an extra benefit enjoyed by our algorithm is that, together with a carefully chosen variant of the exponentiated gradient descent ($\mathrm{EG}\pm$) algorithm as the mirror descent solver, we are able to obtain a \emph{uniform} and \emph{closed-form} update of the dual variables for \emph{all} balancing regularizers. In contrast, \cite{balseiro2020regularized} have to design the dual update step on a case-by-case basis for the balancing regularizers. Also, the closed-form update improves the computational efficiency of the algorithm and is a desired feature in \cite{balseiro2020regularized} that is partially achieved for a few selected balancing regularizers.

\section{Demand and Reward Estimation}\label{sec:ucb}

The regularized least-squares estimator (Line~\ref{line:alg-main-RLS} of Algorithm~\ref{algucb}) for demand parameters is frequently used in the linear bandit literature (see, e.g., \cite{dani2008stochastic,rusmevichientong2010linearly,abbasi2011improved}). However, as mentioned before, we need to work with the upper confidence bound of a specially defined \emph{adjusted revenue function}. Also, we employ an additional step (Line~\ref{line:alg-main-check-B}) to make sure the estimated parameters $(\check{\bm{\alpha}}, \check{\bm{B}})$ are bounded, which is crucial to the regret analysis (more specifically, the analysis of the mirror descent solver). Line~\ref{line:alg-main-check-B} also guarantees the negative semi-definiteness of $\check{\bm{B}}$; when computing the Upper Confidence Bound for the estimation, we use an $\ell_\infty$-norm confidence radius instead of the usual $\ell_2$-norm confidence radius  -- both ingredients help the algorithm to compute the upper confidence bound \emph{in polynomial time}. We will show how to computationally efficiently find $(\check{\bm{\alpha}}, \check{\bm{B}}) \in \M_t$ (Line~\ref{line:alg-main-check-B}) and implement the UCB-type primal update (Line~\ref{line:alg-main-primal-update}) in Section~\ref{sec:find} and Section~\ref{sec:optim} respectively.

To explain our Upper Confidence Bound method in more detail, we first introduce some notations.
For convenience, we define the stopping time 
\[
\tau=\max \left\{t: \min_{i} [\bm{I}_t]_i>0,t\leq T\right\},
\]
which is the last time period when the inventory levels of all resources remain positive. Most of our analysis will be done only for time periods up to $\tau$. 
We define
\begin{align}
f_{t}(\bm{p}):=r(\bm{p})-\bm{\mu}_t^\top \bm{A}\bm{\bm{D}}(\bm{p})
\end{align}
to be the \emph{adjusted revenue function} at price $p$ and with respect to $\bm{\mu}_t$. Note that this corresponds to the optimization objective in Eq.~\eqref{eq:rsharp} when $\bm{\mu} = \bm{\mu}_t$. Our estimation for $f_t(\bm{p})$ is 
\begin{align}
\check{f}_{t}(\bm{p}):=\check{r}_{t}(\bm{p})-\bm{\mu}_t^\top A\check{\bm{D}}_{t}(\bm{p}), 
\end{align}
which corresponds to the first part in the optimization objective of $\bm p_t$ in Eq.~\eqref{eq:primal-update}.  We also define estimators with regard to $(\hat{\bm{\alpha}}_t,\hat{\bm{B}}_t)$ as 
\[ \hat{\bm{D}}_{t}(\bm{p})=\hat{\bm{\alpha}}_t+\hat{\bm{B}}_t p,\qquad \hat{r}_{t}(\bm{p})=\Braket{\bm{p}_t,\hat{\bm{D}}_{t}(\bm{p})}, \qquad  \hat{f}_{t}(\bm{p}):=\hat{r}_{t}(\bm{p})-\bm{\mu}_t^\top A\hat{\bm{D}}_{t}(\bm{p}).
\]

\noindent \underline{\bf Bounding the estimation errors.} We now derive the estimation errors of $\hat{\bm{D}}_{t}, \hat{r}_{t},  \hat{f}_{t}$ and  $\check{\bm{D}}_{t}, \check{r}_{t},  \check{f}_{t}$, as well as their corresponding upper confidence bounds.
For any price (column) vector $\bm{p}$, we let $\widetilde{\bm{p}} := (\bm{p}^\top, 1)^\top$, and then introduce the \emph{regularized information matrix} at time $t$ to be
\begin{align}\label{eq:def-lambda}
\bm{\Lambda}_t  :=(N+1)\cdot\mathbf{I}_{N+1}+\sum_{s<t}\widetilde{\bm p}_s\widetilde{\bm p}_s^\top .
\end{align}
Let 
\vspace{-3ex}
\begin{align}
\label{eq:kappa}
&\kappa :=2\sqrt{2\bar d^2(N+1)\ln\left(NT(1+\bar p^2T)\right)+2(N+1)L_B^2},\qquad \text{and} \\
\label{eq:CI-D-r}
& \Delta_{t}^{\bm{D}}(\bm{p}) :=\sqrt{N+1}\k \|\bm{\Lambda}_t^{-1/2}\bm{\p}_t\|_\infty \qquad \text{and} \qquad \Delta_{t}^{r}(\bm{p}) :=\sqrt{N+1}N\bar p \k \|\bm{\Lambda}_t^{-1/2}\bm{\p}_t\|_\infty
\end{align}
to be the confidence radii of $\hat{r}_t(\bm{p})$ and $\hat{\bm{D}}_t(\bm{p})$ respectively. Note that here we use $\|\bm{\Lambda}_t^{-1/2}\bm{\p}_t\|_\infty$ instead of the $\ell_2$-norm confidence radius $\|\bm{\Lambda}_t^{-1/2}\bm{\p}_t\|_2$ commonly seen in literature. We finally define 
\begin{align}\label{eq:CI-f}
\Delta_t^f(\bm{p}) := \Delta_t^r(\bm{p})+\|\bm{\mu}_t\|_1 \cdot \|\bm{A}\|_\infty\Delta_t^{\bm{D}}(\bm{p})
\end{align}
to be the confidence radius for the adjusted reward estimator $\hat{f}_t$. We utilize the famous Confidence Ellipsoid Lemma in \cite{abbasi2011improved}
to analyze our $\ell_\infty$-type confidence region, and have the following lemma. Our lemma states that the confidence radii defined above (Eq.~\eqref{eq:CI-D-r} and Eq.~\eqref{eq:CI-f}) hold with overwhelming probability.
\begin{lemma}\label{le:ucb0}
With probability at least $(1-\mathcal O\left(T^{-1}\right))$, for all $t\leq\tau$ and all $\bm{p}\in[\underline{p},\bar p]^N$, we have 
\[
\|\hat{\bm{D}}_{t}(\bm{p})-\bm{D}(\bm{p})\|_{\infty} \leq \Delta_{t}^{\bm{D}}(\bm{p}), \qquad 
\left|\hat{r}_{t}(\bm{p})-r(\bm{p})\right| \leq \Delta_{t}^{r}(\bm{p}), \qquad \text{and} \qquad
|\hat{f}_t(\bm{p}) - f_t(\bm{p})| \leq \Delta_t^f(\bm{p}).
\]
\end{lemma}

As we will later show  in Lemma~\ref{le:find}, with probability at least $(1-\mathcal O\left(T^{-1}\right))$, we are able to find a feasible $(\check{\bm{\alpha}}_t,\check{\bm{B}}_t) \in \mathcal{M}_t$ in Line~\ref{line:alg-main-check-B}. Combining the definition of $\mathcal{M}_t$ and Lemmas~\ref{le:ucb0} and \ref{le:find}, we have the following corollary. \begin{corollary}\label{cor:check-error-bound}
With probability at least $(1-\mathcal O\left(T^{-1}\right))$, for all $t\leq\tau$ and all $\bm{p}\in[\underline{p},\bar p]^N$, we have 
\[
\|\check{\bm{D}}_{t}(\bm{p})-\bm{D}(\bm{p})\|_{\infty} \leq 2\Delta_{t}^{\bm{D}}(\bm{p}), \qquad 
\left|\check{r}_{t}(\bm{p})-r(\bm{p})\right| \leq 2\Delta_{t}^{r}(\bm{p}), \qquad \text{and} \qquad
|\check{f}_t(\bm{p}) - f_t(\bm{p})| \leq 2\Delta_t^f(\bm{p}).
\]
\end{corollary}

\noindent \underline{\bf The program $\M_t$.} Note that by Lemma~\ref{le:ucb0}, $\hat{\bm{D}}_t, \hat{r}_t, \hat{f}_t$ already serve as good estimators. However, in the rest part of the algorithm (as well as the analysis), we will mainly work with $\check{\bm{D}}_t, \check{r}_t, \check{f}_t$, which are defined based on $(\check{\bm{\alpha}}_t, \check{\bm{B}}_t)$ derived by solving the program $\M_t$ in Line~\ref{line:alg-main-check-B} of Algorithm~\ref{algucb}. This is due to the following two requirements.
\begin{enumerate}
\item The analysis of the mirror descent solver requires an upper bound on the estimated gradient $\|\check{\bm{g}}_t\|_\infty$ which relies on the bound of $\max_i \|\check{\bm{B}}_t^\top \bm{e}_i\|_2$ (Eqs.~(\ref{eq:boundcheckD},\ref{eq:bound_checkg})). 
\item The primal update (Eq.~\eqref{eq:primal-update}) involves maximizing $\check{r}_t$, a quadratic form of $\check{\bm{B}}_t$, which can be efficiently optimized only when $\check{\bm{B}}_t$ is negative semi-definite so that $\check r_t$ is concave.
\end{enumerate}
By solving the program $\mathcal{M}_t$, we find $(\check{\bm{\alpha}}_t, \check{\bm{B}}_t)$ that simultaneously satisfies the above two requirements and stays close to $(\hat{\bm{\alpha}}_t, \hat{\bm{B}}_t)$ (in terms of the $\|\cdot \|_{\bm{\Lambda}_t}$ norm). In this way, we facilitate both the regret analysis and the efficient computation of the algorithm. Please refer to Sections~\ref{sec:find} and \ref{sec:optim} for the efficient implementations of solving $\mathcal{M}_t$ and the primal update respectively.

\noindent \underline{\bf UCB of the adjusted revenue function.} When the desired event in Corollary~\ref{cor:check-error-bound} happens, we define 
\begin{align}
\bar{f}_{t}(\bm{p}):= \check{f}_{t}(\bm{p})+2\Delta_t^f(\bm{p}) 
\end{align}
and have that $f_t(\bm{p}) \leq \bar{f}_t(\bm{p})$ for all $t \leq \tau$ and $\bm{p} \in [\underline{p}, \bar p]^N$. Note that $\bar{f}_{t}(\cdot)$ is exactly the optimization objective of $\bm{p}_t$ (at Line~\ref{line:alg-main-primal-update} of Algorithm~\ref{algucb}), which is indeed an Upper Confidence Bound (UCB) of the maximization objective in $r^\sharp(\cdot)$ (Eq.~\eqref{eq:rsharp}, namely $f_t(\cdot)$).

Since $\|\bm{\mu}_t\|_1\in\mathcal D$ for all $t\in[T]$ and $\D =\{  \bm{\mu} \in \mathbb{R}^{M} \mid \|\bm{\mu}\|_1\leq C \}$, we further upper bound $\Delta_t^f(\bm{p})$ by
\begin{equation}\label{eq:boundmu}
\Delta_t^f(\bm{p}) \leq\Delta_t^r(\bm{p})+C \|\bm{A}\|_\infty\Delta_t^{\bm{D}}(\bm{p}).
\end{equation}
Therefore, we set 
\vspace{-3ex}
\begin{align}
\Delta_t(\bm{p}) := \Delta_t^r(\bm{p})+ \max\{C\|\bm{A}\|_\infty, 1\} \Delta_t^{\bm{D}}(\bm{p})
\end{align}
to upper bound all the confidence radii $\Delta_t^r(\bm{p})$, $\Delta_t^{\bm{D}}(\bm{p})$ and $\Delta_t^f(\bm{p})$.

\noindent \underline{\bf Bounding the total estimation error.} In our regret analysis, we will relate the regret incurred at time $t$ to the confidence radii at price $\bm p_t$ at the time (which aligns with the general Upper Confidence Bound principle -- bounding the regret by the confidence radii of the selected actions). And thus we will be interested in the summation of the estimation errors. The following lemma adapts the celebrated Elliptical Potential Lemma (see, e.g., Theorem 11.7 in \cite{cesa2006prediction} and Lemma 9 in \cite{dani2008stochastic}) 
to upper bound the total estimation error.
\begin{lemma}\label{le:ucb00}
With probability $1$, we have the following upper bound for the total estimation error:
\begin{align*}
 \sum_{t=1}^\tau \Delta_{t}\left(\bm{p}_t\right)&\leq\O\left(\sqrt{N+1}\kappa\max\{\bar p,1\}(N\bar p+\max\{C\|\bm{A}\|_\infty, 1\})
 \right)\times\sqrt{NT\log(N+1+\bar p^2T)},
\end{align*}
where only a universal constant is hidden in the $\mathcal{O}(\cdot)$ notation.
\end{lemma}
\subsection{Solving $\mathcal{M}_t$ via Ellipsoid Method}\label{sec:find}

In this subsection, we describe how to implement Line~\ref{line:alg-main-check-B} and find a feasible solution to $\mathcal{M}_t$ in polynomial time via the Ellipsoid method. The main lemma of this subsection is Lemma~\ref{le:find}.

We first introduce the definition of a separation oracle for a convex set $K$, which is closely related to the Ellipsoid method.

\begin{definition}[Separation Oracle]
For a closed convex set $K\subseteq \R^n$, a \emph{separation oracle} for $K$, namely $\mathrm{SEP}_K$, is an algorithm that takes a point $\bm{x}\in\R^n$ as input and correctly decides whether $\bm{x}\in K$. In the case that $\bm{x} \not\in K$, the separation oracle also returns a hyperplane that separates $\bm{x}$ from $K$. The hyperplane may be characterized by its norm vector $\bm{c}\in\R^n$ such that $\bm{c}^\top \bm{x} > \bm{c}^\top \bm{y}$ for all $\bm{y}\in K$.
\end{definition}

The ellipsoid method reduces a convex program feasibility problem to the construction of an efficient separation oracle for the corresponding convex body. The following lemma characterizes such a reduction. The lemma is a simplification of Theorem 3.2.1 in \cite{grotschel2012geometric} modulo the numerical error due to the arithmetic operations on real numbers.\footnote{The numerical error analysis is often tedious but straightforward, which is also the case in this subsection. Therefore, we choose to omit this part and emphasize the main algorithmic idea more clearly.}

\begin{lemma}\label{le:ellipsoid-convergence}
Suppose we could perform exact arithmetic operations on real numbers.  Let $\mathrm{Ball}(\bm{x}, r)$ denote the closed ball with radius $r$ and centered at $\bm{x} \in \mathbb{R}^n$. Given a closed convex set $K \subseteq \mathbb{R}^n$, suppose that there exist $R, r > 0$ such that $K \subseteq \mathrm{Ball}(\bm{x}_0, R)$ and $\mathrm{Ball}(\bm{x}_1, r) \subseteq K$ for some $\bm{x}_0, \bm{x}_1 \in \mathbb{R}^n$. Given $R$, $r$, $\bm{x}_0$, and a separation oracle for $K$, namely $\mathrm{SEP}_K$, the Ellipsoid method will return a point in $K$ using $\mathcal O(n^2 \log(R/r))$ calls to the separation oracle and  {$O(n^4 \log (R/r))$}  arithmetic operations.
\end{lemma}

It is easy to verify that our $\M_t$ is a closed convex set in $\mathbb{R}^{N \times (N+1)}$. To apply Lemma~\ref{le:ellipsoid-convergence} to $\M_t$, we first upper and lower bound the shape of $\M_t$ as follows.
\begin{lemma}\label{le:R_r}
Given the desired event described in Lemma~\ref{le:ucb0}, we have that
\[
\M_t\subseteq \mathrm{Ball}(\hat{\bm{\mathbfcal{B}}}_t, \kappa \sqrt{N}) \qquad \text{and} \qquad \mathrm{Ball}([\bm{B} - T^{-2} \cdot \mathbf{I}_{N} |\bm{\alpha}], T^{-4}) \subseteq \M_t ,
\]
where we treat the matrices $\hat{\bm{\mathbfcal{B}}}_t$ and $[\bm{B} - T^{-2} \cdot \mathbf{I}_{N} |\bm{\alpha}]$ as $N \times (N+1)$-dimensional vectors.
\end{lemma}

\noindent \underline{\bf The separation oracle.} It remains to design the separation oracle $\mathrm{SEP}_{\M_t}$. Given $\tilde{\bm{\mathbfcal{B}}} = [\tilde{\bm{B}}|\tilde{\bm{\alpha}}]$, we need to verify the following two types of constraints specified in the definition of $\M_t$ (Eq.~\eqref{eq:program-check-B}).
\begin{itemize}
\item $\|(\tilde{\bm{\mathbfcal{B}}}-\hat{\bm{\mathbfcal{B}}}_{t})^\top \bm{e}_i\|_{\bm{\Lambda}_t}\le\kappa, \forall i\in[N]$. This condition can be verified for each $i \in [N]$ by straightforward computation. When the condition is not met for some $i \in [N]$, we have that $\kappa < \|(\tilde{\bm{\mathbfcal{B}}}-\hat{\bm{\mathbfcal{B}}}_{t})^\top \bm{e}_i\|_{\bm{\Lambda}_t} = \|\bm{\Lambda}_t^{1/2}(\tilde{\bm{\mathbfcal{B}}}-\hat{\bm{\mathbfcal{B}}}_{t})^\top \bm{e}_i\|_2$, and there exists $\bm{c} = \frac{\bm{\Lambda}_t^{1/2}(\tilde{\bm{\mathbfcal{B}}}-\hat{\bm{\mathbfcal{B}}}_{t})^\top \bm{e}_i}{ \|\bm{\Lambda}_t^{1/2}(\tilde{\bm{\mathbfcal{B}}}-\hat{\bm{\mathbfcal{B}}}_{t})^\top \bm{e}_i\|_2}$ such that 
\[
\bm{c}^\top \bm{\Lambda}_t^{1/2}(\tilde{\bm{\mathbfcal{B}}}-\hat{\bm{\mathbfcal{B}}}_{t})^\top \bm{e}_i > \kappa \geq \bm{c}^\top \bm{\Lambda}_t^{1/2}(\mathbfcal{B}'-\hat{\bm{\mathbfcal{B}}}_{t})^\top \bm{e}_i 
\]
for every $\mathbfcal{B}' = [\bm{B}' |\bm{\alpha}']$ where $(\bm{\alpha}', \bm{B}') \in \M_t$, which defines the separation hyperplane.
\item $\|\tilde{\bm{\mathbfcal{B}}}^\top \bm{e}_i\|_2\le2L_B, \forall i\in[N]$. This condition can also be verified for each $i \in [N]$ by straightforward computation. When the condition is not met for some $i \in [N]$, we have that $2L_B < \|\tilde{\bm{\mathbfcal{B}}}^\top \bm{e}_i\|_2$, and there exists $\bm{c} = \frac{\tilde{\bm{\mathbfcal{B}}}^\top \bm{e}_i}{ \|\tilde{\bm{\mathbfcal{B}}}^\top \bm{e}_i\|_2}$ such that 
\[
\bm{c}^\top\tilde{\bm{\mathbfcal{B}}}^\top \bm{e}_i > 2L_B \geq \bm{c}^\top \mathbfcal{B}'^\top \bm{e}_i 
\]
for every $\mathbfcal{B}' = [\bm{B}' |\bm{\alpha}']$ where $(\bm{\alpha}', \bm{B}') \in \M_t$, which defines the separation hyperplane.
\item $\tilde{\bm{B}}+\tilde{\bm{B}}^\top  \preceq 0$. This condition is equivalent to $\lambda_{\max}(\tilde{\bm{B}}+\tilde{\bm{B}}^\top) \leq 0$ which can be efficiently verified. If the condition is not satisfied, we can efficiently find a vector $\bm{c} \in \mathbb{R}^N$ such that 
\[
\langle \tilde{\bm{B}}+\tilde{\bm{B}}^\top, \bm{c} \bm{c}^\top \rangle > 0 \geq \langle \bm{B}'+(\bm{B}')^\top, \bm{c} \bm{c}^\top \rangle 
\]
for every $(\bm{\alpha}', \bm{B}')\in \M_t$, which defines the separation hyperplane.
\end{itemize}

 {The above separation oracle can be implemented using $O(N^3)$ arithmetic operations (required by both the first and the third steps).} Combining Lemma~\ref{le:R_r}, and the separation oracle constructed above, we may invoke Lemma~\ref{le:ellipsoid-convergence} with  {$n = N^2$, $R = \kappa \sqrt{N}$ (therefore $\log R \leq O(\log ( NT \bar{d} \bar p L_B))$) and $r = T^{-4}$}, and conclude this subsection with the following lemma.

\begin{lemma}\label{le:find}
With probability at least $(1-\mathcal O\left(T^{-1}\right))$, for all $t\leq\tau$, $\mathcal M_t$ is feasible, and we can find $ (\check{\bm{\alpha}}_t,\check{\bm{B}}_t)\in \mathcal M_t$  via the Ellipsoid method  {using $O(N^4 \log (N T \bar{d}  \bar p L_B))$ calls to the separation oracle and  $O(N^8 \log (N T \bar{d} \bar p L_B))$ arithmetic operations on real numbers}.
\end{lemma}

\subsection{Efficient Primal Update}\label{sec:optim}

We now show that thanks to the new $\ell_\infty$-norm-based confidence region, we may efficiently implement the primal update (Line~\ref{line:alg-main-primal-update}) by solving $\mathcal{O}(N)$ convex optimization problems. We focus on the optimization problem for $\bm p_t$ as the one for $\bm s_t$ is already convex. Note that
\begin{align*}
&\max _{\bm{p} \in [\underline{p},\bar p]^N}\left\{\check{f}_{t}(\bm{p})+2\Delta_t^f(\bm{p})\right\} \\& =\max_{\bm{p} \in [\underline{p},\bar p]^N} \left\{\Braket{\bm{p}-\bm{A}^\top\bm{\mu}_t,\check{\bm{D}}_{t}(\bm{p})}+ 2\sqrt{N+1}\kappa (N\bar p +\|\bm{\mu}_t\|_1\cdot\|\bm{A}\|_\infty) \|\bm{\Lambda}_t^{-1/2}\bm{\p}\|_\infty\right\} \\
&=\max_{\bm{p} \in [\underline{p},\bar p]^N} \left\{\Braket{\bm{p}-\bm{A}^\top\bm{\mu}_t,\check{\bm{D}}_{t}(\bm{p})}+2\sqrt{N+1}\kappa (N\bar p +\|\bm{\mu}_t\|_1\cdot\|\bm{A}\|_\infty)  \max\limits_{\bm{\lambda}\in\{\pm \bm{e}_1,\dots,\pm \bm{e}_{N+1}\}}\bm{\lambda}^\top\bm{\Lambda}_t^{-1/2}\bm{\p}\right\}\\
&=\max\limits_{\bm{\lambda}\in\{\pm \bm{e}_1,\dots,\pm \bm{e}_{N+1}\}} \max_{\bm{p} \in [\underline{p},\bar p]^N} \left\{\Braket{\bm{p}-\bm{A}^\top\bm{\mu}_t,\check{\bm{D}}_{t}(\bm{p})}+2\sqrt{N+1}\kappa (N\bar p+\|\bm{\mu}_t\|_1\cdot\|\bm{A}\|_\infty)  \bm{\lambda}^\top\bm{\Lambda}_t^{-1/2}\bm{\p}\right\},
\end{align*}
where $\bm{e}_i$ ($i \in \{1, 2, \dots, N+1\})$ is the $i$-th canonical basis vector in $\mathbb{R}^{N+1}$. For any $\lambda\in\{\pm e_1,\dots,\pm e_{N+1}\}$, define the convex program (which is convex due to the negative semi-definiteness of $\check{\bm{B}}$ guaranteed in Line~\ref{line:alg-main-check-B})
\[
P_t^{(\bm{\lambda})} := \argmax_{\bm{p} \in [\underline{p},\bar p]^N} \left\{\Braket{\bm{p}-\bm{A}^\top\bm{\mu}_t,\check{\bm{D}}_{t}(\bm{p})}+ 2\sqrt{N+1}\kappa(N\bar p+\|\bm{\mu}_t\|_1\cdot\|\bm{A}\|_\infty)  \bm{\lambda}^\top\bm{\Lambda}_t^{-1/2}\bm{\p}\right\} .
\]

It is easy to verify that $\argmax_{\bm{p} \in [\underline{p},\bar p]^N}\left\{\check{f}_{t}(\bm{p})+2\Delta_t^f(\bm{p})\right\} \subseteq \cup_{\bm{\lambda} \in\{\pm \bm{e}_1,\dots,\pm \bm{e}_{N+1}\}} P_t^{(\bm{\lambda})}$. Therefore, to compute the primal update for $\bm p_t$, we only need to first solve $2(N+1)$ convex programs to identify $\bm{p}_t^{(\bm{\lambda})} \in P_t^{(\bm{\lambda})}$ for every $\bm{\lambda} \in\{\pm \bm{e}_1,\dots,\pm \bm{e}_{N+1}\}$, and then select 
\begin{equation}\label{eq:updatept}
\bm{p}_t \in \argmax\limits_{\bm{p} \in \{\bm{p}_t^{(\bm{\lambda})} : \bm{\lambda} \in \{\pm \bm{e}_1, \dots, \pm \bm{e}_{N+1}\}\}} \left\{\Braket{\bm{p}-\bm{A}^\top\bm{\mu}_t,\check{\bm{D}}_{t}(\bm{p})} + 2\sqrt{N+1}\kappa(N\bar p+\|\bm{\mu}_t\|_1\cdot\|\bm{A}\|_\infty) \|\bm{\Lambda}_t^{-1/2}\bm{\p}\|_\infty\right\}.
\end{equation}

\section{Mirror Descent Solver $\varsigma^{\textrm{D}}$ and its Closed-form Dual Update}\label{sec:mirrordescent}
In this section, we design the mirror descent solver $\varsigma^{\textrm{D}}$ to satisfy Definition~\ref{def:dualdescent}. Given the dual space $\D$, for any \emph{reference function} $h$ that is $\sigma$-strongly convex with respect to $\|\cdot\|_1$ over $\D$, the online mirror descent (OMD) algorithm operates in the following way to update the dual variable:
\begin{align}
\bm{\mu}_{t+1} \in \argmin_{\bm{\mu}\in\mathcal D} \left\{ \langle  \bm{\mu}, \check{\bm{g}}_{t}\rangle + \frac{1}{\eta} D_h(\bm{\mu}, \bm{\mu}_t) \right\}, \label{eq:OMD-update}
\end{align}
where $D_{h}(\bm{x}, \bm{y})=h(\bm{x})-h(\bm{y})-\nabla h(\bm{y})^{\top }(\bm{x}-\bm{y})$ is the Bregman divergence. It is well-known (see, e.g., \cite{hazan2016introduction}) that if $\left\|\check{\bm{g}}_t\right\|_{\infty} \leq G$ for all $t$, then if we start with any given $\bm{\mu}_1 \in \D$, the $\{\bm{\mu}_t\}$ sequence produced by Eq.~\eqref{eq:OMD-update} guarantees that for any stationary benchmark $\bm{\mu} \in \D$,
\[
\sum_{t=1}^T\left\langle\bm{\mu}_{t}, \check{\bm{g}}_{t}\right\rangle\leq \sum_{t=1}^T\left\langle\bm{\mu}, \check{\bm{g}}_{t}\right\rangle+\frac{\sup_{\bm{\mu}\in\D}D_{h}\left(\bm{\mu},\bm{\mu}_{1}\right)}{\eta}+\frac{\eta G^{2}}{2 \sigma}T ,
\]
which matches the requirement of Definition~\ref{def:dualdescent}.

The popular choices of the reference functions are the negative entropy function $h(\bm{x}) = \sum_{i=1}^{n}x_i\ln x_i$ (so that $D_h(\bm{x}, \bm{y}) = \sum_{i=1}^{n}x_i\ln (x_i/y_i)$, and the OMD algorithm becomes the exponentiated gradient algorithm) and the Euclidean norm $h(\bm{x}) = \frac12 \|\bm{x}\|_2^2$ (so that $D_h(\bm{x}, \bm{y}) = \frac12\|\bm{x}-\bm{y}\|_2$ and the OMD algorithm becomes the projected gradient descent algorithm). However, based on the different shapes of the dual space $\D$, one has to carefully choose $h$ to guarantee its strong convexity and proper definition (e.g., the negative entropy function is not properly defined when any of the coordinates becomes negative). Due to this reason, \cite{balseiro2020regularized} have to design the reference function on a case-by-case basis for various balancing  regularizers $\phi$ which shape their dual space $\D$. When designing $h$, \cite{balseiro2020regularized} also aim to simplify the update rule (Eq.~\eqref{eq:OMD-update}) with the hope of a closed-form update, so as to reduce the computational cost. However, they are only able to achieve this goal for selected balancing regularizers.

In our work, thanks to the simplicity of newly designed dual space $\D =\{  \bm{\mu} \in \mathbb{R}^{M} \mid \|\bm{\mu}\|_1\leq C \}$ (Eq.~\eqref{eq:dual-space}), we are able to use a uniform mirror descent solver $\varsigma^{\text{D}}$ that enjoys the closed-form update for all balancing regularizers. Our $\varsigma^{\text{D}}$ is similar to the OMD algorithm with the negative entropy function. The only issue, however, is that the negative entropy function does not apply to negative coordinates covered by our dual space $\D$. To this end, we employ the special variant of the algorithm that separately deals with the positive weights and negative weights in $\bm{\mu}_t$. The algorithm was proposed by \cite{kivinen1997exponentiated} and called $\EG^\pm$ (Exponentiated Gradient Algorithm with Positive and Negative Weights).

The $\EG^\pm$ algorithms is formally described in Algorithm~\ref{alg:eg}. Note that instead of a single vector $\bm{\mu}_t$, the algorithm keeps two vectors $\bm{\mu}^+_{t+1}$ and $\bm{\mu}^-_{t+1}$, and the update of the two vectors are in simple closed forms. While both vectors are in $\mathbb{R}_{+}^M$, they respectively represent (the absolute values of) the positive and negative weights in $\bm{\mu}_t$ (see Eq.~\eqref{eq:EG-pm-dual-update}). Due to this technical reason, to use $\EG^\pm$ as our mirror descent solver, we need to slightly modify the description of our main Algorithm~\ref{algucb}. First, we initialize the two vectors as 
\vspace{-3ex}
\begin{align}
\bm{\mu}^+_{1} = \bm{\mu}^-_{1} = (C/M,\dots,C/M)^\top, \label{eq:EG-pm-initialization}
\end{align}
which replaces the initialization (Line~\ref{line:alg-main-initialization}) of Algorithm~\ref{algucb}. We also replace the dual update (Eq.~\eqref{eq:alg-dual-update}) of Algorithm~\ref{algucb} by
\vspace{-3ex}
\begin{align}
(\bm{\mu}^+_{t+1},\bm{\mu}^-_{t+1}) = \EG^\pm(\bm{\mu}^+_t,\bm{\mu}^-_t,\check{\bm{g}}_{t}; \D,\eta), \qquad \qquad \bm{\mu}_{t+1} = \bm{\mu}^+_{t+1} - \bm{\mu}^-_{t+1} . \label{eq:EG-pm-dual-update}
\end{align}

\begin{algorithm}[!t]
\caption{$\EG^\pm(\bm{\mu}^+_t,\bm{\mu}^-_t,\check{\bm{g}}_{t}; \D,\eta)$}
\label{alg:eg}
\begin{algorithmic}[1]
\State Compute the $\bm{\mu}^+_{t+1}$ and $\bm{\mu}^-_{t+1}$ vectors as follows:
\For{$i = 1,2,\dots,M$}
\begin{align*}
[\bm{\mu}_{t+1}^+]_i& =\frac{C[\bm{\mu}_t^+]_i\exp(-\eta C[\check{\bm{g}}_{t}]_i)}{\sum_{i=1}^{M}\left([\bm{\mu}_t^+]_i\exp(-\eta C[\check{\bm{g}}_{t}]_i)+[\bm{\mu}_t^-]_i\exp(\eta C[\check{\bm{g}}_{t}]_i)\right)} , \\
[\bm{\mu}_{t+1}^-]_i& = \frac{C[\bm{\mu}_t^-]_i\exp(\eta C[\check{\bm{g}}_{t}]_i)}{\sum_{i=1}^{M}\left([\bm{\mu}_t^+]_i\exp(-\eta C[\check{\bm{g}}_{t}]_i)+[\bm{\mu}_t^-]_i\exp(\eta C[\check{\bm{g}}_{t}]_i)\right)} .
\end{align*} 
\EndFor
\State {\bf return} $(\bm{\mu}^+_{t+1},\bm{\mu}^-_{t+1})$.
\end{algorithmic}
\end{algorithm}

It remains to choose $G$ as the upper bound of $\|\check{\bm{g}}_t\|_\infty$. To this end, we set 
\[
G := 2\max\{\bar p ,1\}(N+1)L_B\|\bm{A}\|_\infty+\bar{\gamma}.
\]
 Since for all $i\in[N]$ we have $\|\check{\bm{B}}_t^\top \bm{e}_i\|_1\leq(N+1) \|\check{\bm{B}}_t^\top \bm{e}_i\|_2 $ and $\|\check{\bm{B}}_t^\top \bm{e}_i\|_2\leq 2L_B$, which is guaranteed in Line~\ref{line:alg-main-check-B}, it is easy to obtain $\|\check{\mathbfcal{B}}_t\|_{\infty}=\max_i\|\check{\bm{B}}_t^\top \bm{e}_i\|_1 \leq 2(N+1)L_B$. And thus we could have the following upper bound of $\|\check{\bm{D}}_{t}(\bm{p}_t)\|_{\infty}$
\begin{align}\label{eq:boundcheckD}
    \|\check{\bm{D}}_{t}(\bm{p}_t)\|_{\infty} = \|\check{\mathbfcal{B}}_t\bm{\p}_t\|_{\infty} \leq \|\check{\mathbfcal{B}}_t\|_{\infty}\max\{\bar p ,1\} \leq 2(N+1)L_B\max\{\bar p ,1\}.
\end{align}
Therefore, we may upper bound $\|\check{\bm{g}}_{t}\|_{\infty}$ by $G$:
\begin{align}\label{eq:bound_checkg}
    \|\check{\bm{g}}_{t}\|_{\infty}=\|\bm{A}\check{\bm{D}}_{t}(\bm{p}_t)-\bm{s}_t\|_{\infty}\leq\|\bm{A}\|_{\infty}\|\check{\bm{D}}_{t}(\bm{p}_t)\|_{\infty}+\bar{\gamma}\leq 2\max\{\bar p ,1\}(N+1)L_B\|\bm{A}\|_\infty+\bar{\gamma} = G.
\end{align}

By directly applying Theorem 2 in \cite{hoeven2018many}, we have the following lemma showing that $\EG^\pm$ satisfies our requirement of the mirror descent solver.
\begin{lemma}
By adopting $\EG^\pm$ as our mirror descent solver $\varsigma^{\textrm{D}}$,  Definition~\ref{def:dualdescent} is satisfied with 
\begin{align}
C_1=\ln(2M) \qquad \qquad \text{and} \qquad \qquad C_2=C^2G^2/2. \label{eq:C1-C2-values}
\end{align}
\end{lemma}
 {It is worth noting that when $h(\bm{x})=\frac{1}{2}\|\bm{x}\|_2^2$,  the online mirror descent algorithm \eqref{eq:OMD-update} is known as the projected gradient descent method (PGD).
Since $h(x)=\frac{1}{2}\|\bm{x}\|_2^2$ is $1/M$-strongly convex over $\mathcal D=\{  \bm{\mu} \in \mathbb{R}^{M} \mid \|\bm{\mu}\|_1\leq C \}$ with respect to $\|\cdot\|_1$, the PGD solver also satisfies Definition~\ref{def:dualdescent}. 
However, the updating step of the PGD solver has no closed form due to the projection onto the $L_1$ ball in each update. Numerical results in Section~\ref{sec:addexp} demonstrate that $\EG^\pm$ has better empirical performance than the PGD solver.}

\section{Main Result}\label{sec:mainresults}

With the main technical tools ready in hand, we now prove the following main theorem which upper bounds the regret of our Algorithm~\ref{algucb}.
\begin{theorem}\label{thm:main}
When combining Algorithm~\ref{algucb} with our $\EG^\pm$ mirror descent solver (Algorithm~\ref{alg:eg}), we may upper bound the regret of the algorithm by
\begin{align*}
\mathcal R(T) & \leq \left(\|\bm{A}\|_{\infty}\bar d/\underline{\gamma} + \mathcal{O}(1)\right)[r(\bm{p}^*)+\phi(\bm{A}\bm{\bm{D}}(\bm{p}^*))]+ \mathcal{O}(\|\bm{A}\|_{\infty}\bar d\sqrt{T\log (MT)})\\
&\qquad  +2\sqrt{C_1 C_2 T}+ \O\left(\sqrt{N+1}\kappa\max\{\bar p,1\}(N\bar p+\max\{C\|\bm{A}\|_\infty, 1\})
 \right)\times\sqrt{NT\log(N+1+\bar p^2T)} ,
\end{align*}
where we may choose values for $C_1$ and $C_2$ according to Eq.~\eqref{eq:C1-C2-values} and only universal constants are hidden in the $\mathcal{O}(\cdot)$ notations.
\end{theorem}

\begin{remark}
Recall that $C= L+((\bar r+\bar \phi)/\underline{\gamma})$, $\bar r=N\bar p\bar d$, and $\kappa$ is defined in Eq.~\eqref{eq:kappa}. Assuming the problem parameters $\bar{d}, \bar{p}, \bar\phi, L, L_B,\|\bm{A}\|_\infty \leq \O(1)$ and $\underline{\gamma} \geq \Omega(1)$, we have that $C_1 \leq \wo(1)$, $C, C_2 \leq \O(N)$,   $\kappa \leq \wo(N)$, and $\mathcal{R}(T) \leq \wo(N^{5/2} \sqrt{T})$.
\end{remark}
The proof of our main theorem, which is presented in Section~\ref{sec:regretanalysis} in E-Companion, follows the general framework of the primal-dual analysis of online optimization problems (e.g., \cite{beck2003mirror,hazan2016introduction,balseiro2022best,balseiro2020regularized}), and will be detailed in 5 steps. The main differences from \cite{balseiro2020regularized} is that in Step II, we need to deal with the estimation error in the dual expression that relates to the balancing  regularizer (note the $\check{\bm{g}}_{t}$ term in $\sum_{t=1}^\tau\left\langle\bm{\mu}_{t}, \check{\bm{g}}_{t}\right\rangle+\sum_{t=1}^\tau\phi(\bm{s}_t)-T\phi\left(\frac{1}{T}\sum_{t=1}^T  \bm{A}\bm D(\bm{p}_t)\right)$). We bound this part in Steps III and IV. This error can be upper bounded by estimation error of $\check{\bm{g}}_{t}$ multiplied by the $\ell_1$-norm of the dual variables. Our definition of the dual space $\D = \left\{  \bm{\mu} \in \mathbb{R}^{M} \mid \|\bm{\mu}\|_1\leq C \right\}$ again kicks in to help upper bound the error.

\section{Numerical Experiments}\label{sec:experiments}
In this section, we present the numerical experiments on the synthetic data sets to illustrate the effectiveness of our algorithm. We use an NRM example, in which the retailer sells five products ($N=5$) using ten resources ($M=10$), and the (transpose of the) resource consumption matrix is defined as
\setlength{\arraycolsep}{8pt}
\renewcommand{\arraystretch}{1.5}
\begin{align}\label{eq:resource matrix}
    \bm{A}^\top =
 \begin{bmatrix}
1 & 1 & 0 & 2 & 1 & 1 & 0 & 2 & 2 & 3 \\
3 & 1 & 2 & 1 & 3 & 1 & 2 & 1 & 1 & 0 \\
2 & 3 & 1 & 0 & 2 & 3 & 1 & 0 & 0 & 3 \\
1 & 0 & 0 & 0 & 1 & 0 & 0 & 0 & 0 & 0 \\
0 & 1 & 1 & 2 & 0 & 1 & 1 & 2 & 2 & 3 \\
\end{bmatrix}.
\end{align}

The underlying linear demand function is defined as \[
\bm{D}(\bm{p}) =  \begin{bmatrix}
20 \\
25 \\
19 \\
14 \\
23 \\
\end{bmatrix} +  \begin{bmatrix}
-5 & 0.10 & 0.09 & 0.1 & 0.11 \\
0.11 & -7 & 0.10 & 0.02 & 0.12 \\
0.03 & 0.1 & -3.5 & 0.18 & 0.07 \\
0.10 & 0.02 & 0 & -2.5 & 0 \\
0.04 & 0.05 & 0.10 & 0 & -6 \\
\end{bmatrix} \bm{p}.
\]

In addition, we choose the weighted min-max fairness regularizer 
\[
\phi(\bm{s}):=\lambda\min_i(w_is_i)
\]
with $w_i = 1$ for all $i$. 
We  generate the demand noise from the truncated Gaussian distribution 
\[
\mathrm{clip}(\mathcal{N}(0,1),1), \text{~~~where~}
\mathrm{clip}(x,1) = \left\{ \begin{array}{rr}
-1 & x<-1;\\
x & |x|\leq 1;\\
1 &x> 1
\end{array} \right\}.
\]
We set the time horizon as $T\in\{100,500,1000,2000,3000,4000,5000,6000,7000,8000,9000,10000\}$ and the price range for each product as $[1,4]$. We
 test two initial inventory levels ($\bm{\gamma}_1=(60, 50, 45, 40, 55, 45, 60, 40, 40, 80)^\top$ and $\bm{\gamma}_2 = (80, 70, 65, 60, 75, 65, 70, 60, 60, 100)^\top$) and four regularization level ($\lambda \in \{0, 0.5, 1.0, 1.5\}$) using two mirror descent solvers ($\EG^\pm$ and PGD). 
We conduct $10$ trials independently for each case and plot the average result of these trials in all figures. We also use the shaded region around each curve to indicate the $95\%$ confidence interval across the $10$ trials.\footnote{ {For normal distribution, the $Z$- value for $95\% $ confidence is $1.96$. Slightly abusing the terminology, we define the $95\%$ confidence interval here as  $[\textrm{mean}+1.96*\textrm{standard error}, \textrm{mean}+1.96*\textrm{standard error}]$}.}

For brevity, we present the numerical results of $\bm{\gamma}_1=(60, 50, 45, 40, 55, 45, 60, 40, 40, 80)^\top$  in this section and leave the numerical results of initial inventory level $\bm{\gamma}_2 = (80, 70, 65, 60, 75, 65, 70, 60, 60, 100)^\top$  to Section~\ref{sec:addexp} in the supplementary materials.

\begin{figure}[t]
\centering
\includegraphics[width =0.48\textwidth]{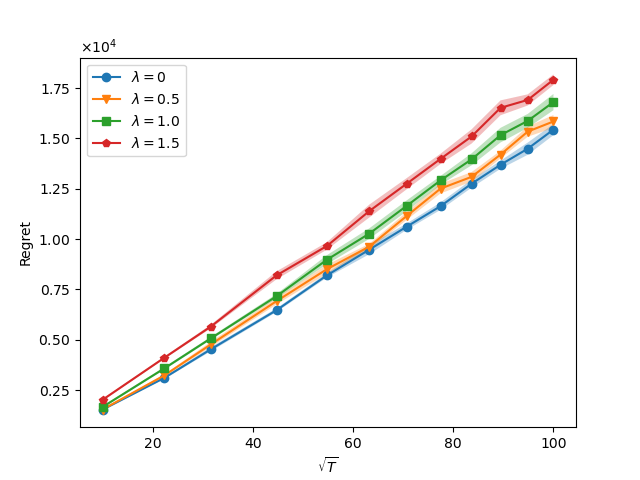}
\includegraphics[width =0.48\textwidth]{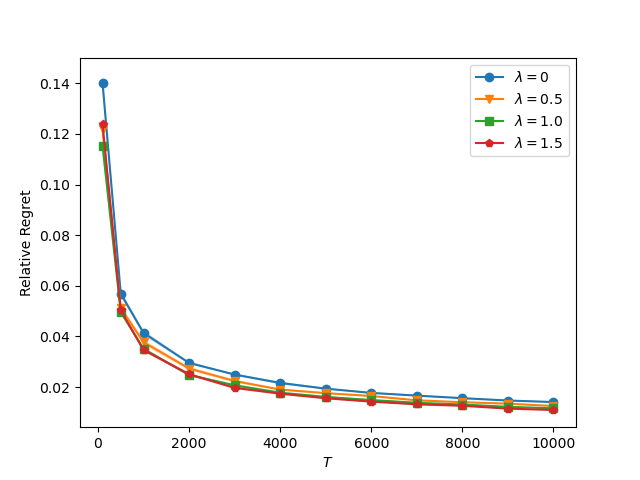}
\caption{ {The performance of Algorithm~\ref{algucb}+$\EG^\pm$ with $\bm{\gamma}_1$ and $\lambda \in\{ 0, 0.5,1.0, 1.5\}$.  Here the $x$-axis of the left figure is the square root of the total time periods $T$ and the $y$-axis is the cumulative regret defined in Eq.~(\ref{regretform2}). The $x$-axis of the right figure is the total time periods $T$ and the $y$-axis is the relative regret defined in \eq{\ref{eq:relativeregret}}.}}
\label{fig:linearegregretlow}
\includegraphics[width =0.48\textwidth]{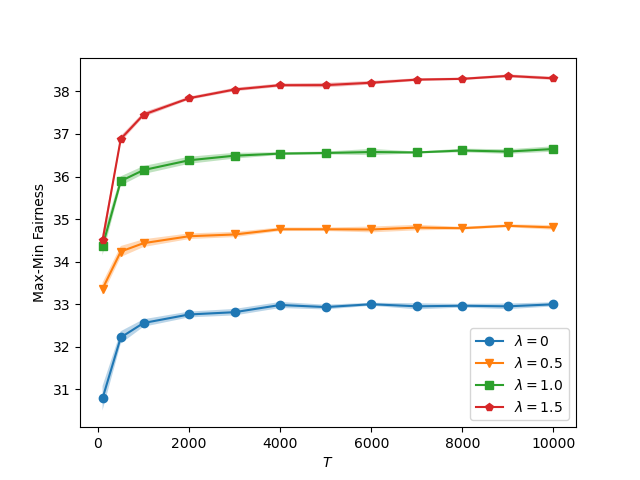}
\includegraphics[width =0.48\textwidth]{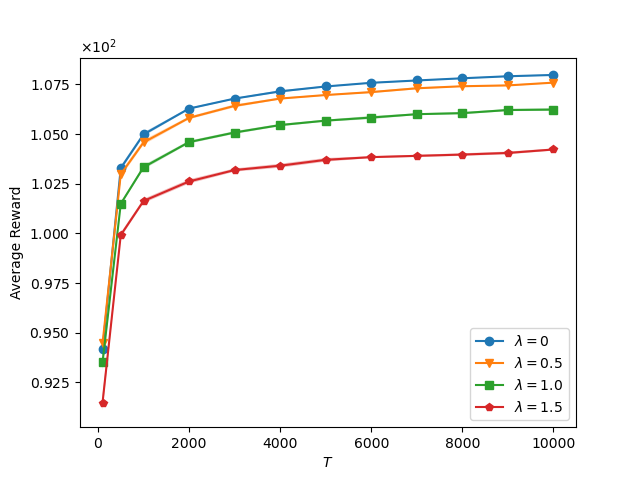}
\caption{ {The max-min fairness $\min_i\left(\frac{1}{T}\sum_{t=1}^T[\bm{A}\bm{d}_t]_i\right)$ and the average reward $\frac{1}{T}\sum_{t=1}^Tr(\bm{p}_t)$ of Algorithm~\ref{algucb}+$\EG^\pm$ at regularization levels $\lambda \in\{ 0, 0.5,1.0, 1.5\}$ under the initial inventory level $\bm{\gamma}_1$. } }
\label{fig:linearegfairnesslow}
\end{figure}

\noindent \underline{\bf Numerical results of Algorithm~\ref{algucb}+ $\EG^\pm$  under inventory level $\bm{\gamma}_1$.}
In the left of Figure~\ref{fig:linearegregretlow} is the plot of the regret of Algorithm~\ref{algucb} with the inventory level $\bm{\gamma}_1$ and $\lambda \in\{ 0, 0.5,1.0, 1.5\}$ versus the square root of the total time periods $T$. This figure clearly demonstrates the regret of our algorithm grows at rate $\wo(\sqrt{T})$ for all regularization levels $\lambda \in\{ 0, 0.5,1.0, 1.5\}$, which is consistent with the theoretical guarantee of Theorem~\ref{thm:main}. In the right of Figure~\ref{fig:linearegregretlow} we plot the relative regret of Algorithm~\ref{algucb} versus the total time periods $T$, where the relative regret is defined as 
\begin{equation}\label{eq:relativeregret}
    \frac{T[r(\bm{p}^*)+\phi(\bm{A}\bm{\bm{D}}(\bm{p}^*))]-\E\left[\sum_{t=1}^Tr(\bm{p}_t)+T\phi\left(\frac{1}{T}\sum_{t=1}^T\bm{A}\bm{d}_t\right)\right]}{T[r(\bm{p}^*)+\phi(\bm{A}\bm{\bm{D}}(\bm{p}^*))]}.
\end{equation}
Note that the narrow $95\%$ confidence intervals indicate the stability and robustness of our algorithm.

We further empirically study the impact of the regularization level $\lambda$ to the utilization of the resources. In the left of Figure~\ref{fig:linearegfairnesslow} is the plot of the max-min fairness versus the total time periods $T$ with the inventory level $\bm{\gamma}_1$ and $\lambda \in\{0, 0.5,1.0, 1.5\}$, where the max-min fairness is defined as the minimum element of the average resource consumption vector $\min_i\left(\frac{1}{T}\sum_{t=1}^T[\bm{A}\bm{d}_t]_i\right)$.
In the right of Figure~\ref{fig:linearegfairnesslow} we plot the average reward versus the total time periods $T$, where the average reward is defined as $\frac{1}{T}\sum_{t=1}^Tr(\bm{p}_t)$. 
These figures show the max-min fairness increases and the average reward decreases as $\lambda$ grows, indicating the natural trade-off between fairness and the average reward. We also find that the max-min fairness could be enhanced greatly with a small sacrifice of the average reward reduction. 

In Section~\ref{sec:addexp} of the supplementary materials, we present  additional numerical results, which include 1) computational time report; 2) numerical results of Algorithm~\ref{algucb}+$\EG^\pm$ under the inventory level $\bm{\gamma}_2$; 3) numerical results of Algorithm~\ref{algucb}+$\mathrm{PGD}$; 4) performance Comparison of $\EG^\pm$ and $\mathrm{PGD}$; 5) numerical results of Algorithm~\ref{algucb}+$\EG^\pm$ in a model misspecified setting;  6) numerical results on a classic NRM example studied in \cite{besbes2012blind,ferreira2018online}.

\section{Conclusion}\label{sec:conclusion}
This paper studies the price-based network revenue management with both fair resource-consumption balancing and demand learning, which is motivated by the practical needs of industries such as online retailing and airline applications. To tackle the challenges of this task, we make several innovative technical contributions, which have the potential to be applied to other operations management problems.  We propose a primal-dual-type online policy with Upper-Confidence-Bound (UCB) learning method to simultaneously learn the unknown demand function and optimize the composite objective concerning both the NRM revenue and the balancing metric.  Both theoretical analysis and numerical results show the effectiveness and the ability {in simultaneously achieving revenue maximization and fair resource-consumption balancing.

For future directions, one can consider adapting the framework in this paper to other revenue management applications with both resource-consumption balancing and demand learning. One could also study the demand learning of non-parametric demand functions and consider adapting our framework to other global ancillary objectives beyond balanced consumption across resources.
As discussed in the introduction section, in some scenarios resource-consumption balancing could enhance the customer satisfactions.
Addressing customer dissatisfaction through the inclusion of customer dissatisfaction costs into revenue maximization objective is also a noteworthy consideration.
 While modeling and formulating customer dissatisfaction costs comes with its own set of complexities and involves intricate challenges, we believe it's a promising avenue to explore for future research.

\bibliographystyle{pomsref}

\let\oldbibliography\thebibliography
\renewcommand{\thebibliography}[1]{%
	\oldbibliography{#1}%
	\baselineskip14pt 
	\setlength{\itemsep}{10pt}
}
\bibliography{main.bib}

\ECSwitch



\begin{center}
 		\large{\bf E-Companion to `` Network Revenue Management with Demand Learning and Fair Resource-Consumption Balancing''} 
 	\end{center}
 	
 	\vspace{10pt}

\section{Some useful technical lemmas}
\begin{lemma}[Azuma-Hoeffding Inequality]\cite{}\label{le:azuma}
 Let $\left(\left\{\left(D_{k}, \mathcal{F}_{k}\right)\right\}_{k=1}^{\infty}\right)$ be a martingale difference sequence for which there are constants $\left\{\left(a_{k}, b_{k}\right)\right\}_{k=1}^{n}$ such that $D_{k} \in\left[a_{k}, b_{k}\right]$ almost surely for all $k=1, \ldots, n$. Then, for all $t \geq 0$,
$$
\mathbb{P}\left[\left|\sum_{k=1}^{n} D_{k}\right| \geq t\right] \leq 2 \mathrm{exp}\left(-\frac{2 t^{2}}{\sum_{k=1}^{n}\left(b_{k}-a_{k}\right)^{2}}\right).
$$
\end{lemma}

Recall that $\bm{\mathbfcal{B}}=[\bm{B}|\bm{\alpha}]\in\R^{N\times(N+1)}$ $\hat{\mathbfcal{B}}=[\hat{\bm{B}}|\hat{\bm{\alpha}}]\in\R^{N\times(N+1)}$, and  $\p := (\bm p, 1)$. Thus we have $\bm d_{t}^\top \bm{e}_i=(\mathbfcal{B}^\top \bm e_{i})^\top\p_t+\bm{\varepsilon}_i$. Noting $\left\|\mathbfcal{B}^\top \bm e_{i}\right\|_{2} \leq L_B$, $\varepsilon_i\leq\bar d$ and $\|\widetilde{\bm p}_t\|_2\leq\bar p\sqrt{N+1}$ and applying the Theorem 2 in \cite{abbasi2011improved}, we have the following confidence ellipsoid lemma.
\begin{lemma}\label{le:ellipsoid}
Recall that $\bm{\Lambda}_t=(N+1)\cdot\mathbf{I}_{N+1}+\sum_{s<t}\widetilde{\bm p}_s\widetilde{\bm p}_s^\top$, for any $\delta>0$, with probability at least $1-\delta$, for all $t$ we have
\begin{equation}
    \P\left[\|(\mathbfcal{B}-\hat{\mathbfcal{B}}_{t})^\top \bm{e}_i \|_{\bm{\Lambda}_t} \leq \bar d \sqrt{(N+1)\ln \left(\frac{1+t\bar p^2}{\delta}\right)}+\sqrt{N+1}L_B\right]\geq 1-\delta.
\end{equation}

\end{lemma}

\section{Proofs Omitted in Section~\ref{sec:model}}
\subsection{Proof of Proposition~\ref{prop1}}
\proof{Proof of Proposition~\ref{prop1}.}
By the constraint of Eq.~(\ref{eq:dp}) , it is easy to obtain $\E\left[\sum_{t=1}^T\bm{A}\bm{d}_t\right]  \leq T\bm{\gamma}$. Therefore,
\begin{equation}\nonumber
\begin{aligned}
J_{\mathrm{opt}}&\leq \left\{
\begin{array}{cl}
\max\limits_{\pi} \ &\ \E\left[\sum_{t=1}^Tr(\bm{p}_t)+T\phi\left(\frac{1}{T}\sum_{t=1}^T\bm{A}\bm{d}_t\right)\right] \\
    \text{s.t.} \ &\ \E\left[\sum_{t=1}^T\bm{A}\bm D(\bm{p}_t)\right]  \leq T\bm{\gamma}
\end{array}
    \right\}.
\end{aligned}
\end{equation}
With the concavity of $r(\bm{p}_t)$ by Jenson's inequality we obtain $$\sum_{t=1}^Tr(\bm{p}_t)\leq Tr\left(\frac{1}{T}\sum_{t=1}^T\bm{p}_t\right). $$
Therefore, we have
\begin{equation}\nonumber
\begin{aligned}
&\left\{
\begin{array}{cl}
\max\limits_{\pi} \ &\ \E\left[\sum_{t=1}^Tr(\bm{p}_t)+T\phi\left(\frac{1}{T}\sum_{t=1}^T\bm{A}\bm{d}_t\right)\right] \\
    \text{s.t.} \ &\ \E\left[\sum_{t=1}^T\bm{A}\bm D(\bm{p}_t)\right]  \leq T\bm{\gamma}
\end{array}
    \right\}
    \\&\leq\left\{
\begin{array}{cl}
\max\limits_{\pi} \ &\ \E\left[Tr\left(\frac{1}{T}\sum_{t=1}^T\bm{p}_t\right)+T\phi\left(\frac{1}{T}\sum_{t=1}^T\bm{A}\bm{d}_t\right)\right] \\
    \text{s.t.} \ &\ \E\left[\sum_{t=1}^T\bm{A}\bm D(\bm{p}_t)\right]  \leq T\bm{\gamma}
\end{array}
    \right\}.
\end{aligned}
\end{equation}
With the concavity of $r(\cdot)$ and $\phi(\cdot)$, using Jenson's inequality again, we have 
\begin{equation}\nonumber
\begin{aligned}
&\left\{
\begin{array}{cl}
\max\limits_{\pi} \ &\ \E\left[Tr\left(\frac{1}{T}\sum_{t=1}^T\bm{p}_t\right)+T\phi\left(\frac{1}{T}\sum_{t=1}^T\bm{A}\bm{d}_t\right)\right] \\
    \text{s.t.} \ &\ \E\left[\sum_{t=1}^T\bm{A}\bm D(\bm{p}_t)\right]  \leq T\bm{\gamma}
\end{array}
    \right\}\\&\leq\left\{
\begin{array}{cl}
\max\limits_{\pi} \ &\ Tr\left(\E\left[\frac{1}{T}\sum_{t=1}^T\bm{p}_t\right]\right)+T\phi\left(\E\left[\frac{1}{T}\sum_{t=1}^T\bm{A}\bm D(\bm{p}_t)\right]\right) \\
    \text{s.t.} \ &\ \E\left[\sum_{t=1}^T\bm{A}\bm D(\bm{p}_t)\right]  \leq T\bm{\gamma}
\end{array}
    \right\}\\&\leq J_D,
\end{aligned}
\end{equation}
where the last inequality is due to \eq{\ref{eq:fluid3}}.
Combining the above inequalities, we complete the proof.\Halmos
\endproof
\subsection{Assumptions Validation of the Balancing Regularizers}\label{sec:assumptions validations}
We will prove the balancing regularizers proposed in Section~\ref{sec:regularizer} satisfy Assumption~\ref{assump:phi1} (we only validate Assumption~1.1 and Assumption~1.3, since Assumption~1.2 can be easily validated).

\noindent \textit{\underline{Example 1: Weighted Max-min Fairness Regularizer:  $\phi(\bm{s}):=\lambda\min_i(w_is_i)$.}}  

We first show $\phi(\cdot)$ is $L$-Lipschitz continuous with respect to the $\|\cdot\|_{\infty}$-norm in the following way,
\begin{align*}
    \phi(\bm{s})-\phi(\bm t)&= \lambda(\min_i(w_is_i)-\min_i(w_it_i))\\&\leq \lambda \max_i(|w_is_i-w_it_i|) \\&\leq(\lambda\max_iw_i)\|\bm s-\bm t\|_\infty.
\end{align*}
Next, we will show the concavity of $\phi(\cdot)$. For any $s,t$ and $\bm{\alpha}\in[0,1]$, we have 
\begin{align*}
    \phi(\alpha \bm s+(1-\alpha)\bm t)&= \lambda(\min_i(\alpha w_is_i+(1-\alpha)w_it_i)\\&\geq \lambda(\alpha\min_i(w_is_i)+(1-\alpha)\min_i(w_it_i)) \\&\geq\alpha\phi(\bm{s})+(1-\alpha)\phi(\bm t).
\end{align*}
\noindent \textit{\underline{Example 2: Group Max-min Fairness Regularizer: $\phi(\bm{s}):=\lambda\min_i( (\bm{U}\tilde{\bm{s}})_i)$, where $\tilde{\bm{s}}=(w_1s_1,\cdots,w_ms_m)^\top$.}}

We first show $\phi(\cdot)$ is $L$-Lipschitz continuous with respect to the $\|\cdot\|_{\infty}$-norm in the following way,
\begin{align*}
    \phi(\bm{s})-\phi(\bm t)&= \lambda(\min_i((\bm{U}\tilde{\bm{s}})_i)-\min_i((\bm{U}\tilde{\bm{t}})_i))\\&\leq \lambda \|\bm{U}(\tilde{\bm{s}}-\tilde{\bm{t}})\|_\infty\\&\leq \lambda\|\bm{U}\|_\infty\|\tilde{\bm{s}}-\tilde{\bm{t}}\|_\infty\\&\le(\lambda\|\bm{U}\|_\infty\max_iw_i)\|\bm s-\bm t\|_\infty.
\end{align*}
Next, we will show the concavity of $\phi(\cdot)$. For any $\bm s,\bm t$ and $\alpha\in[0,1]$, we have 
\begin{align*}
    \phi(\alpha \bm s+(1-\alpha)\bm t)&= \lambda(\min_i(\alpha (\bm{U}\tilde{\bm{s}})_i+(1-\alpha)(\bm{U}\tilde{\bm{t}})_i)\\&\geq \lambda(\alpha\min_i((\bm{U}\tilde{\bm{s}})_i)+(1-\alpha)\min_i((\bm{U}\tilde{\bm{t}})_i)) \\&\geq\alpha\phi(\bm{s})+(1-\alpha)\phi(\bm t).
\end{align*}

\noindent \textit{\underline{Example 3: Range Fairness Regularizer:$\phi(\bm{s}):=\lambda(\min_i(w_is_i)-\max_i(w_is_i)+\max_i(w_i\gamma_i))$.}} 
\noindent \textit{\underline{Example 4: Load Balancing Regularizer:$\phi(\bm{s}):=\lambda(\min_i((\gamma_i-s_i)/\gamma_i)$.}}

We have shown 
$\min_i(w_is_i)$ is $L$-Lipschitz continuous with respect to the $\|\cdot\|_{\infty}$-norm and concave.  By this fact, it is easy to note that  Example 3 and Example 4 also satisfy Assumption~\ref{assump:phi1}.
\Halmos

\section{Proofs Omitted in Section~\ref{sec:alg}}

\subsection{Proof of Lemma~\ref{le:ucb0}}
\proof{Proof of Lemma~\ref{le:ucb0}.}
Let $\delta$ in Lemma~\ref{le:ellipsoid} be $1/(NT)$, and thus with probability $1-1/T$ for any $i\in[N]$ and $t\le\tau$ it holds that
\begin{align}\nonumber
    \|(\mathbfcal{B}-\hat{\mathbfcal{B}}_{t})^\top \bm{e}_i \|^2_{\bm{\Lambda}_t}&=\|\bm{\Lambda}_t^{1/2}((\mathbfcal{B}-\hat{\mathbfcal{B}}_{t})^\top \bm{e}_i )\|^2_2\\&\leq 2\bar d^2(N+1)\ln\left(NT(1+\bar p^2T)\right)+2(N+1)L_B^2,\label{eq:leucb00}
\end{align}
where the inequality is due to $(a+b)^2\leq 2a^2+2b^2$.

By \eq{\ref{eq:leucb00}} and Cauchy-Schwarz inequality, we have
\begin{align}
     \nonumber \|(\hat{\mathbfcal{B}}_t -\mathbfcal{B})\p\|_\infty&\leq\max_{i\in[N]}|((\mathbfcal{B}-\hat{\mathbfcal{B}}_{t})^\top \bm{e}_i  )^\top\p|
    \\&\leq \max_{i\in[N]}\nonumber |((\mathbfcal{B}-\hat{\mathbfcal{B}}_{t})^\top \bm{e}_i  )^\top\bm{\Lambda}_t^{1/2}\bm{\Lambda}_t^{-1/2}\p|
    \\\nonumber&\leq\max_{i\in[N]}\|\bm{\Lambda}_t^{1/2}((\mathbfcal{B}-\hat{\mathbfcal{B}}_{t})^\top \bm{e}_i  )\|_2\|\bm{\Lambda}_t^{-1/2}\p\|_2
    \\&\leq\sqrt{2\bar d^2(N+1)\ln\left(NT(1+\bar p^2T)\right)+2(N+1)L_B^2}\sqrt{\p^\top\bm{\Lambda}_t^{-1}\p}.\label{eq:leucb2.9}
\end{align}
Noting that $\|(\hat{\mathbfcal{B}}_t -\mathbfcal{B})\p\|_2\le\sqrt{N}\|(\hat{\mathbfcal{B}}_t -\mathbfcal{B})\p\|_\infty$, by \eq{\ref{eq:leucb2.9}} we have 
\begin{align}\nonumber
    \|(\hat{\mathbfcal{B}}_t -\mathbfcal{B})\p\|_2&\le\sqrt{N}\|(\hat{\mathbfcal{B}}_t -\mathbfcal{B})\p\|_\infty
    \\\label{eq:leucb2}&\leq\sqrt{2\bar d^2N(N+1)\ln\left(NT(1+\bar p^2T)\right)+2N(N+1)L_B^2}\sqrt{\p^\top\bm{\Lambda}_t^{-1}\p}.
\end{align}

\noindent\textit{\underline{Proof of $\left| \hat{r}_{t}(\bm{p})-r(\bm{p})\right| \leq \Delta_{t}^{r}(\bm{p})$.}}

By \eq{\ref{eq:leucb2}}, we have
\begin{align}\nonumber
    \left|\langle \bm{p},\hat{\mathbfcal{B}}_t \p\rangle-\Braket{\bm{p},\mathbfcal{B}\p}\right|&\leq\|\bm{p}\|_2\|(\hat{\mathbfcal{B}}_t -\mathbfcal{B})\p\|_2\\&\leq \sqrt{N}\bar p\sqrt{2\bar d^2N(N+1)\ln\left(NT(1+\bar p^2T)\right)+2N(N+1)L_B^2}\sqrt{\p^\top\bm{\Lambda}_t^{-1}\p},\label{eq:leucb2.5}
\end{align}
where the first inequality is due to Cauchy-Schwarz inequality.

Since for any vector $\bm x\in\R^d$ it holds that
$$\|\bm{x}\|_2\leq\sqrt{d}\|\bm{x}\|_\infty\leq\sqrt{d}\|\bm{x}\|_2.$$ 
Therefore, we have
\begin{align}\label{eq:leucb3}
   \sqrt{\p^\top\bm{\Lambda}_t^{-1}\p}\leq\sqrt{N+1}\|\bm{\Lambda}_t^{-1/2}\p_t\|_\infty\leq\sqrt{N+1}\sqrt{\p^\top\bm{\Lambda}_t^{-1}\p}.
\end{align}
Combining \eq{\ref{eq:leucb2.5}} and \eq{\ref{eq:leucb3}}, we have 
\begin{align}\nonumber
 \left| \hat{r}_{t}(\bm{p})-r(\bm{p})\right|&=\left|\langle \bm{p},\hat{\mathbfcal{B}}_t \p\rangle-\Braket{\bm{p},\mathbfcal{B}\p}\right|\\&\leq \sqrt{N}\bar p\sqrt{2\bar d^2N(N+1)\ln\left(NT(1+\bar p^2T)\right)+2N(N+1)L_B^2}\sqrt{\p^\top\bm{\Lambda}_t^{-1}\p}\nonumber
 \\&\leq N(N+1)\bar p \sqrt{ 2\bar d^2\ln\left(NT(1+\bar p^2T)\right)+2L_B^2}\|\bm{\Lambda}_t^{-1/2}\p\|_\infty\nonumber
 \\&\leq\Delta_t^r(\bm{p}).\label{eq:deltar}
\end{align}

\noindent\textit{\underline{Proof of $\| \hat{\bm{D}}_{t}(\bm{p})-\bm{D}(\bm{p})\|_{\infty} \leq \Delta_{t}^{\bm{D}}(\bm{p})$.}}

Combining \eq{\ref{eq:leucb2.9}} and \eq{\ref{eq:leucb3}}, it is easy to obtain that
\begin{align}\nonumber
    \| \hat{\bm{D}}_{t}(\bm{p})-\bm{D}(\bm{p})\|_{\infty}&\nonumber=\|(\hat{\mathbfcal{B}}_t -\mathbfcal{B})\p\|_\infty\\&\leq \sqrt{2\bar d^2(N+1)\ln\left(NT(1+\bar p^2T)\right)+2(N+1)L_B^2}\sqrt{\p^\top\bm{\Lambda}_t^{-1}\p}\nonumber
    \\&\leq(N+1)\sqrt{2\bar d^2\ln\left(NT(1+\bar p^2T)\right)+2L_B^2}\|\bm{\Lambda}_t^{-1/2}\p\|_\infty\nonumber\\&\leq\Delta_t^{\bm{D}}(\bm{p}).\label{eq:deltad}
\end{align}

\noindent\textit{\underline{Proof of $| \hat{f}_t(\bm{p}) - f_t(\bm{p})| \leq \Delta_t^f(\bm{p})$.}}

By Eq.~(\ref{eq:deltad}), with probability $1-1/T$ we have 
\begin{align}\nonumber
    \left|\left\langle\bm{\mu}_t, \bm A \hat{\bm{D}}_{t}(\bm{p})-\bm{A}\bm{\bm{D}}(\bm{p})\right\rangle\right| &\leq \|\bm{\mu}_t\|_1\cdot\| \bm{A}( \hat{\bm{D}}_{t}(\bm{p})-\bm{D}(\bm{p}))\|_\infty\\&\leq\|\bm{\mu}_t\|_1\cdot\|\bm{A}\|_\infty\Delta_t^{\bm{D}}(\bm{p}).\label{eq:adeltad}
\end{align}
Since $f_{t}(\bm{p})=r(\bm{p})-\langle\bm{\mu}_t, \bm{A}\bm{\bm{D}}(\bm{p})\rangle$, $ \hat{f}_{t}(\bm{p})= \hat{r}_{t}(\bm{p})-\left\langle\bm{\mu}_t, \bm A \hat{\bm{D}}_{t}(\bm{p})\right\rangle $ and $\Delta_t^f(\bm{p})=\Delta_t^r(\bm{p})+\|\bm{\mu}_t\|_1\cdot\|\bm{A}\|_\infty\Delta_t^{\bm{D}}(\bm{p})$.
Combining \eq{\ref{eq:deltar}} and \eq{\ref{eq:adeltad}}, we have 
\begin{align*}
    | \hat{f}_t(\bm{p}) - f_t(\bm{p})|&\leq \left| \hat{r}_{t}(\bm{p})-r_{t}(\bm{p})\right|+\left|\left\langle\bm{\mu}_t, \bm A \hat{\bm{D}}_{t}(\bm{p})-\bm{A}\bm{\bm{D}}(\bm{p})\right\rangle\right|\\&\leq \Delta_t^r(\bm{p})+ \|\bm{\mu}_t\|_1\cdot\|\bm{A}\|_\infty\Delta_t^{\bm{D}}(\bm{p})
    \\& = \Delta_t^f(\bm{p}).
\end{align*}
Therefore, we complete the proof of Lemma~\ref{le:ucb0}.
\Halmos
\endproof
\subsection{Proof of Corollary~\ref{cor:check-error-bound}}

\proof{Proof of Corollary~\ref{cor:check-error-bound}.}
The proof will be conditioned on when we find a feasible $(\check{\bm{\alpha}}_t,\check{\bm{B}}_t) \in \mathcal{M}_t$ for all $t \leq \tau$ (which happens with probability $(1-\mathcal O\left(T^{-1}\right))$ by Lemma~\ref{le:find}) and the desired event of Lemma~\ref{le:ucb0}.

For each $t \leq \tau$ and $\bm p \in [\underline{p}, \bar{p}]^N$, we first upper bound $\|\check{\bm{D}}_{t}(\bm{p})-\bm{D}(\bm{p})\|_{\infty}$. Note that
\begin{align*}
\|\check{\bm{D}}_{t}(\bm{p})-\bm{D}(\bm{p})\|_{\infty} \leq \|\check{\bm{D}}_{t}(\bm{p})-\hat{\bm{D}}(\bm{p})\|_{\infty} + \|\hat{\bm{D}}_{t}(\bm{p})-\bm{D}(\bm{p})\|_{\infty} \leq \|\check{\bm{D}}_{t}(\bm{p})-\hat{\bm{D}}(\bm{p})\|_{\infty} + \Delta_t^{\bm{D}}(\bm{p}).
\end{align*}
Therefore, we only need to show that $\|\check{\bm{D}}_{t}(\bm{p})-\hat{\bm{D}}(\bm{p})\|_{\infty} \leq \Delta_t^{\bm{D}}(\bm{p})$ to prove that $\|\check{\bm{D}}_{t}(\bm{p})-\bm{D}(\bm{p})\|_{\infty} \leq 2\Delta_{t}^{\bm{D}}(\bm{p})$. For every $i \in [N]$, let $\check{\mathbfcal{B}}_t = [\check{\bm{B}}_t | \check{\bm{\alpha}}_t]$, and we verify that
\begin{align*}
|\bm{e}_i^\top (\check{\bm{D}}_{t}(\bm{p})-\hat{\bm{D}}(\bm{p}))|& = |\bm{e}_i^\top (\check{\mathbfcal{B}}_t - \hat{\mathbfcal{B}}_t) \p| = |\bm{e}_i^\top (\check{\mathbfcal{B}}_t - \hat{\mathbfcal{B}}_t) \bm{\Lambda}_{t}^{1/2} \bm{\Lambda}_{t}^{-1/2} \p| \nonumber\\
&\leq \| \bm{\Lambda}_{t}^{1/2} ( \check{\mathbfcal{B}}_t - \hat{\mathbfcal{B}}_t)^\top \bm{e}_i \| \cdot \|\bm{\Lambda}_{t}^{-1/2} \p\| \leq \kappa \|\bm{\Lambda}_{t}^{-1/2} \p\| \leq \sqrt{N+1}\kappa \|\bm{\Lambda}_{t}^{-1/2} \p\|_{\infty} = \Delta_t^{\bm{D}}(\bm{p}) .
\end{align*}
Here, the first inequality is due to Cauchy-Schwarz, and the second inequality is due to that $(\check{\bm{\alpha}}_t,\check{\bm{B}}_t) \in \mathcal{M}_t$.

We now upper bound $\left|\check{r}_{t}(\bm{p})-r(\bm{p})\right|$ as follows.
\begin{align*}
\left|\check{r}_{t}(\bm{p})-r(\bm{p})\right| \leq \left|\check{r}_{t}(\bm{p})-\hat{r}(\bm{p})\right| + \left|\hat{r}_{t}(\bm{p})-r(\bm{p})\right| \leq N \bar{p} \|\check{\bm{D}}_{t}(\bm{p})-\hat{\bm{D}}(\bm{p})\|_{\infty} + \left|\hat{r}_{t}(\bm{p})-r(\bm{p})\right| \leq \Delta_t^r(\bm{p}) + \Delta_t^r(\bm{p}),
\end{align*}
where in the inequality, we upper bound $\|\check{\bm{D}}_{t}(\bm{p})-\hat{\bm{D}}(\bm{p})\|_{\infty}$ by $\Delta_t^{\bm{D}}(\bm{p})$ due to the paragraph above.

We finally upper bound $|\check{f}_t(\bm{p}) - f_t(\bm{p})|$. Note that
\begin{align*}
|\check{f}_t(\bm{p}) - f_t(\bm{p})|& \leq |\check{f}_t(\bm{p}) - \hat{f}_t(\bm{p})| + |\hat{f}_t(\bm{p}) - f_t(\bm{p})|\\
&\leq \left|\check{r}_{t}(\bm{p})-\hat{r}(\bm{p})\right| + \|\bm{\mu}_t\|_1 \cdot \|\bm{A}\|_\infty \|\check{\bm{D}}_{t}(\bm{p})-\hat{\bm{D}}(\bm{p})\|_{\infty} + |\hat{f}_t(\bm{p}) - f_t(\bm{p})|\\
&\leq \Delta_t^r(\bm{p}) + \|\bm{\mu}_t\|_1 \cdot \|\bm{A}\|_\infty \Delta_t^{\bm{D}}(\bm{p}) + \Delta_t^f(\bm{p}) = 2\Delta_t^f(\bm{p}) ,
\end{align*}
where the second inequality is by the definitions of $\hat{f}_t$ and $\check{f}_t$, and the third inequality uses the upper bounds for $\left|\check{r}_{t}(\bm{p})-\hat{r}(\bm{p})\right|$ and $\|\check{\bm{D}}_{t}(\bm{p})-\hat{\bm{D}}(\bm{p})\|_{\infty}$ derived in the previous parts of this proof.  \Halmos 
\endproof

\subsection{Proof of Lemma~\ref{le:ucb00}}
\proof{Proof of Lemma~\ref{le:ucb00}.}
Since $\bm{\Lambda}_t=(N+1)\cdot\mathbf{I}_{N+1}+\sum_{s<t}\widetilde{\bm p}_s\widetilde{\bm p}_s^\top$, for every $t\geq1$, $\p_t^\top\bm{\Lambda}_t^{-1}\p_t\leq\p_t^\top\p_t/(N+1)\leq\bar p^2$. By this fact, we have
\begin{align}\nonumber
    \sqrt{\p_t^\top\bm{\Lambda}_t^{-1}\p_t}&= \min\{\bar p, \sqrt{\p_t^\top\bm{\Lambda}_t^{-1}\p_t}\}\\&\leq \max\{\bar p,1\}\min\{1,\sqrt{\p_t^\top\bm{\Lambda}_t^{-1}\p_t}\}.\label{eq:lepo2}
\end{align}
Note $\|\bm{\Lambda}_t^{-1/2}\p_t\|_\infty\leq\sqrt{\p_t^\top\bm{\Lambda}_t^{-1}\p_t}$
and recall the definition of $\Delta_{t}^{r}(\bm{p})$ and $\Delta_{t}^{\bm{D}}(\bm{p})$ we have
\begin{align}
    &\Delta_{t}^{r}(\bm{p}) \leq 2N(N+1)\bar p \sqrt{ 2\bar d^2\ln\left(NT(1+\bar p^2T)\right)+2L_B^2}\sqrt{\p_t^\top\bm{\Lambda}_t^{-1}\p_t},\label{eq:bounddeltar}\\&
    \Delta_{t}^{\bm{D}}(\bm{p})\leq 2(N+1)\sqrt{ 2\bar d^2\ln\left(NT(1+\bar p^2T)\right)+2L_B^2}\sqrt{\p_t^\top\bm{\Lambda}_t^{-1}\p_t}.\label{eq:bounddeltad}
\end{align}

Recalling the definition $\kappa=2\sqrt{2\bar d^2(N+1)\ln\left(NT(1+\bar p^2T)\right)+2(N+1)L_B^2}
$  and $\Delta_t(\bm{p}) := \Delta_t^r(\bm{p})+ \max\{C\|\bm{A}\|_\infty, 1\} \Delta_t^{\bm{D}}(\bm{p})$,  by Eqs.~(\ref{eq:lepo2}, \ref{eq:bounddeltar}, \ref{eq:bounddeltad}), we obtain
\begin{align}\nonumber
\Delta_t(\bm{p})&\nonumber\leq
\sqrt{N+1}\kappa(N\bar p+\max\{C\|\bm{A}\|_\infty, 1\})
 \sqrt{\p_t^\top\bm{\Lambda}_t^{-1}\p_t}\nonumber\\&\leq
\sqrt{N+1}\kappa\max\{\bar p,1\}(N\bar p+\max\{C\|\bm{A}\|_\infty, 1\})
 \min\left\{1,\sqrt{\p_t^\top\bm{\Lambda}_t^{-1}\p_t}\right\}.\label{eq:minrphi}
\end{align}

Since $\bm{\Lambda}_t=\bm{\Lambda}_{t-1}+\p_t\p_t^\top=\bm{\Lambda}_{t-1}^{1/2}(\mathbf{I}_{N+1}+\bm{\Lambda}_{t-1}^{-1/2}\p_t\p_t^\top\bm{\Lambda}_{t-1}^{-1/2})\bm{\Lambda}_{t-1}^{1/2}$, we have 
\begin{align}\nonumber
    \mathrm{det}(\bm{\Lambda}_t)&= \mathrm{det}(\bm{\Lambda}_{t-1})(1+\p_t^\top\bm{\Lambda}_{t-1}^{-1}\p_t)\\&\geq\mathrm{det}(\bm{\Lambda}_{t-1})\exp\left(\frac{1}{2}\min\{1,\p_t^\top\bm{\Lambda}_{t-1}^{-1}\p_t\}\right),\label{eq:lepo3}
\end{align}
where the first equility is due to $\mathrm{det}(\mathbf{I}+\bm{x}\bm{x}^\top)=1+\|\bm{x}\|_2^2$ and the last inequality is due to $\exp(x/2)\leq 1+x$ when $x\in[0,1]$.

Therefore , with \eq{\ref{eq:lepo3}} we have 
\begin{align}\nonumber
    \sum_{t=1}^\tau \min\left\{1,\sqrt{\p_t^\top\bm{\Lambda}_t^{-1}\p_t}\right\}&\leq \sqrt{\tau}\sqrt{\sum_{t=1}^\tau \min\{1,\p_t^\top\bm{\Lambda}_t^{-1}\p_t\}}\\\nonumber&\leq \sqrt{T}\sqrt{2\ln\mathrm{det}(\bm{\Lambda}_\tau)-2\ln\mathrm{det}(\bm{\Lambda}_0)}\\\nonumber&\leq \sqrt{T}\sqrt{2\ln\left(\frac{\mathrm{trace}(\bm{\Lambda}_\tau)}{N+1}\right)^{N+1}}\\&\leq \sqrt{T}\O(\sqrt{N\log(N+1+\bar p^2T)},\label{eq:lepo5}
\end{align}
where the first inequality is due to Cauchy-Schwarz inequality, the third inequality is due to the AM-GM inequality, and the last inequality is due to $\mathrm{trace}(\bm{\Lambda}_t)=\mathrm{trace}((N+1)\cdot\mathbf{I}_{N+1})+\mathrm{trace}(\sum_{s<t}\widetilde{\bm p}_s\widetilde{\bm p}_s^\top)\leq (N+1)(N+1+T\bar p^2)$.

 Combining \eq{\ref{eq:minrphi}} and \eq{\ref{eq:lepo5}}, we have
\begin{align*}
    \sum_{t=1}^\tau \Delta_{t}\left(\bm{p}_t\right)&\leq\O\left(\sqrt{N+1}\kappa\max\{\bar p,1\}(N\bar p+\max\{C\|\bm{A}\|_\infty, 1\})
\right)\times\sqrt{NT\log(N+1+\bar p^2T)},
\end{align*}
where we complete the proof.
\Halmos
\endproof

\subsection{Proof of Lemma~\ref{le:R_r}}
\begin{lemma}\label{le:existence}
With probability at least $(1-\mathcal O\left(T^{-1}\right))$, for all $t\le\tau$, we have $(\bm{\alpha}, \bm{B}-T^{-2}\cdot \mathbf{I}_{N})\in\M_t$, where $(\bm{\alpha},\bm{B})$ is the underlying true parameter and $\mathcal M_t = \left\{(\tilde{\bm{\alpha}},\tilde{\bm{B}}) : 
    \|(\tilde{\mathbfcal{B}}-\hat{\mathbfcal{B}}_{t})^\top \bm{e}_i\|_{\bm{\Lambda}_t}\le\kappa,   \|\tilde{\mathbfcal{B}}^\top \bm{e}_i\|_2 \leq 2L_B ~ \forall i\in[N] \text{~and~} \tilde{\bm{B}}+\tilde{\bm{B}}^\top  \preceq 0
    \right\}$.
\end{lemma}
\proof{Proof of Lemma~\ref{le:existence}.}
Note that by the triangle inequality, for all $t\leq \tau$ and $i\in[N]$ we have
\begin{align}
   \label{eq:letriangle} \|([\bm{B} - T^{-2} \cdot \mathbf{I}_{N} |\bm{\alpha}]-\hat{\mathbfcal{B}}_{t})^\top \bm{e}_i\|_{\bm{\Lambda}_t}&\leq  \|(\mathbfcal{B}-\hat{\mathbfcal{B}}_{t})^\top \bm{e}_i \|_{\bm{\Lambda}_t}+\|[ T^{-2} \cdot \mathbf{I}_{N} |0]^\top \bm{e}_i\|_{\bm{\Lambda}_t}.
\end{align}
For $\|([\bm{B} - T^{-2} \cdot \mathbf{I}_{N} |\bm{\alpha}]-\hat{\mathbfcal{B}}_{t})^\top \bm{e}_i\|_{\bm{\Lambda}_t}$, by \eq{\ref{eq:leucb00}},
with probability $1-1/T$ for any $i\in[N]$ and $t\le\tau$ it holds that
\begin{align}
    \|(\mathbfcal{B}-\hat{\mathbfcal{B}}_{t})^\top \bm{e}_i \|_{\bm{\Lambda}_t}\leq \sqrt{2\bar d^2(N+1)\ln\left(NT(1+\bar p^2T)\right)+2(N+1)L_B^2},\label{eq:letriangle1}
\end{align}
Let $\mathrm{diag}(\bm{\Lambda}_t)_i$ be the $i$-th element of the diagonal of  $\bm{\Lambda}_t$.
For $\|[ T^{-2} \cdot \mathbf{I}_{N} | \bm{0}]^\top \bm{e}_i\|_{\bm{\Lambda}_t}$, we have 
\begin{align}
   \nonumber\|[ T^{-2} \cdot \mathbf{I}_{N} |0]^\top \bm{e}_i\|_{\bm{\Lambda}_t}^2&\leq\mathrm{diag}(\bm{\Lambda}_t)_i\cdot T^{-4}\nonumber\\&\leq (N+1+\bar p^2T)/T^{4}\label{eq:ei},
\end{align}
where the last inequality is due to $\mathrm{diag}(\bm{\Lambda}_t)_i=N+1+\mathrm{diag}(\sum_{s<t}\widetilde{\bm p}_s\widetilde{\bm p}_s^\top)_i\leq N+1+\bar p^2T$.

Invoking Eq.~(\ref{eq:letriangle1}) and Eq.~(\ref{eq:ei}) into Eq.~(\ref{eq:letriangle}), we have
\begin{align}
    &\|([\bm{B} - T^{-2} \cdot \mathbf{I}_{N} |\bm{\alpha}]-\hat{\mathbfcal{B}}_{t})^\top \bm{e}_i\|_{\bm{\Lambda}_t}\nonumber
    \\&\leq \sqrt{2\bar d^2(N+1)\ln\left(NT(1+\bar p^2T)\right)+2(N+1)L_B^2}+\sqrt{(N+1+\bar p^2T)/T^{4}}\label{eq:b-t}\\&
    \leq 2\sqrt{2\bar d^2(N+1)\ln\left(NT(1+\bar p^2T)\right)+2(N+1)L_B^2}=\kappa.\nonumber
\end{align}
By assumption $\|\mathbfcal{B}^\top \bm{e}_i \|_{2}\leq L_B$ and $L_B\geq1$, we have
\begin{align}\nonumber
   \|[\bm{B} - T^{-2} \cdot \mathbf{I}_{N} |\bm{\alpha}]^\top \bm{e}_i\|_{2}&\leq  \|\mathbfcal{B}^\top \bm{e}_i \|_{2}+\|[ T^{-2} \cdot \mathbf{I}_{N} |0]^\top \bm{e}_i\|_{2}\\&\leq L_B+1/T^2\label{eq:leboundcheckB1}\\&\leq 2L_B.\nonumber
\end{align}
And it is easy to note that $(\bm B - T^{-2} \cdot \mathbf{I}_{N})+(\bm B - T^{-2} \cdot \mathbf{I}_{N})^\top\preceq 0$ with the assumption $\bm B+\bm B^\top\preceq 0$. Therefore, we complete the proof of this lemma.
\Halmos
\endproof
\begin{lemma}\label{le:robustness}
When the desired event in Lemma~\ref{le:existence} happens
, $[\bm{B} - T^{-2} \cdot \mathbf{I}_{N} |\bm{\alpha}]+\bm X\in\M_t$ for any $\bm{X}\in\{\bm{X}\in\R^{N\times (N+1)}|\|\bm{X}\|_F\leq 1/T^{4}\}$.
\end{lemma}
\proof{Proof of Lemma~\ref{le:robustness}}

Note that by the triangle inequality, for all $t\leq \tau$ and $i\in[N]$ we have
\begin{align}
   \label{eq:le2triangle} \|([\bm{B} - T^{-2} \cdot \mathbf{I}_{N} |\bm{\alpha}]+\bm{X}-\hat{\mathbfcal{B}}_{t})^\top \bm{e}_i\|_{\bm{\Lambda}_t}&\leq  \|([\bm{B} - T^{-2} \cdot \mathbf{I}_{N} |\bm{\alpha}]-\hat{\mathbfcal{B}}_{t})^\top \bm{e}_i\|_{\bm{\Lambda}_t}+\|\bm{X}^\top \bm{e}_i\|_{\bm{\Lambda}_t}.
\end{align}
When the desired event in Lemma~\ref{le:existence} happens, by \eq{\ref{eq:b-t}},
for any $i\in[N]$ and $t\le\tau$ it holds that
\begin{align}
   \|([\bm{B} - T^{-2} \cdot \mathbf{I}_{N} |\bm{\alpha}]-\hat{\mathbfcal{B}}_{t})^\top \bm{e}_i\|_{\bm{\Lambda}_t}\leq \sqrt{2\bar d^2(N+1)\ln\left(NT(1+\bar p^2T)\right)+2(N+1)L_B^2}+\sqrt{(N+1+\bar p^2T)/T^{4}}.\label{eq:2b-t}
\end{align}
And for any $\bm{X}\in\{\bm{X}\in\R^{N\times (N+1)}|\|\bm{X}\|_F\leq 1/T^{4}\}$,  we can upper bound $\|\bm{X}^\top \bm{e}_i\|_{\bm{\Lambda}_t}$ as follows
\begin{align}
   \nonumber\|\bm{X}^\top \bm{e}_i\|_{\bm{\Lambda}_t}^2&\leq\lambda_{\mathrm{max}}(\bm{\Lambda}_t) \|\bm{X}^\top \bm{e}_i\|_2^2\nonumber\\&\leq \lambda_{\mathrm{max}}(\bm{\Lambda}_t)\cdot T^{-8}
   \nonumber\\&\leq (N+1)(1+\bar p^2T)/T^{8}\label{eq:2ei},
\end{align}
where the first inequality is due to $x^\top \bm{\Lambda} x\leq \lambda_{\mathrm{max}}(\bm{\Lambda})\|\bm{x}\|_2^2$ for any symmetric matrix $\bm{\Lambda}$, the second  inequality is due to $\|\bm{X}^\top \bm{e}_i\|_2\leq\|\bm{X}\|_F\leq 1/T^4$, and the last inequality is due to $\lambda_{\mathrm{max}}(\bm{\Lambda}_t)=N+1+\lambda_{\mathrm{max}}(\sum_{s<t}\widetilde{\bm p}_s\widetilde{\bm p}_s^\top)\leq N+1+\mathrm{trace}(\sum_{s<t}\widetilde{\bm p}_s\widetilde{\bm p}_s^\top)\leq (N+1)(1+\bar p^2T)$.

Invoking Eq.~(\ref{eq:2b-t}) and Eq.~(\ref{eq:2ei}) into Eq.~(\ref{eq:le2triangle}), we have
\begin{align}
    &\|([\bm{B} - T^{-2} \cdot \mathbf{I}_{N} |\bm{\alpha}]+\bm{X}-\hat{\mathbfcal{B}}_{t})^\top \bm{e}_i\|_{\bm{\Lambda}_t}\nonumber
    \\&\leq \sqrt{2\bar d^2(N+1)\ln\left(NT(1+\bar p^2T)\right)+2(N+1)L_B^2}+\sqrt{(N+1+\bar p^2T)/T^{4}}+\sqrt{(N+1)(1+\bar p^2T)/T^{8}}\nonumber\\&
    \leq 2\sqrt{2\bar d^2(N+1)\ln\left(NT(1+\bar p^2T)\right)+2(N+1)L_B^2}=\kappa.\label{eq:lerobustness}
\end{align}
And by \eq{\ref{eq:leboundcheckB1}}, we have
\begin{align}
    \|([\bm{B} - T^{-2} \cdot \mathbf{I}_{N} |\bm{\alpha}]+\bm{X})^\top \bm{e}_i\|_{2}&\leq  \|[\bm{B} - T^{-2} \cdot \mathbf{I}_{N} |\bm{\alpha}]^\top \bm{e}_i\|_{2}+\|\bm{X}^\top \bm{e}_i\|_{2}\nonumber\\&\leq L_B+1/T^2+1/T^4\nonumber\\&\leq 2L_B .\label{eq:boundcheckD2}
\end{align}

Now we need to prove the third constraint is satisfied by $[\bm{B} - T^{-2} \cdot \mathbf{I}_{N} |\bm{\alpha}]+\bm{X}$. To facilitate our discussion let $\tilde{\bm{X}}\in \R^{N\times N}$ be the square matrix after deleting the last column of $\bm{X}$. 

Since we have the assumption that $\bm B+\bm B^\top\preceq 0$, we only need to show $(\tilde{\bm{X}} -T^{-2}\cdot \mathbf{I}_{N})+(\tilde{\bm{X}} -T^{-2}\cdot \mathbf{I}_{N})^\top\preceq 0$ to prove that 
$(\bm{B}-T^{-2}\cdot \mathbf{I}_{N} +\tilde{\bm{X}})+(\bm{B}-T^{-2}\cdot \mathbf{I}_{N} +\tilde{\bm{X}})\preceq 0$.

By the fact  that $\lambda_{\mathrm{max}}(\bm{\Lambda})=\|\bm{\Lambda}\|_2$  and $\|\bm{\Lambda}\|_2\leq\|\bm{\Lambda}\|_F$ if $\bm{\Lambda}$ is symmetric, we have 
\begin{align*}
    \lambda_{\mathrm{max}}(\tilde{\bm{X}}+\tilde{\bm{X}}^\top-2T^{-2}\cdot \mathbf{I}_{N})&= \lambda_{\mathrm{max}}(\tilde{\bm{X}}+\tilde{\bm{X}}^\top)-2/T^{2}\\&\leq \|\tilde{\bm{X}}+\tilde{\bm{X}}^\top\|_F-2/T^{2}\\&\leq 2\|\tilde{\bm{X}}\|_F-2/T^{2}\\&\leq0,
\end{align*}
where the last inequality is due to $\|\tilde{\bm{X}}\|_F\leq\|\bm{X}\|_F\leq 1/T^{4}$.

Therefore, combining $(\bm{B}-T^{-2}\cdot \mathbf{I}_{N} +\tilde{\bm{X}})+(\bm{B}-T^{-2}\cdot \mathbf{I}_{N} +\tilde{\bm{X}})^\top\preceq 0$ with \eq{\ref{eq:lerobustness}} and \eq{\ref{eq:boundcheckD2}} , we complete the proof of this lemma.

\Halmos
\endproof
\proof{Proof of Lemma~\ref{le:R_r}.}
Note that $\bm{\Lambda}_t=(N+1)\cdot\mathbf{I}_{N+1}+\sum_{s<t}\widetilde{\bm p}_s\widetilde{\bm p}_s^\top$,
For all $i\in [N]$, we have
$$\|(\tilde{\mathbfcal{B}}-\hat{\mathbfcal{B}}_{t})^\top \bm{e}_i\|_2\leq\|(\tilde{\mathbfcal{B}}-\hat{\mathbfcal{B}}_{t})^\top \bm{e}_i\|_{\bm{\Lambda}_t}.$$
And thus, for $(\tilde{\bm{\alpha}},\tilde{\bm{B}})\in \M_t$, it holds that
$$\|(\tilde{\mathbfcal{B}}-\hat{\mathbfcal{B}}_{t})^\top \bm{e}_i\|_2\leq \k \qquad\forall i\in[N].$$
Therefore, we have $\|\tilde{\mathbfcal{B}}-\hat{\mathbfcal{B}}_{t}\|_F\leq \k\sqrt{N}$, i.e., $\M_t\subseteq \mathrm{Ball}(\hat{\mathbfcal{B}}_t, \kappa \sqrt{N})$ when treating the matrix $\hat{\mathbfcal{B}}_t$  as an $N \times (N+1)$-dimensional vector.

Combining Lemma~\ref{le:existence} and Lemma~\ref{le:robustness}, we will have the following conclusion to complete the proof of the second part of this lemma.

Given the desired event \eq{\ref{eq:leucb00}} in Lemma~\ref{le:ucb0}, for all $t\le\tau$, it holds that $[\bm{B} - T^{-2} \cdot \mathbf{I}_{N} |\bm{\alpha}]+X\in\M_t$ for any $X\in\{X\in\R^{N\times (N+1)}|\|\bm{X}\|_F\leq 1/T^{4}\}$, i.e., 
$\mathrm{Ball}([\bm{B} - T^{-2} \cdot \mathbf{I}_{N} |\bm{\alpha}], T^{-4}) \subseteq \M_t$, when treating the matrix $[\bm{B} - T^{-2} \cdot \mathbf{I}_{N} |\bm{\alpha}]$ as an $N \times (N+1)$-dimensional vector. \Halmos
\endproof
\section{Proof of Theorem~\ref{thm:main}}\label{sec:regretanalysis}
This section is devoted to the proof of Theorem~\ref{thm:main}, which is detailed in 5 steps. Recall that $\tau = \max \{t : \min_{i \in [M]} [\bm{I}_t]_i > 0, t \leq T\}$ is the stopping time till when the inventory levels of all resources remain positive. For convenience, for $t \leq \tau$, we define
\begin{align}
	\bm g_t := -\bm{A}\bm D(\bm{p}_t) + \bm s_t. \label{eq:def-g-t}
\end{align}
For $t > \tau$, we set all relevant quantities to zeros:\footnote{We will treat $\bm D(\bm{p}_t)$ as a symbol rather than a function of $\bm p_t$ for $t > \tau$.}
\[
\bm g_t = \hat{\bm g}_t = \bm s_t = 0, \qquad  \bm p_t = 0, \qquad \bm D(\bm{p}_t) = \bm{d}_t = 0, \qquad\qquad \forall t > \tau.
\]

By Eq.~(\ref{regretform2}), we have that
\begin{align}\label{eq:regretform2}
	\mathcal{R}(T) &\leq T[r(\bm{p}^*)+\phi(\bm{A}\bm{\bm{D}}(\bm{p}^*))]-\E\left[\sum_{t=1}^Tr(\bm{p}_t)+T\phi\left(\frac{1}{T}\sum_{t=1}^T\bm{A}\bm{d}_t\right)\right].
\end{align}

\subsection{Step I: Replacing $\bm{d}_t$ with $\bm{D}(\bm{p}_t)$ and Introducing $\mathcal{\widetilde{R}}$}
The first step is to replace the real demand $\bm{d}_t$ on the Right-Hand-Side of Eq.~\eqref{eq:regretform2} with the expected demand $\bm D(\bm{p}_t)$ so that the resulting expression $\mathcal{\widetilde{R}}$ is easier to deal with. By applying the Azuma-Hoeffding inequality and a union bound, we have the following lemma.

\begin{lemma}\label{le:concentration}
	With probability at least $(1-1/T)$, it holds that 
	\begin{equation}\label{eq:hoefding00}
		\forall i\in[M], \qquad \left|\sum_{t=1}^T [\bm{A}\bm{d}_t]_i-\sum_{t=1}^T [\bm{A}\bm D(\bm{p}_t)]_i\right|\leq \mathcal{O}(\|\bm{A}\|_{\infty}\bar d\sqrt{T\log(MT)}).
	\end{equation}
\end{lemma}
Lemma~\ref{le:concentration} implies that
\begin{align}\label{eq:hoefding01}
	\left\|\sum_{t=1}^T\bm{A}\bm{d}_t-\sum_{t=1}^T \bm A\bm D(\bm{p}_t)\right\|_{\infty}\leq \mathcal{O}(\|\bm{A}\|_{\infty}\bar d\sqrt{T\log (MT)}).
\end{align}
Together with the Lipschtz continuity of $\phi(\cdot)$, Eq.~\eqref{eq:hoefding01} implies that
\begin{align}\label{eq:lipphi}
	\left|\phi\left(\frac{1}{T}\sum_{t=1}^T\bm{A}\bm{d}_t\right)-\phi\left(\frac{1}{T}\sum_{t=1}^T\bm{A}\bm D(\bm{p}_t)\right)\right|&\leq \frac{L}{T}\left\|\sum_{t=1}^T\bm{A}\bm{d}_t-\sum_{t=1}^T\bm{A}\bm D(\bm{p}_t)\right\|_\infty \leq \mathcal{O}\left(L\|\bm{A}\|_{\infty}\bar d\sqrt{\frac{\log (MT)}{T}}\right).
\end{align}

Now we define the random variable 
\begin{align}
	\mathcal{\widetilde{R}} := T[r(\bm{p}^*)+\phi(\bm{A}\bm{\bm{D}}(\bm{p}^*))]-\left[\sum_{t=1}^\tau r(\bm{p}_t)+T\phi\left(\frac{1}{T}\sum_{t=1}^T\bm{A}\bm D(\bm{p}_t)\right)\right]. \label{eq:def-tildeR}
\end{align}
Eq.~\eqref{eq:lipphi} (which holds with probability at least $(1-1/T)$) implies that
\begin{equation}\label{eq:thm1jenson}
	\mathcal{R}(T)\leq \E[\mathcal{\widetilde{R}}] +\mathcal{O}(L\|\bm{A}\|_{\infty}\bar d\sqrt{T\log (MT)}),
\end{equation}
and we may turn to upper bound $\E[\mathcal{\widetilde{R}}]$ instead. In the following steps, we will upper bound the value of $\mathcal{\widetilde{R}}$ conditioned on that the desired events of Lemma~\ref{le:concentration} and Lemma~\ref{le:ucb0} hold. Note that this scenario happens with probability at least $(1 - \mathcal{O}(T^{-1}))$, and $\widetilde{\mathcal{R}}$ is at most $T[r(\bm{p}^*)+\phi(\bm{A}\bm{\bm{D}}(\bm{p}^*))]$ in the rare opposite case.

\subsection{Step II: Bounding the Fluid Optimum by the UCB of the Dual} 
The goal of the second step is, given the desired event of Corollary~\ref{cor:check-error-bound}, to establish for each $t \leq \tau$ that 
\begin{align}
	r(\bm{p}^*)+\phi\left(\bm{A}\bm{D}(\bm{p}^*)\right) \leq r(\bm{p}_t)+\phi\left(\bm{s}_{t}\right) + \left\langle\bm{\mu}_{t}, -\bm{A}\hat{\bm{D}}_{t}(\bm{p}_t) + \bm s_{t}\right\rangle +2( \Delta_{t}^{r}\left(\bm{p}_t\right)+\Delta_t^f\left(\bm{p}_t\right)) ,
	\label{eq:rplushphi}
\end{align}
where $\bm p^*$ is the optimal solution of the fluid model (Eq.~\eqref{eq:primal-opt-def-1}), and recall the definitions of the primal variables $\bm p_t$ and $\bm s_t$ in Eq.~\eqref{eq:primal-update}. The Right-Hand-Side of Eq.~\eqref{eq:rplushphi} can be viewed as the Upper-Confidence-Bound of the dual function $\mathfrak{q}(\bm{\mu}_t)$.

To prove Eq.~\eqref{eq:rplushphi}, we first introduce $\bm p^*_t :=  \argmax_{\bm{p}\in[\underline{p},\overline{p}]^N} \left\{ r(\bm{p})- \langle \bm{\mu}_t, \bm{A}\bm{\bm{D}}(\bm{p})\rangle \right\}$ which can be viewed as the desired choice for $\bm p_t$ (without the estimation errors of $r(\cdot)$ and $\bm{D}(\cdot)$). In the following claim we upper bound the fluid optimum by the exact dual function.
\begin{claim} \label{claim:rplushphi-pf-0}
	$r(\bm{p}^*)+\phi\left(\bm{A}\bm{D}(\bm{p}^*)\right) \leq r(\bm{p}_t^*) + \phi(\bm{s}_t) + \langle \bm{\mu}_t, -\bm{A}\bm{D}(\bm{p}_t^*) +\bm{s}_t\rangle $ .
\end{claim}
Claim~\ref{claim:rplushphi-pf-0} is essentially a restatement of the weak duality and its proof is deferred to Section~\ref{sec:proofofclaim}. Comparing Claim~\ref{claim:rplushphi-pf-0} and our goal (Eq.~\eqref{eq:rplushphi}), we only need to upper bound the estimation errors. In particular, it suffices to have that
\begin{align*}
	|r\left(\bm{p}_t\right) - \check r_t\left(\bm{p}_t\right) | \leq  2\Delta_{t}^{r}\left(\bm{p}_t\right) \qquad  \text{and}  \qquad  r(\bm{p}_t^*) +  \left\langle\bm{\mu}_{t}, -\bm{A}\bm{D}(\bm{p}_{t}^{*})\right\rangle \leq \check r_t\left(\bm{p}_t\right) + \left\langle\bm{\mu}_{t}, -\bm{A} \check{\bm{D}}_t(\bm{p}_t) \right\rangle +2\Delta_t^f(\bm{p}_t),
\end{align*}
where the first inequality is exactly guaranteed by the first item of Corollary~\ref{cor:check-error-bound}, for the second inequality,
we have that 
\[
r\left(\bm p_{t}^{*}\right)-\left\langle\bm{\mu}_{t},\bm{A}\bm{D}(\bm{p}_{t}^{*})\right\rangle = f_{t}\left(\bm p_{t}^{*}\right) \leq \bar{f}_{t}\left(\bm p_{t}^{*}\right) \leq \bar{f}_{t}\left(\bm{p}_t\right) = \hat r_t\left(\bm{p}_t\right) + \left\langle\bm{\mu}_{t}, -\bm{A} \hat{\bm D}_t(\bm{p}_t) \right\rangle+2\Delta_t^f(\bm{p}_t),
\]
where the first inequality is by the third item of Corollary~\ref{cor:check-error-bound} and the second inequality is due to our Upper-Confidence-Bound-style primal update (Eq.~\eqref{eq:primal-update}).

Now we have established Eq.~\eqref{eq:rplushphi}. Together with the definition of $\mathcal{\widetilde{R}}$ (\eq{\ref{eq:def-tildeR}}) and that $\check{\bm{g}}_{t}=-\bm{A}\check{\bm{D}}_{t}(\bm{p}_t)+\bm{s}_{t}$ (Line~\ref{line:subgradient-g} of Algorithm~\ref{algucb}), we obtain that
\begin{align}
	\mathcal{\widetilde{R}}&\leq (T-\tau)[r(\bm{p}^*)+\phi(\bm{A}\bm{\bm{D}}(\bm{p}^*))]+\sum_{t=1}^\tau 2(\Delta_{t}^{r}(\bm{p}_t)+\Delta_t^f(\bm{p}_t))\nonumber \\
	&\qquad \qquad\qquad \qquad\qquad \qquad\qquad \qquad +\sum_{t=1}^\tau\left\langle\bm{\mu}_{t}, \check{\bm{g}}_{t}\right\rangle+\sum_{t=1}^\tau\phi(\bm{s}_t)-T\phi\left(\frac{1}{T}\sum_{t=1}^T\bm{A}\bm D(\bm{p}_t)\right). \label{eq:thm1regret1}
\end{align}

Observe that in Eq.~\eqref{eq:thm1regret1} we have $\sum_{t=1}^\tau\phi(\bm{s}_t)-T\phi\left(\frac{1}{T}\sum_{t=1}^T\bm{A}\bm D(\bm{p}_t)\right)$ and $\sum_{t=1}^\tau\left\langle\bm{\mu}_{t}, \check{\bm{g}}_{t}\right\rangle$. In the next two steps, we will bound them separately.

\subsection{Step III: Upper Bounding $\sum_{t=1}^\tau\phi(\bm{s}_t)-T\phi\left(\frac{1}{T}\sum_{t=1}^T\bm{A}\bm D(\bm{p}_t)\right)$}
In this step we upper bound the term $\sum_{t=1}^\tau\phi(\bm{s}_t)-T\phi\left(\frac{1}{T}\sum_{t=1}^T\bm{A}\bm D(\bm{p}_t)\right)$ in \eq{\ref{eq:thm1regret1}} by the dot products between a carefully selected dual variable $\bar{\bm{\mu}}$ and $\{\bm g_t\}$ (Eq.~\eqref{eq:def-g-t}). It is also important to guarantee that $\bar{\bm{\mu}}$ stays in the range of our novel dual space $\D$, which is used by our later analysis. Formally, we prove the following lemma.

\begin{lemma}\label{le:duala}
	Let 
	\begin{align} \label{eq:def-bar-mu}
		\bar{\bm s} =\frac{1}{T} \sum_{t=1}^{T} \bm{A} \bm D(\bm{p}_t) \qquad  \text{and}  \qquad \bar{\bm{\mu}} =\argmax _{  \bm{\mu} \in \mathbb{R}^{M}}\{-(-\phi)^*(\bm{\mu})+\langle\bm{\mu}, \bar{\bm s}\rangle\}.
	\end{align}
	It holds that
	\begin{align}
		\sum_{t=1}^{T} \phi\left(\bm{s}_{t}\right)-T\phi\left(\frac{1}{T} \sum_{t=1}^{T} \bm{A} \bm D(\bm{p}_t)\right)\leq -\sum_{t=1}^T\left\langle\bar{\bm{\mu}},\bm{g}_{t}\right\rangle=-\sum_{t=1}^\tau\left\langle\bar{\bm{\mu}},\bm{g}_{t}\right\rangle. \label{eq:duala}
	\end{align}
	Moreover, we have that $\bar{\bm{\mu}}\in \D$.
\end{lemma}

Noting that when $t>\tau$ we have $s_t=0$ and thus $\phi(\bm{s}_t)\geq 0$ (by Assumption~\ref{assump:phi1}). Together with Eq.~\eqref{eq:thm1regret1} and Lemma~\ref{le:duala} we further upper bound $\mathcal{\widetilde{R}}$ by
\begin{align}\label{eq:thm1regret1.5}
	\mathcal{\widetilde{R}}&\leq (T-\tau)[r(\bm{p}^*)+\phi(\bm{A}\bm{\bm{D}}(\bm{p}^*))]+\sum_{t=1}^\tau 2(\Delta_{t}^{r}(\bm{p}_t)+\Delta_t^f(\bm{p}_t))+\sum_{t=1}^\tau\left\langle\bm{\mu}_{t}, \check{\bm{g}}_{t}\right\rangle-\sum_{t=1}^\tau\left\langle\bar{\bm{\mu}},\bm{g}_{t}\right\rangle.
\end{align}

\subsection{Step IV: Upper Bounding $\sum_{t=1}^\tau\left\langle\bm{\mu}_{t}, \check{\bm{g}}_{t}\right\rangle$}

In this step, we upper bound the term $\sum_{t=1}^\tau\left\langle\bm{\mu}_{t}, \check{\bm{g}}_{t}\right\rangle$ in \eq{\ref{eq:thm1regret1.5}} (as well as Eq.~\eqref{eq:thm1regret1}) by combining the properties of the mirror descent solver and our Upper-Confidence-Bound-type estimator. Intuitively, we would like to replace $\check{\bm{g}}_{t}$ by $\bm{g}_t$ so that the term could be compared with the other term $\sum_{t=1}^\tau\left\langle \bar{\bm{\mu}}, \bm{g}_{t}\right\rangle$ in \eq{\ref{eq:thm1regret1.5}}. Formally, we will establish Eq.~\eqref{eq:mughat}. 

For any $\bm{\mu}\in\mathcal{\bm{D}}$, by applying the definition of the mirror descent solver in Definition~\ref{def:dualdescent}, we have that
\begin{align}
	\sum_{t=1}^\tau\left\langle\bm{\mu}_{t}, \check{\bm{g}}_{t}\right\rangle &\leq \sum_{t=1}^\tau\left\langle\bm{\mu}, \check{\bm{g}}_{t}\right\rangle+\frac{C_1}{\eta}+C_2\eta T \nonumber\\
	&\leq \sum_{t=1}^\tau\left\langle\bm{\mu},\bm{g}_{t}\right\rangle+ \sum_{t=1}^\tau\left|\Braket{\bm{\mu},\bm{A}\check{\bm{D}}_{t}(\bm{p}_t)-\bm{A}\bm D(\bm{p}_t)}\right|+\frac{C_1}{\eta}+C_2\eta T,\label{eq:ghatg}
\end{align}
where $C_1$ and $C_2$ are the constant parameters in Definition~\ref{def:dualdescent}.  By H\"older's Inequality, (for any $\bm{\mu} \in \D$) we have that
\begin{align}\label{eq:cs}
	\left|\Braket{\bm{\mu},\bm A\check{\bm{D}}_{t}(\bm{p}_t)-\bm{A}\bm D(\bm{p}_t)}\right|\leq\|\bm{\mu}\|_1\cdot\|\bm A\check{\bm{D}}_{t}(\bm{p}_t)-\bm{A}\bm D(\bm{p}_t)\|_{\infty}\leq 2C\|\bm{A}\|_\infty\Delta_{t}^{\bm{D}}\left(\bm{p}_t\right).
\end{align}
Note that here we crucially rely on our definition of the dual space $\D = \left\{  \bm{\mu} \in \mathbb{R}^{M} \mid \|\bm{\mu}\|_1\leq C \right\}$.
Combining Eq.~\eqref{eq:ghatg} and Eq.~\eqref{eq:cs}, for any $\bm{\mu} \in \D$, we establish that
\begin{align}\label{eq:mughat}
	\sum_{t=1}^\tau\left\langle\bm{\mu}_{t}, \check{\bm{g}}_{t}\right\rangle &\leq \sum_{t=1}^\tau\left\langle\bm{\mu},\bm{g}_{t}\right\rangle+\frac{C_1}{\eta}+C_2\eta T+\sum_{t=1}^\tau 2C\|\bm{A}\|_\infty\Delta_{t}^{\bm{D}}\left(\bm{p}_t\right),
\end{align}

Recall the definition of $\bar{\bm{\mu}}$ in Eq.~\eqref{eq:def-bar-mu}. Let $\bm{\mu}=\bar{\bm{\mu}}+\bm \delta$ where $\bm \delta\in\R^M_+$ satisfying $\bar{\bm{\mu}}+\bm \delta \in \mathcal D$ will be determined later. Plugging our choice of $\bm{\mu}$ into Eq.~\eqref{eq:mughat}, we have that
\begin{align}\label{eq:thm1mirrordual}
	\sum_{t=1}^\tau\left\langle\bm{\mu}_{t}, \check{\bm{g}}_{t}\right\rangle&\leq \sum_{t=1}^\tau\left\langle\bar{\bm{\mu}},\bm{g}_{t}\right\rangle+\sum_{t=1}^\tau\left\langle\bm \delta,\bm{g}_{t}\right\rangle+\frac{C_1}{\eta}+C_2\eta T+\sum_{t=1}^\tau2 C\|\bm{A}\|_\infty\Delta_{t}^{\bm{D}}\left(\bm{p}_t\right).
\end{align}

Recalling $\Delta_t(\bm{p}) := \Delta_t^r(\bm{p})+ \max\{C\|\bm{A}\|_\infty, 1\} \Delta_t^{\bm{D}}(\bm{p})$ and combining Eq.~\eqref{eq:thm1regret1.5} and Eq.~\eqref{eq:thm1mirrordual}, we obtain that
\begin{align}\label{eq:thm1regret2}
	\mathcal{\widetilde{R}}&\leq (T-\tau)[r(\bm{p}^*)+\phi(\bm{A}\bm{\bm{D}}(\bm{p}^*))]+\sum_{t=1}^\tau\left\langle\bm{\delta},\bm{g}_{t}\right\rangle+\frac{C_1}{\eta}+C_2\eta T+\sum_{t=1}^\tau 4\Delta_t(\bm{p}_t).
\end{align}

\subsection{Step V: Choosing Parameters and Putting Things Together} 
We finally choose the proper parameters to upper bound $\mathcal{\widetilde{R}}$ and conclude the proof. For the choice of $\bm{\delta}$, we discuss the following two cases.

\paragraph{\underline{Case 1: $\tau=T$.}} If none of the resources depletes before time horizon $T$, i.e., $\tau=T$, we set $\bm{\delta} = 0$. Now, Eq.~\eqref{eq:thm1regret2} implies that
\begin{align*}
	\mathcal{\widetilde{R}}\leq \frac{C_1}{\eta}+C_2\eta T+ \sum_{t=1}^\tau 4\Delta_t(\bm{p}_t).
\end{align*}
\paragraph{\underline{Case 2: $\tau<T$.}}If $\tau<T$, then there exists a resource $i \in [M]$ such that
\begin{equation}\label{eq:deplete}
	\sum_{t=1}^\tau [\bm{A}\bm{d}_t]_i+\|\bm{A}\|_{\infty}\bar d \geq T \gamma_i.  
\end{equation}
We now set $\bm{\delta}=\left([r(\bm{p}^*)+\phi(\bm{A}\bm{\bm{D}}(\bm{p}^*))]/\gamma_i\right)\bm{e}_i$ where $\bm{e}_i$ is the $i$-th unit vector, it is easy to verify that $\bm{\mu}=\bar{\bm{\mu}}+\bm{\delta} \in \mathcal D$. Thus, combining Eq.~(\ref{eq:hoefding00}) and Eq.~(\ref{eq:deplete}), we have that
\begin{align}
	\sum_{t=1}^\tau\left\langle\bm{\delta},\bm{g}_{t}\right\rangle &\leq\left([r(\bm{p}^*)+\phi(\bm{A}\bm{\bm{D}}(\bm{p}^*))]/\gamma_i\right)\sum_{t=1}^\tau([\bm s_t]_i-[\bm{A}\bm D(\bm{p}_t)]_i)\nonumber
	\\&\leq\left([r(\bm{p}^*)+\phi(\bm{A}\bm{\bm{D}}(\bm{p}^*))]/\gamma_i\right)\sum_{t=1}^\tau(\gamma_i-[\bm{A}\bm{d}_t]_i)+\mathcal{O}(\|\bm{A}\|_{\infty}\bar d\sqrt{T\log (MT)})\nonumber
	\\&\leq\left([r(\bm{p}^*)+\phi(\bm{A}\bm{\bm{D}}(\bm{p}^*))]/\gamma_i\right)\left((\tau- T)\gamma_i+\|\bm{A}\|_{\infty}\bar d\right)+\mathcal{O}(\|\bm{A}\|_{\infty}\bar d\sqrt{T\log (MT)}).\label{eq:wdelta}
\end{align}
Plugging Eq.~(\ref{eq:wdelta}) back into Eq.~(\ref{eq:thm1regret2}), we obtain that
\begin{align*}
	\mathcal{\widetilde{R}}&\leq\left(\|\bm{A}\|_{\infty}\bar d/\underline{\gamma}\right)[r(\bm{p}^*)+\phi(\bm{A}\bm{\bm{D}}(\bm{p}^*))]+\mathcal{O}(\|\bm{A}\|_{\infty}\bar d\sqrt{T\log (MT)})+\frac{C_1}{\eta}+C_2\eta T+ \sum_{t=1}^\tau4\Delta_t(\bm{p}_t).
\end{align*}
Combining the above two cases and setting $\eta=\sqrt{\frac{C_1}{C_2T}}$, together with Lemma~\ref{le:ucb00}, we get that
\begin{align*}
	{\widetilde{\mathcal R}}&\leq \left(\|\bm{A}\|_{\infty}\bar d/\underline{\gamma}\right)[r(\bm{p}^*)+\phi(\bm{A}\bm{\bm{D}}(\bm{p}^*))]+\mathcal{O}(\|\bm{A}\|_{\infty}\bar d\sqrt{T\log (MT)})\\
	&\qquad+2\sqrt{C_1 C_2 T}+ \O\left(\sqrt{N+1}\kappa\max\{\bar p,1\}(N\bar p+\max\{C\|\bm{A}\|_\infty, 1\}) \right)\times\sqrt{NT\log(N+1+\bar p^2T)}.
\end{align*}
Together with Eq.~\eqref{eq:thm1jenson} and the discussion about the rare case when either of the desired events of Corollary~\ref{cor:check-error-bound} and Lemma~\ref{le:concentration} fails, we conclude that 
\begin{align*}
	\mathcal R(T) & \leq \left(\|\bm{A}\|_{\infty}\bar d/\underline{\gamma} + \mathcal{O}(1)\right)[r(\bm{p}^*)+\phi(\bm{A}\bm{\bm{D}}(\bm{p}^*))]+ \mathcal{O}(\|\bm{A}\|_{\infty}\bar d\sqrt{T\log (MT)})\\
	&+2\sqrt{C_1 C_2 T}+ \O\left(\sqrt{N+1}\kappa\max\{\bar p,1\}(N\bar p+\max\{C\|\bm{A}\|_\infty, 1\}) \right)\times\sqrt{NT\log(N+1+\bar p^2T)}. \Halmos
\end{align*}

\section{Proofs Omitted in Section~\ref{sec:regretanalysis}}
\subsection{Proof of Lemma~\ref{le:concentration}}
\proof{Proof of Lemma~\ref{le:concentration}.}

 Combining the definition of $\|\bm{A}\|_\infty$ and the boundedness of $\bm{d}_t$, we have $$\|\bm{A}\bm{\varepsilon}_t\|_\infty\leq\|\bm{A}\|_\infty\|\bm \varepsilon_t\|_\infty\leq\|\bm{A}\|_\infty\bar d.$$

By the assumption on the demand noise, $\{\bm \varepsilon_t\}_{t=1}^T$ is martingale difference sequence, so as $\{[\bm{A}\bm \varepsilon_t]_i\}_{t=1}^T$. Therefore, applying Azuma-Hoeffding's inequality (Lemma \ref{le:azuma}) , for all $i\in[M]$ with probability $1-1/(MT)$ we have that
\begin{equation}\label{eq:hoefding}
    \left|\sum_{t=1}^T [\bm{A}\bm{d}_t]_i-\sum_{t=1}^T [\bm{A}\bm D(\bm{p}_t)]_i\right|\leq \mathcal{O}(\|\bm{A}\|_{\infty}\bar d\sqrt{T\log(MT)}).
\end{equation}

By a union bound, we have that with probability at least $(1-1/T)$, Eq.~\eqref{eq:hoefding} holds for every $i \in [M]$, proving the lemma. \Halmos
\endproof

\subsection{Proof of Claim~\ref{claim:rplushphi-pf-0}}\label{sec:proofofclaim}
\proof{Proof of Claim~\ref{claim:rplushphi-pf-0}.}
Recall that the goal is to prove
\begin{align}
r(\bm{p}^*)+\phi\left(\bm{A}\bm{D}(\bm{p}^*)\right) \leq r(\bm{p}_t^*) + \phi(\bm{s}_t) + \langle \bm{\mu}_t, -\bm{A}\bm{D}(\bm{p}_t^*) +\bm{s}_t\rangle . \label{eq:rplushphi-pf-0}
\end{align}
By Eq.~\eqref{eq:rsharp} and the definitions of $p^*_t$ and $\bm s_t$ (Eq.~\eqref{eq:primal-update}), it holds that
\begin{align*}
r\left(\bm p_{t}^{*}\right)& = \max_{\bm{p}\in[\underline{p},\overline{p}]^N} \left\{ r(\bm{p})- \langle \bm{A}^\top \bm{\mu}_t, \bm D(\bm{p})\rangle \right\} =
r^\sharp\left(\bm{A}^\top \bm{\mu}_{t}\right) + \left\langle\bm{\mu}_{t},
 \bm A\bm D(\bm{p}_{t}^{*})\right\rangle,\\
\phi\left(\bm{s}_{t}\right) + \left\langle\bm{\mu}_{t}, \bm s_{t}\right\rangle & =\max_{-\bm{\gamma} \leq \bm s \leq \bm{\gamma}} \{\phi(\bm{s}) + \langle \bm{\mu}_t, \bm s \rangle\} = (-\phi)^*\left(\bm{\mu}_{t}\right),
\end{align*}
which leads to 
\begin{align}
r(\bm{p}_t^*) + \phi(\bm{s}_t) & = r^\sharp\left(\bm{A}^\top \bm{\mu}_{t}\right) +  (-\phi)^*\left(\bm{\mu}_{t}\right) + \left\langle\bm{\mu}_{t} ,  \bm A\bm D(\bm{p}_{t}^{*})\right\rangle  - \left\langle\bm{\mu}_{t}, \bm s_{t}\right\rangle \nonumber\\
& = \mathfrak{q}(\bm{\mu}_t) +\left\langle\bm{\mu}_{t} ,  \bm A\bm D(\bm{p}_{t}^{*}) - \bm s_t\right\rangle \geq \mathfrak{p}^* +\left\langle\bm{\mu}_{t},  \bm A\bm D(\bm{p}_{t}^{*}) - \bm s_t\right\rangle, \label{eq:rplushphi-pf}
\end{align}
where the second equality is by the definition of $\mathfrak{q}$ (Eq.~\eqref{eq:def-dual-q}) and the last inequality is due to the weak duality (Eq.~\eqref{eq:weak-duality-2}). Combining Eq.~\eqref{eq:rplushphi-pf} and the definition of $\mathfrak{p}^*$  (Eq.~\eqref{eq:primal-opt-def-1}), we prove Eq.~\eqref{eq:rplushphi-pf-0}.\Halmos
\endproof

\subsection{Proof of Lemma~\ref{le:duala}}
\proof{Proof of Lemma~\ref{le:duala}.}
We start by proving Eq.~\eqref{eq:duala}. By the definition that $\bar{\bm s} =\frac{1}{T} \sum_{t=1}^{T} \bm{A} \bm D(\bm{p}_t)$ and $\bar{\bm{\mu}} =\argmax _{  \bm{\mu} \in \mathbb{R}^{M}}\{-(-\phi)^*(\bm{\mu})+\langle\bm{\mu}, \bar{\bm s}\rangle\}$, we have that
\[
-(-\phi)^*(\bar{\bm{\mu}})+\langle\bar{\bm{\mu}}, \bar{\bm s}\rangle=\max _{  \bm{\mu} \in \mathbb{R}^{M}}\left\{-(-\phi)^*(\bm{\mu})+\langle\bm{\mu}, \bar{\bm s}\rangle\right\}.
\]

By Assumption~\ref{assump:phi1}, we have that $-\phi(\bm{s}) $ is convex and closed with the closed domain $\{\bm s : -\bm{\gamma} \leq \bm s \leq \bm{\gamma}\}$ (since $\phi(\cdot)$ is continuous), which implies that 
\[
(-\phi)^{**}(\bm s) = -\phi(\bm{s}) \qquad \forall \bm s : -\bm{\gamma} \leq \bm s \leq \bm{\gamma}.
\]
Thus, for $\bar{\bm s} \in \{\bm s : -\bm{\gamma} \leq \bm s \leq \bm{\gamma}\}$,  it holds that
\begin{align}\label{eq:conjugateof}
-\phi(\bar{\bm s}) = \max_{\bm{\mu} \in \mathbb{R}^M} \{ -(-\phi)^{*}(\bm{\mu}) + \langle \bm{\mu}, \bar{\bm s} \rangle  \} = \max_{\bm{\mu} \in \mathbb{R}^M} \{ -(-\phi)^{*}(-\bm{\mu}) + \langle \bm{\mu}, \bar{\bm s} \rangle  \} = -(-\phi)^*(\bar{\bm{\mu}})+\langle\bar{\bm{\mu}}, \bar{\bm s}\rangle
\end{align}

Let $\widetilde{\bm{s}}=\frac{1}{T} \sum_{t=1}^{T} \bm s_{t}$. Recall the definition of $(-\phi)^*(\cdot)$ (Eq.~\eqref{eq:dual-functions}), we obtain that
\begin{align*}
(-\phi)^*(\bar{\bm{\mu}}) &=\max_{-\bm{\gamma}\leq\bm{s}\leq \bm{\gamma} }\{\phi(\bm{s})+\langle\bar{\bm{\mu}}, \bm s\rangle\} 
\geq \frac{1}{T} \sum_{t=1}^{T} (\phi\left(\bm{s}_{t}\right)+\left\langle\bar{\bm{\mu}},\bm s_{t}\right\rangle) 
=  \frac{1}{T} \sum_{t=1}^{T}( \phi\left(\bm{s}_{t}\right)+\left\langle\bar{\bm{\mu}},\widetilde{\bm{s}}\right\rangle).
\end{align*}

Combining the two equations above, we have that
\begin{align}\label{eq:pf-le-duala-1}
\phi(\bar{\bm s})\geq \frac{1}{T} \sum_{t=1}^{T} \phi\left(\bm{s}_{t}\right)+T\left\langle\bar{\bm{\mu}},\widetilde{\bm{s}}-\bar{\bm s}\right\rangle.
\end{align}

Finally, by the definition of $\bm g_t$ (Eq.~\eqref{eq:def-g-t}), we have that 
$T\left\langle\bar{\bm{\mu}},\widetilde{\bm{s}}-\bar{\bm s}\right\rangle = \sum_{t=1}^T\left\langle\bar{\bm{\mu}},\bm{g}_{t}\right\rangle$.
Together with Eq.~\eqref{eq:pf-le-duala-1}, we prove Eq.~\eqref{eq:duala}.

\medskip

We now turn to show that $\bar{\bm{\mu}} \in \D$. By the definition of $(-\phi)^*(\cdot)$ (Eq.~\eqref{eq:dual-functions}), for all $\bm s:  -\bm{\gamma}\leq\bm{s}\leq \bm{\gamma} $, we have that 
\[
(-\phi)^*(\bar{\bm{\mu}})\geq\phi\left(\bm s\right)+\left\langle\bar{\bm{\mu}}, \bm s\right\rangle.
\]
Together with Eq.~(\ref{eq:conjugateof}), for all $\bar{\bm{s}} \in \{\bm s :  -\bm{\gamma}\leq\bm{s}\leq \bm{\gamma}\}$, we have that
\begin{equation}\label{eq:phibars}
    \phi(\bar{\bm{s}})\geq \phi(\bm{s})+\left\langle\bar{\bm{\mu}},\bm s-\bar{\bm s}\right\rangle.
\end{equation}

By the definition of the dual norm, it holds that 
   $$ \|\bar{\bm{\mu}}\|_1=\max_{\bm v:\|\bm v\|_{\infty}=1}\Braket{\bar{\bm{\mu}},\bm v}.$$
   
Let $\bm v^*=\argmax_{\bm v:\|\bm v\|_{\infty}=1}\Braket{\bar{\bm{\mu}},\bm v}$.  Since $\bar{\bm{s}} $ is an interior point of $[-\bm{\gamma},\bm{\gamma}]$, there exists a small real number ${\alpha}$ such that $\bar{\bm{s}}+{\alpha} \bm v^*\in [-\bm{\gamma},\bm{\gamma}]$. Plug $\bm s = \bar{\bm{s}}+{\alpha} \bm v^*$ into Eq.~(\ref{eq:phibars}), we obtain that
\[
\phi(\bar{\bm{s}})- \phi(\bar{\bm{s}}+{\alpha} \bm v^*) \geq \left\langle\bar{\bm{\mu}},{\alpha} \bm v^*\right\rangle ={\alpha} \|\bar{\bm{\mu}}\|_1.
\]


By Assumption~\ref{assump:phi1}, we have that $\phi(\cdot)$ is $L$-Lipschitz continuous with respect to the $\|\cdot\|_{\infty}$-norm. Therefore, we have $\|\bar{\bm{\mu}}\|_1\leq L$, thus $\bar{\bm{\mu}} \in \D = \left\{  \bm{\mu} \in \mathbb{R}^{M} \mid \|\bm{\mu}\|_1\leq C \right\}$ (since $C\geq L$).  \Halmos
\endproof
 
\section{Additional Numerical Experiments}\label{sec:addexp}

In this section, we present additional numerical results, which include 1) computational time report; 2) numerical results of Algorithm~\ref{algucb}+$\EG^\pm$ under the inventory level $\bm{\gamma}_2$; 3) numerical results of Algorithm~\ref{algucb}+$\mathrm{PGD}$; 4) performance Comparison of $\EG^\pm$ and $\mathrm{PGD}$; 5) numerical results of Algorithm~\ref{algucb}+$\mathrm{PGD}$ in a model misspecified setting;  6) numerical results on a classic NRM example studied in \cite{besbes2012blind,ferreira2018online}.
\subsection{Computational Time Report}
In Table~\ref{tab:time}, we report the running time of Algorithm~\ref{algucb}+$\EG^\pm$. 
Running time is computed by Python 3.9 on a PC with Intel-Core-i7 CPU under the inventory level $\bm \gamma_2$ and regularization level $\lambda = 0.5$. 

We conduct $10$ trials independently and report the average running time of these trials. The numbers in parentheses refer to  the $95\%$ confidence error across the $10$ trials.
\begin{table}[h!]
\centering
\caption{ {Runtime (in seconds) of Algorithm~\ref{algucb}+$\EG^\pm$ under the inventory level $\bm \gamma_2$ and regularization level $\lambda = 0.5$.}}\label{tab:time}
\begin{tabular}{l|rrrrrr}
\toprule
Time Horizon & 100     & 2000    & 4000  & 6000    & 8000 & 10000  \\
\hline
Running Time      &2.41 (0.05)  & 11.97 (0.59)  & 20.51 (0.85) & 27.25 (0.42) & 34.63 (0.57) & 41.20 (0.88)\\
\bottomrule
\end{tabular}
\end{table}

\subsection{Numerical results of Algorithm~\ref{algucb}+$\EG^\pm$ under inventory level $\bm{\gamma}_2$}\label{sec:eginventorygamma2}

\begin{figure}[!h]
\centering
\includegraphics[width =0.4\textwidth]{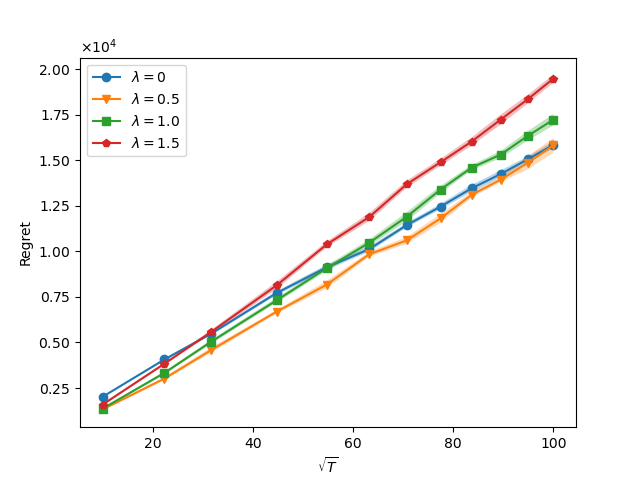}
\includegraphics[width =0.4\textwidth]{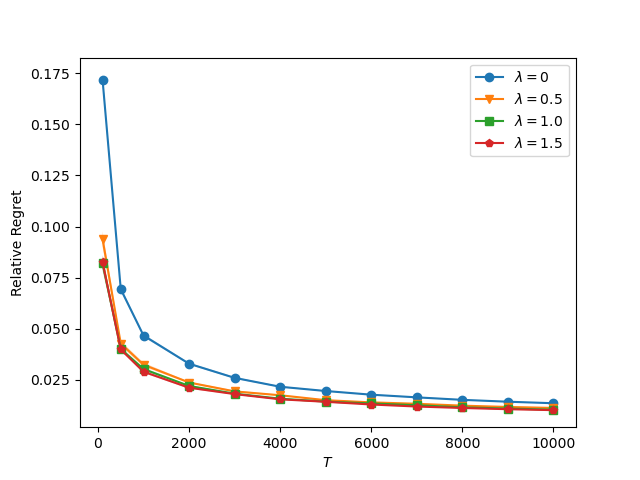}
\caption{ {The performance of Algorithm~\ref{algucb}+$\EG^\pm$ with the inventory level $\bm{\gamma}_2$ and $\lambda \in\{ 0, 0.5,1.0, 1.5\}$.  Here the $x$-axis of the left figure is the square root of the total time periods $T$ and the $y$-axis is the cumulative regret defined in Eq.~(\ref{regretform2}). The $x$-axis of the right figure is the total time periods $T$ and the $y$-axis is the relative regret defined in \eq{\ref{eq:relativeregret}}.}}
\label{fig:linearegregrethigh}
\end{figure}
\begin{figure}[!h]
\centering
\includegraphics[width =0.4\textwidth]{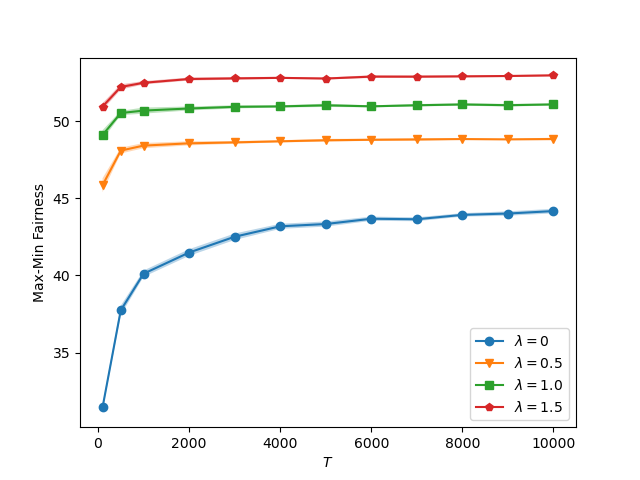}
\includegraphics[width =0.4\textwidth]{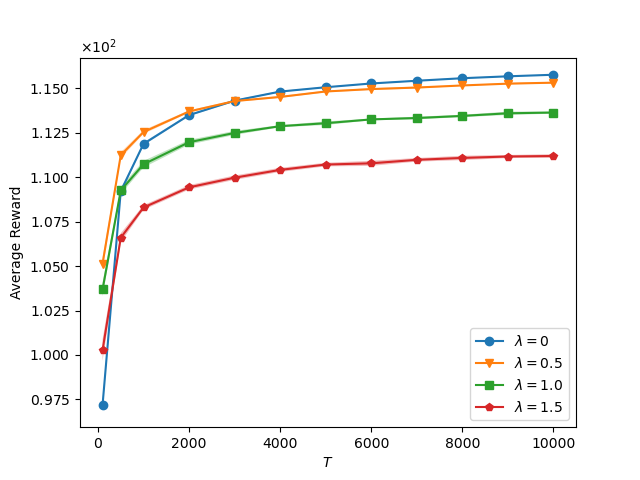}
\caption{ {The max-min fairness $\min_i\left(\frac{1}{T}\sum_{t=1}^T[\bm{A}\bm{d}_t]_i\right)$ and the average reward $\frac{1}{T}\sum_{t=1}^Tr(\bm{p}_t)$ of Algorithm~\ref{algucb}+$\EG^\pm$ at regularization levels $\lambda \in\{ 0, 0.5,1.0, 1.5\}$ under the initial inventory level $\bm{\gamma}_2$.  }}
\label{fig:linearegfairnesshigh}
\end{figure}
\noindent \underline{\bf Results.} In the left of Figure~\ref{fig:linearegregrethigh} is the plot of the regret of Algorithm~\ref{algucb} with the inventory level $\bm{\gamma}_2$  and $\lambda \in\{ 0, 0.5,1.0, 1.5\}$ versus the square root of the total time periods $T$.  In the right of Figure~\ref{fig:linearegregrethigh} we plot the relative regret of Algorithm~\ref{algucb} versus the total time periods $T$.

In the left of Figure~\ref{fig:linearegfairnesshigh}  is the plot of the max-min fairness versus the total time periods $T$ with $\bm{\gamma_2}$ and $\lambda \in\{0, 0.5,1.0, 1.5\}$.
In the right of Figure~\ref{fig:linearegfairnesshigh}  we plot the average reward versus the total time periods $T$. 

It is easy to find that the numerical results of the initial inventory level $\bm{\gamma}_2$ are almost the same as those in the case $\bm{\gamma}_1$ presented in Section~\ref{sec:experiments}, which justifies the effectiveness of our algorithm for different initial inventory levels.

\subsection{Numerical results of Algorithm~\ref{algucb}+$\mathrm{PGD}$}\label{sec:expofpgd}
It is worth noting that when $h(\bm x)=\frac{1}{2}\|\bm{x}\|_2^2$,  the online mirror descent algorithm \eqref{eq:OMD-update} is known as the projected gradient descent method (PGD), which also satisfies Definition~\ref{def:dualdescent}.
In this section, we present the numerical results of Algorithm~\ref{algucb}+$\mathrm{PGD}$.
\begin{figure}[!h]
\centering
\includegraphics[width =0.4\textwidth]{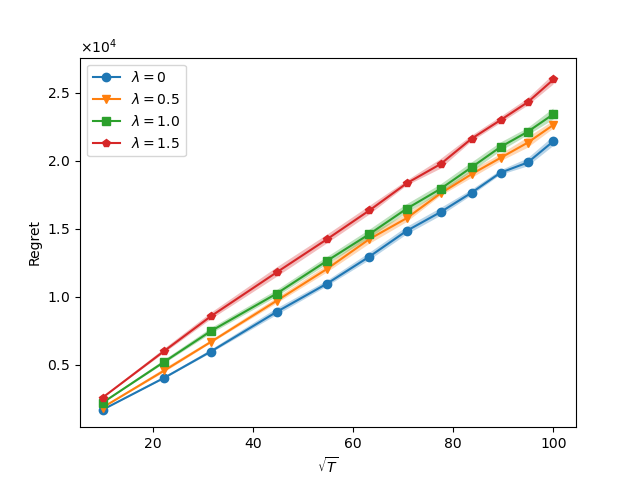}
\includegraphics[width =0.4\textwidth]{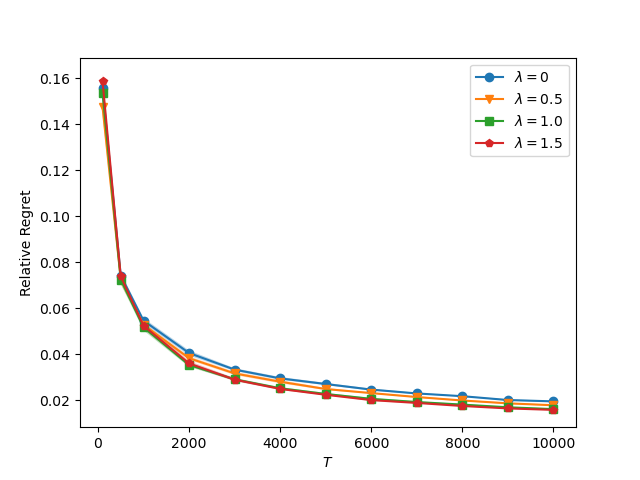}
\caption{ {The performance of Algorithm~\ref{algucb}+$\mathrm{PGD}$ with $\bm{\gamma}_1$ and $\lambda \in\{ 0, 0.5,1.0, 1.5\}$.  Here the $x$-axis of the left figure is the square root of the total time periods $T$ and the $y$-axis is the cumulative regret defined in Eq.~(\ref{regretform2}). The $x$-axis of the right figure is the total time periods $T$ and the $y$-axis is the relative regret defined in \eq{\ref{eq:relativeregret}}.}}
\label{fig:linearpgdregretlow}
\end{figure}
\begin{figure}[!h]
\centering
\includegraphics[width =0.4\textwidth]{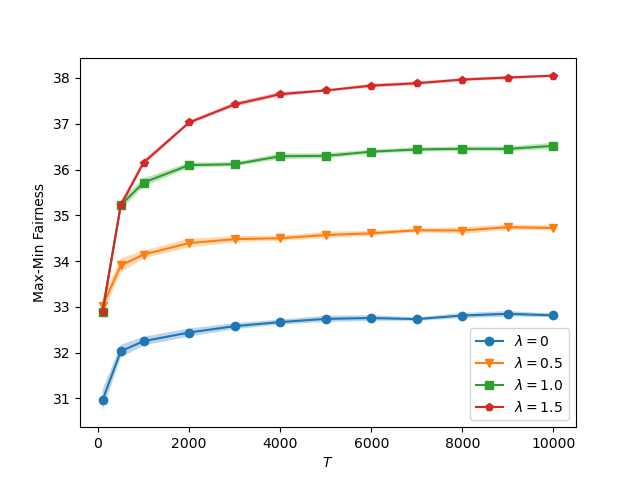}
\includegraphics[width =0.4\textwidth]{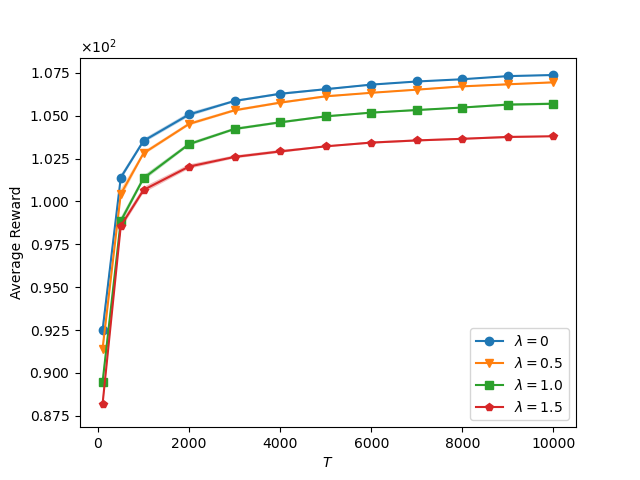}
\caption{ {The max-min fairness $\min_i\left(\frac{1}{T}\sum_{t=1}^T[\bm{A}\bm{d}_t]_i\right)$ and the average reward $\frac{1}{T}\sum_{t=1}^Tr(\bm{p}_t)$ of Algorithm~\ref{algucb}+$\mathrm{PGD}$ at regularization levels $\lambda \in\{ 0, 0.5,1.0, 1.5\}$ under the initial inventory level $\bm{\gamma}_1$.  }}
\label{fig:linearpgdfairnesslow}
\end{figure}

\begin{figure}[t]
\centering
\includegraphics[width =0.4\textwidth]{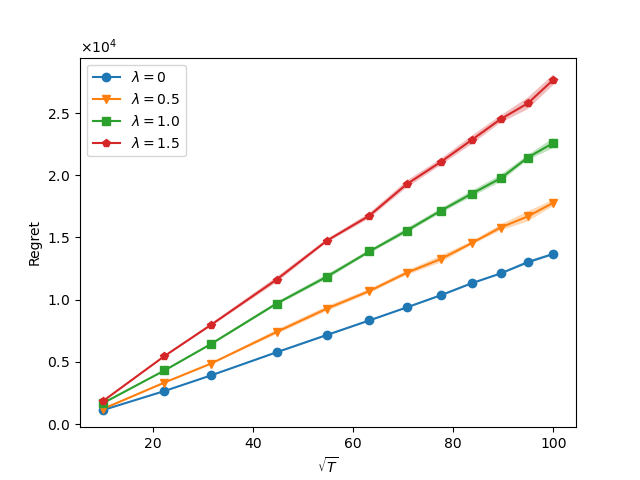}
\includegraphics[width =0.4\textwidth]{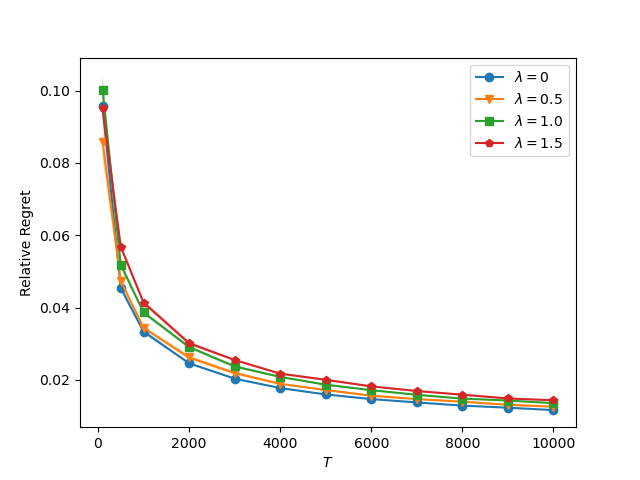}
\caption{ {The performance of Algorithm~\ref{algucb}+$\mathrm{PGD}$ with the inventory level $\bm{\gamma}_2$ and $\lambda \in\{ 0, 0.5,1.0, 1.5\}$.  Here the $x$-axis of the left figure is the square root of the total time periods $T$ and the $y$-axis is the cumulative regret defined in Eq.~(\ref{regretform2}). The $x$-axis of the right figure is the total time periods $T$ and the $y$-axis is the relative regret defined in \eq{\ref{eq:relativeregret}}.}}
\label{fig:linearpgdregrethigh}
\end{figure}
\begin{figure}[!h]
\centering
\includegraphics[width =0.4\textwidth]{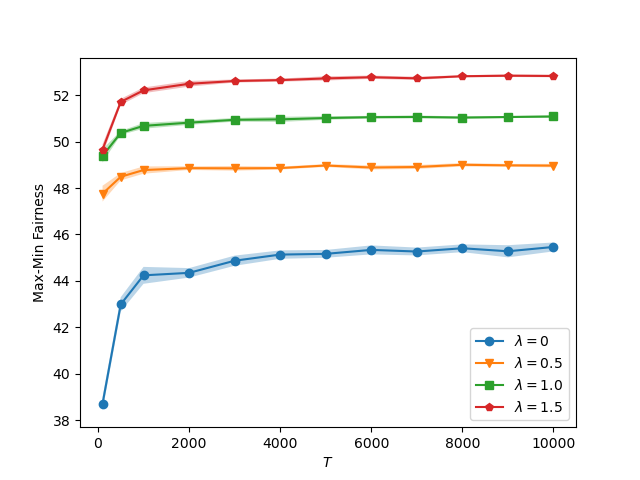}
\includegraphics[width =0.4\textwidth]{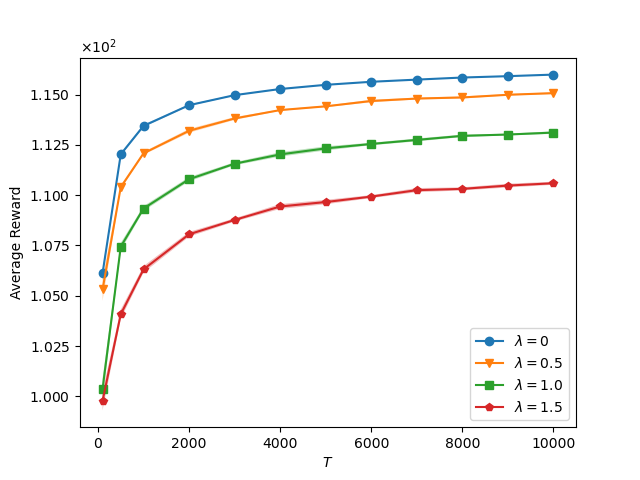}
\caption{ {The max-min fairness $\min_i\left(\frac{1}{T}\sum_{t=1}^T[\bm{A}\bm{d}_t]_i\right)$ and the average reward $\frac{1}{T}\sum_{t=1}^Tr(\bm{p}_t)$ of Algorithm~\ref{algucb}+$\mathrm{PGD}$ at regularization levels $\lambda \in\{ 0, 0.5,1.0, 1.5\}$ under the initial inventory level $\bm{\gamma}_2$.  }}
\label{fig:linearpgdfairnesshigh}
\end{figure}

\noindent \underline{\bf Results.} In the left of Figure~\ref{fig:linearpgdregretlow} (resp. Figure~\ref{fig:linearpgdregrethigh}) is the plot of the regret of Algorithm~\ref{algucb} with the inventory level $\bm{\gamma}_1$  (resp. $\bm{\gamma}_2$) and $\lambda \in\{ 0, 0.5,1.0, 1.5\}$ versus the square root of the total time periods $T$.  In the right of Figure~\ref{fig:linearpgdregretlow} (resp. Figure~\ref{fig:linearpgdregrethigh}) we plot the relative regret of Algorithm~\ref{algucb} versus the total time periods $T$.

In the left of Figure~\ref{fig:linearpgdfairnesslow} (resp. Figure~\ref{fig:linearpgdfairnesshigh}) is the plot of the max-min fairness versus the total time periods $T$ with $\bm{\gamma}_1$  (resp. $\bm{\gamma}_2$) and $\lambda \in\{0, 0.5,1.0, 1.5\}$.
In the right of Figure~\ref{fig:linearpgdfairnesslow} (resp. Figure~\ref{fig:linearpgdfairnesshigh}) we plot the average reward versus the total time periods $T$. 

\subsection{Performance Comparison between $\EG^\pm$ and $\mathrm{PGD}$.}\label{sec:expcomparison}
In this section, we compare the regret performance between Algorithm~\ref{algucb}+$\mathrm{PGD}$ and Algorithm~\ref{algucb}+$\EG^\pm$ under  two initial inventory levels ($\bm{\gamma}_1$ and $\bm{\gamma}_2$) and four  regularization level ($\lambda \in \{0, 0.5, 1.0, 1.5\}$). 

\noindent \underline{\bf Results.} Figure~\ref{fig:comparisonlow} (Figure~\ref{fig:comparisonhigh})
is the regret comparison between $\EG^\pm$ and $\mathrm{PGD}$ under the initial inventory level  $\bm{\gamma}_1$  (resp. $\bm{\gamma}_2$) and four  regularization level ($\lambda \in \{0, 0.5, 1.0, 1.5\}$. Here the $x$-axis of the left figure is the square root of the total time periods $T$ and the $y$-axis is the cumulative regret defined in Eq.~(\ref{regretform2}).

From these comparisons, it is easy to note that the empirical performance of Algorithm~\ref{algucb}+$\EG^\pm$ is better than Algorithm~\ref{algucb}+$\mathrm{PGD}$.
\begin{figure}[!h]
\centering
\includegraphics[width =0.24\textwidth]{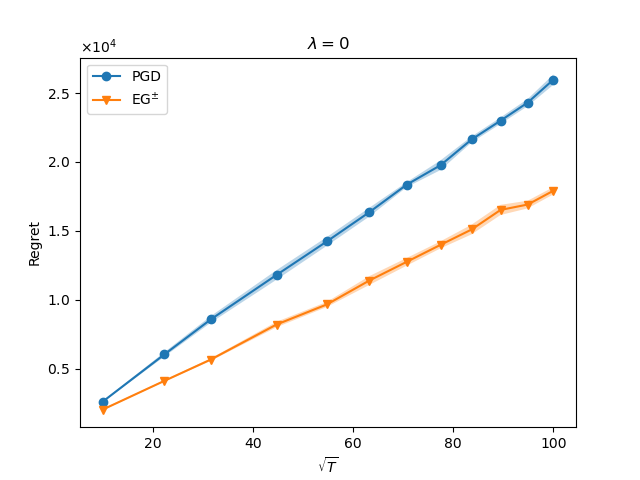}
\includegraphics[width =0.24\textwidth]{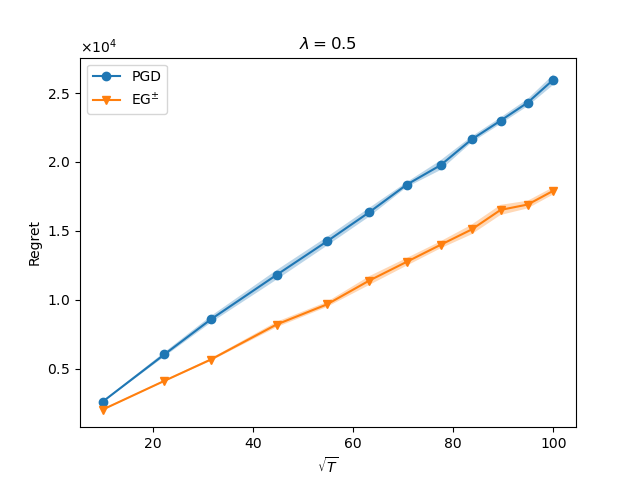}
\includegraphics[width =0.24\textwidth]{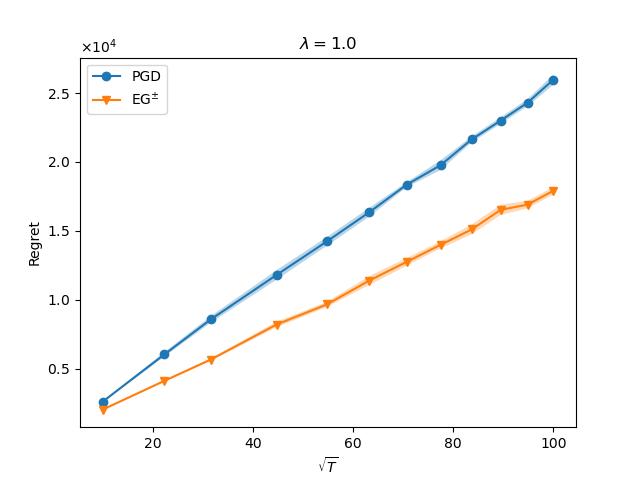}
\includegraphics[width =0.24\textwidth]{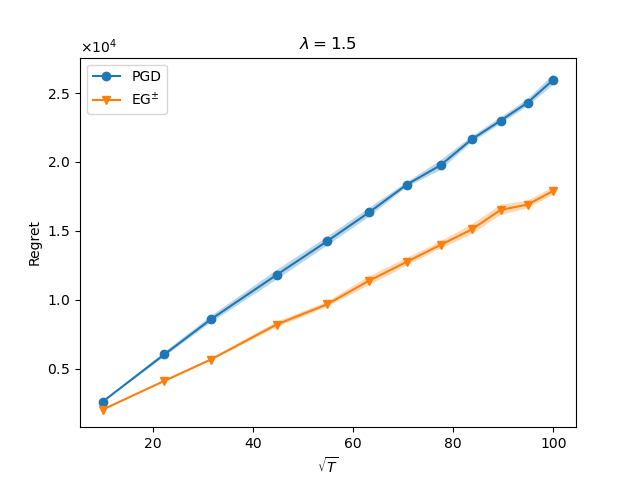}
\caption{ {Performance comparison of $\EG^\pm$ and $\mathrm{PGD}$ under the initial inventory levels $\bm{\gamma}_1$ and four  regularization level ($\lambda \in \{0, 0.5, 1.0, 1.5\}$. }}
\label{fig:comparisonlow}
\includegraphics[width =0.24\textwidth]{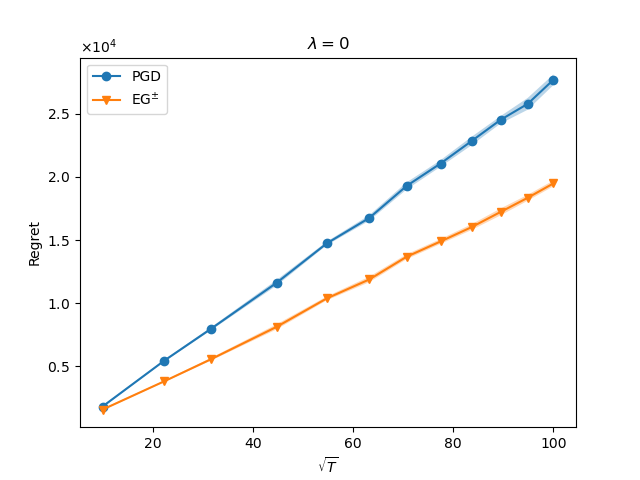}
\includegraphics[width =0.24\textwidth]{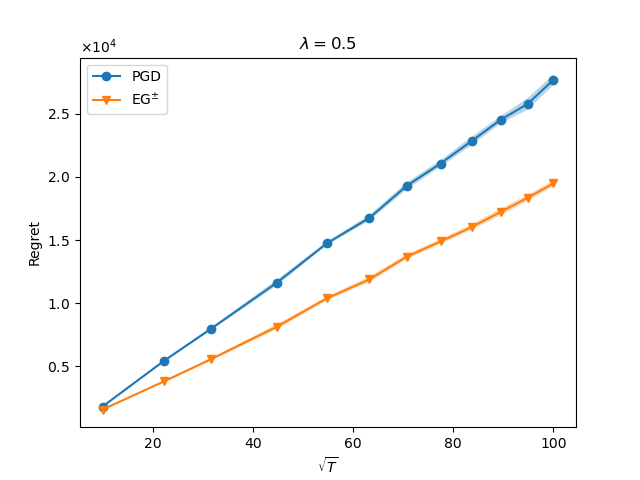}
\includegraphics[width =0.24\textwidth]{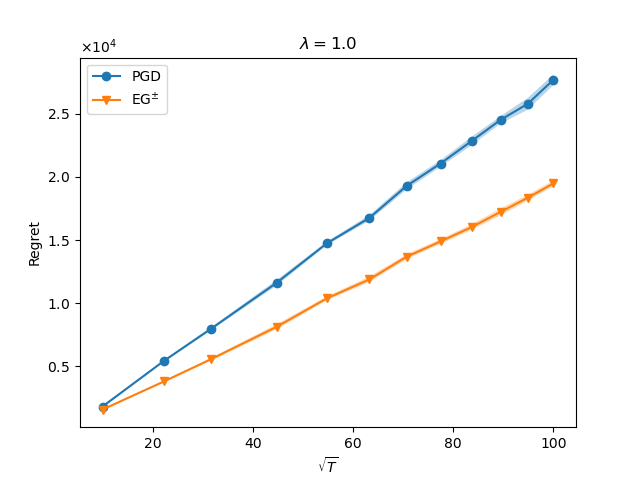}
\includegraphics[width =0.24\textwidth]{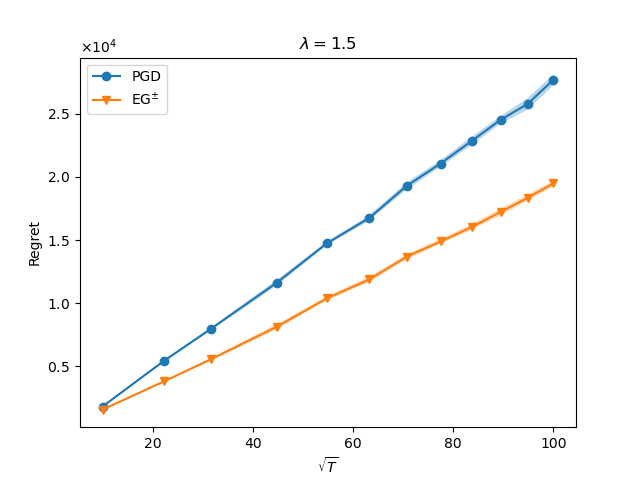}
\caption{ {Performance comparison of $\EG^\pm$ and $\mathrm{PGD}$ under the initial inventory levels $\bm{\gamma}_2$ and four  regularization level ($\lambda \in \{0, 0.5, 1.0, 1.5\}$. }}
\label{fig:comparisonhigh}
\end{figure}

\subsection{Model Misspecified Setting}\label{sec:expmisspecified}
\begin{figure}[t]
\centering
\includegraphics[width =0.4\textwidth]{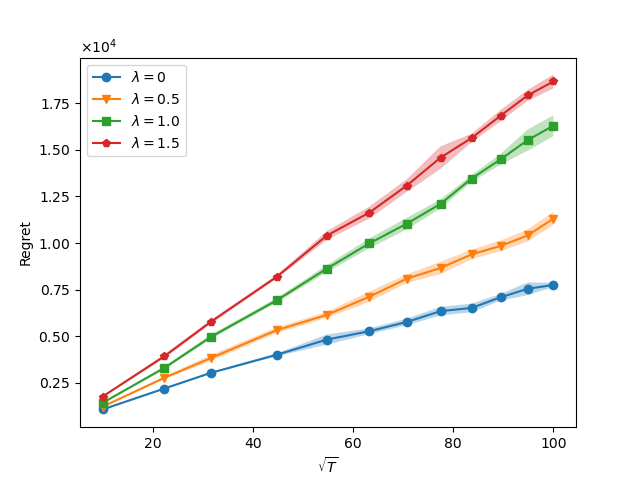}
\includegraphics[width =0.4\textwidth]{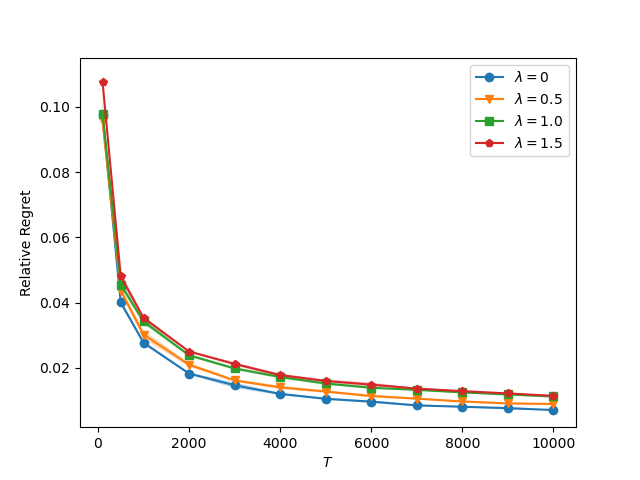}
\caption{ {The performance of Algorithm~\ref{algucb}+$\EG^\pm$ with $\bm{\gamma}_1$ and $\lambda \in\{ 0, 0.5,1.0, 1.5\}$ in the model misspecified setting.  Here the $x$-axis of the left figure is the square root of the total time periods $T$ and the $y$-axis is the cumulative regret defined in Eq.~(\ref{regretform2}). The $x$-axis of the right figure is the total time periods $T$ and the $y$-axis is the relative regret defined in \eq{\ref{eq:relativeregret}}.}}
\label{fig:expegregretlow}
\end{figure}
\begin{figure}[!h]
\centering
\includegraphics[width =0.4\textwidth]{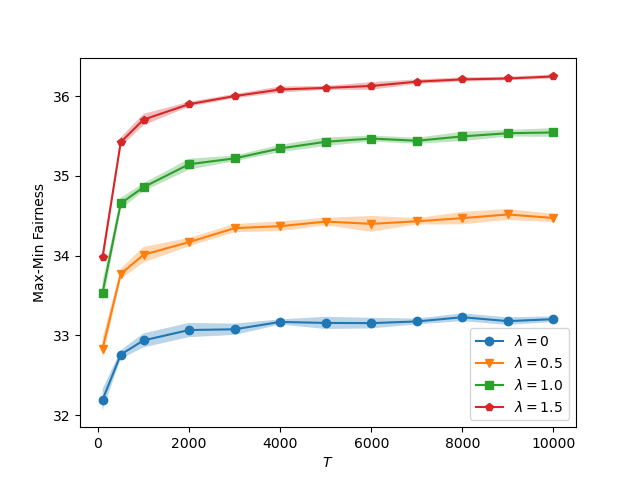}
\includegraphics[width =0.4\textwidth]{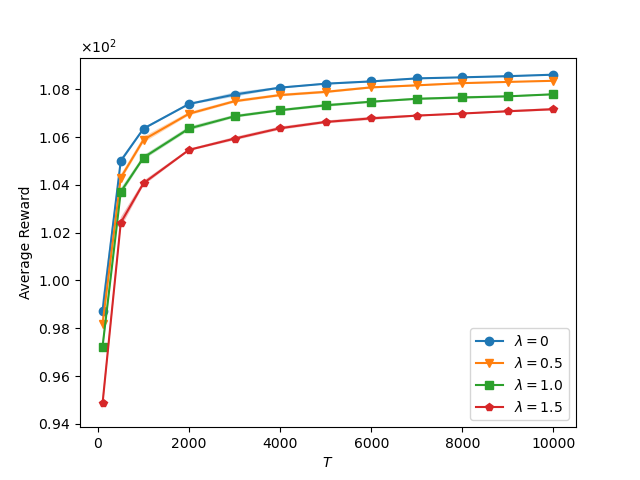}
\caption{ {The max-min fairness $\min_i\left(\frac{1}{T}\sum_{t=1}^T[\bm{A}\bm{d}_t]_i\right)$ and the average reward $\frac{1}{T}\sum_{t=1}^Tr(\bm{p}_t)$ of Algorithm~\ref{algucb}+$\EG^\pm$ at regularization levels $\lambda \in\{ 0, 0.5,1.0, 1.5\}$ under the initial inventory level $\bm{\gamma}_1$ in the model misspecified setting.  }}
\label{fig:expegfairnesslow}
\end{figure}

\begin{figure}[!h]
\centering
\includegraphics[width =0.4\textwidth]{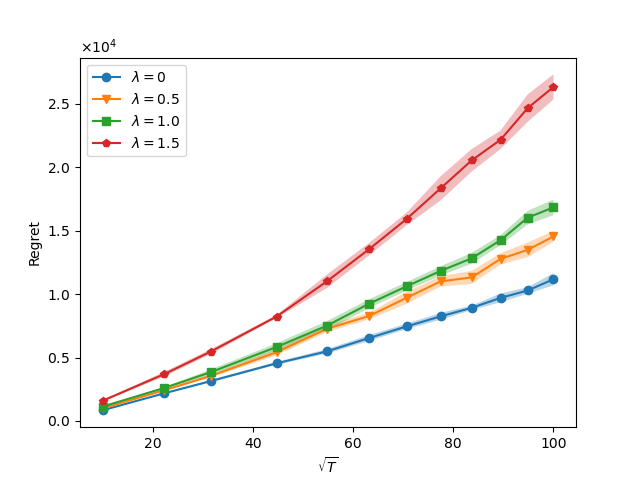}
\includegraphics[width =0.4\textwidth]{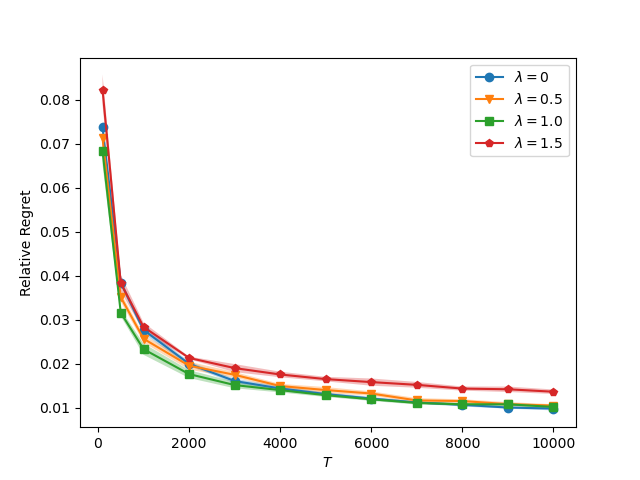}
\caption{ The performance of Algorithm~\ref{algucb}+$\EG^\pm$ with the inventory level $\bm{\gamma}_2$ and $\lambda \in\{ 0, 0.5,1.0, 1.5\}$ in the model misspecified setting.  Here the $x$-axis of the left figure is the square root of the total time periods $T$ and the $y$-axis is the cumulative regret defined in Eq.~(\ref{regretform2}). The $x$-axis of the right figure is the total time periods $T$ and the $y$-axis is the relative regret defined in \eq{\ref{eq:relativeregret}}.}
\label{fig:expegregrethigh}
\end{figure}
\begin{figure}[!h]
\centering
\includegraphics[width =0.4\textwidth]{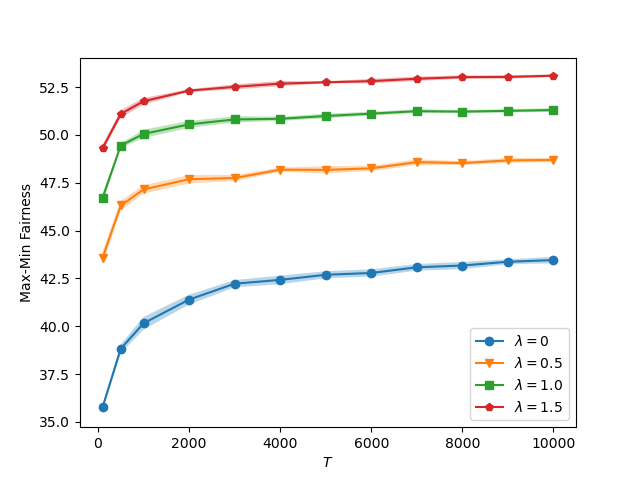}
\includegraphics[width =0.4\textwidth]{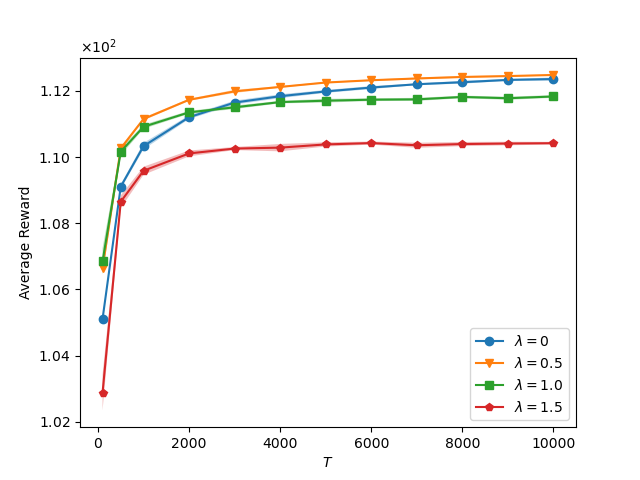}
\caption{ {The max-min fairness $\min_i\left(\frac{1}{T}\sum_{t=1}^T[\bm{A}\bm{d}_t]_i\right)$ and the average reward $\frac{1}{T}\sum_{t=1}^Tr(\bm{p}_t)$ of Algorithm~\ref{algucb}+$\EG^\pm$ at regularization levels $\lambda \in\{ 0, 0.5,1.0, 1.5\}$ under the initial inventory level $\bm{\gamma}_2$ in the model misspecified setting.  }}
\label{fig:expegfairnesshigh}
\end{figure}

We also conduct Algorithm~\ref{algucb}+ $\EG^\pm$ in a model misspecified setting, where the true demand function is an exponential function.  We use this instance to illustrate the robustness of our algorithm when the model assumptions are not satisfied.
The resource consumption matrix of the NRM example is defined the same as Eq.~\eqref{eq:resource matrix}, and the underlying exponential demand function is defined as \[
\bm D(\bm{p}) =  \mathrm{{exp}}\left(\begin{bmatrix}
3.3 \\
3.5 \\
3.2 \\
2.9 \\
3.5 \\
\end{bmatrix} +  \begin{bmatrix}
-0.4 & 0 & 0 & 0 & 0 \\
0 & -0.35 & 0 & 0 & 0 \\
0 & 0 & -0.45 & 0 & 0 \\
0 & 0 & 0 & -0.5 & 0 \\
0 & 0 & 0 & 0 & -0.6 \\
\end{bmatrix}p\right).
\]

\noindent \underline{\bf Results.} In the left of Figure~\ref{fig:expegregretlow} (resp. Figure~\ref{fig:expegregrethigh}) is the plot of the regret of Algorithm~\ref{algucb} with the inventory level $\bm{\gamma}_1$  (resp. $\bm{\gamma}_2$) and $\lambda \in\{ 0, 0.5,1.0, 1.5\}$ versus the square root of the total time periods $T$.  In the right of Figure~\ref{fig:expegregretlow} (resp. Figure~\ref{fig:expegregrethigh}) we plot the relative regret of Algorithm~\ref{algucb} versus the total time periods $T$.

In the left of Figure~\ref{fig:expegfairnesslow} (resp. Figure~\ref{fig:expegfairnesshigh}) is the plot of the max-min fairness versus the total time periods $T$ with $\bm{\gamma}_1$  (resp. $\bm{\gamma}_2$) and $\lambda \in\{0, 0.5,1.0, 1.5\}$.
In the right of Figure~\ref{fig:expegfairnesslow} (resp. Figure~\ref{fig:expegfairnesshigh}) we plot the average reward versus the total time periods $T$. 

\subsection{Experiments on another Classic NRM Example}\label{sec:expclassic}
 For consistency, in this section, we use the NRM example presented in \cite{besbes2012blind,ferreira2018online}. In this example, the retailer sells two products ($N=2$) using three resources ($M=3$), and the resource consumption matrix is defined as 
\setlength{\arraycolsep}{8pt}
\renewcommand{\arraystretch}{1.5}
\[
   \bm A =
  \begin{bmatrix}
  1 & 1\\
  3 & 1\\0 & 5
  \end{bmatrix}.
\]
The underlying linear demand function is defined as 
\begin{align}\label{eq:linear demand}
\bm D(\bm{p}) =   \begin{bmatrix}
   8\\
      9
  \end{bmatrix} +  \begin{bmatrix}
  -1.5 & 0\\
     0 & -3
  \end{bmatrix}\bm p.
\end{align}

In contrast to \cite{besbes2012blind,ferreira2018online} which use a discrete price set in their experiments, we use the continuous price set to test the effectiveness of our algorithm for handling large price sets. 
We assume that the price range for each product is $[1,5]$. 

In addition, we choose the weighted min-max fairness regularizer 
\[
\phi(\bm{s}):=\lambda\min_i(w_is_i)
\]
with $w_i = 1$ for all $i$. 
 We implement Algorithm~\ref{algucb} with $\EG^\pm$ solver, and test two initial inventory levels ($\bm{\gamma} = (10,8,20)$ and $\bm{\gamma}=(15,12,30)$) and four  regularization level ($\lambda \in \{0, 0.5, 1.0, 1.5\}$). 

\noindent \underline{\bf Results.} In the left of Figure~\ref{classicfig:regretlow} (resp. Figure~\ref{classicfig:regret}) is the plot of the regret of Algorithm~\ref{algucb} with $\bm{\gamma} = (10,8,20)$ (resp. $\bm{\gamma} = (15,12,30)$) and $\lambda \in\{ 0, 0.5,1.0, 1.5\}$ versus the square root of the total time periods $T$.  In the right of Figure~\ref{classicfig:regretlow} (resp. Figure~\ref{classicfig:regret}) we plot the relative regret of Algorithm~\ref{algucb} versus the total time periods $T$.

In the left of Figure~\ref{classicfig:fairnesslow} (resp. Figure~\ref{classicfig:fairness}) is the plot of the max-min fairness versus the total time periods $T$ with $\bm{\gamma} = (10,8,20)$ (resp. $\bm{\gamma} = (15,12,30)$) and $\lambda \in\{0, 0.5,1.0, 1.5\}$.
In the right of Figure~\ref{classicfig:fairnesslow} (resp. Figure~\ref{classicfig:fairness}) we plot the average reward versus the total time periods $T$.

\begin{figure}[!h]
\centering
\includegraphics[width =0.4\textwidth]{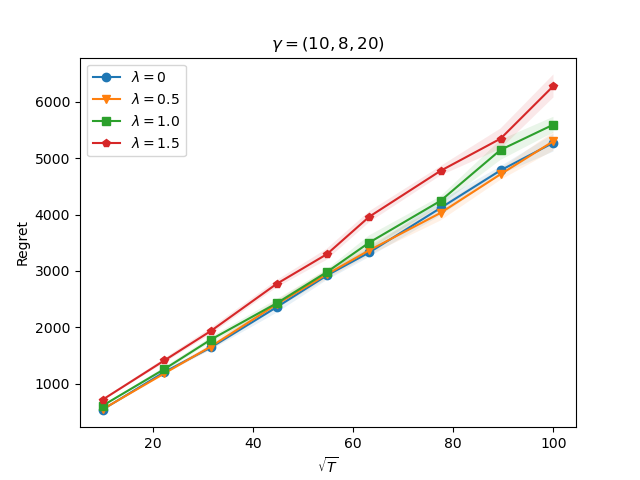}
\includegraphics[width =0.4\textwidth]{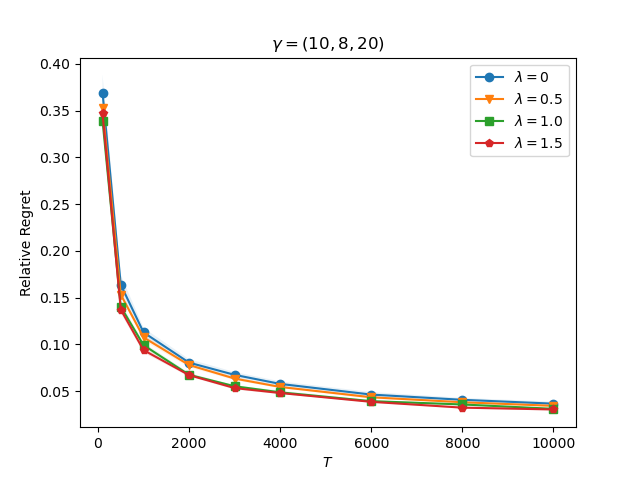}
\caption{ {The performance of Algorithm~\ref{algucb}+$\EG^\pm$ with $\bm{\gamma} = (15,12,30)$ and $\lambda \in\{ 0, 0.5,1.0, 1.5\}$ on the classic NRM example  \eqref{eq:linear demand}.  Here the $x$-axis of the left figure is the square root of the total time periods $T$ and the $y$-axis is the cumulative regret defined in Eq.~(\ref{regretform2}). The $x$-axis of the right figure is the total time periods $T$ and the $y$-axis is the relative regret defined in \eq{\ref{eq:relativeregret}}.}}
\label{classicfig:regretlow}
\end{figure}
\begin{figure}[!h]
\centering
\includegraphics[width =0.4\textwidth]{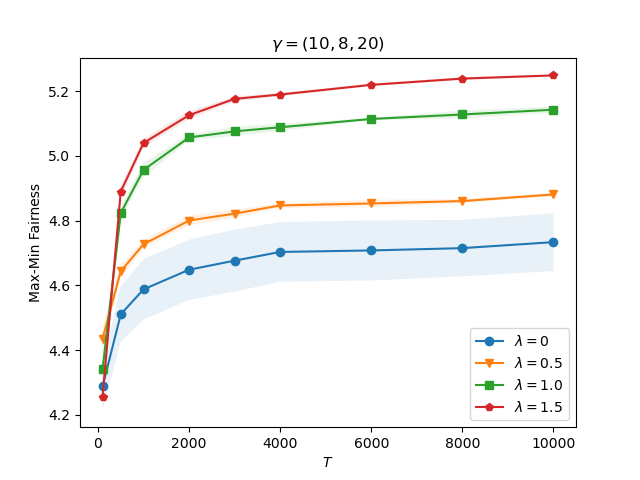}
\includegraphics[width =0.4\textwidth]{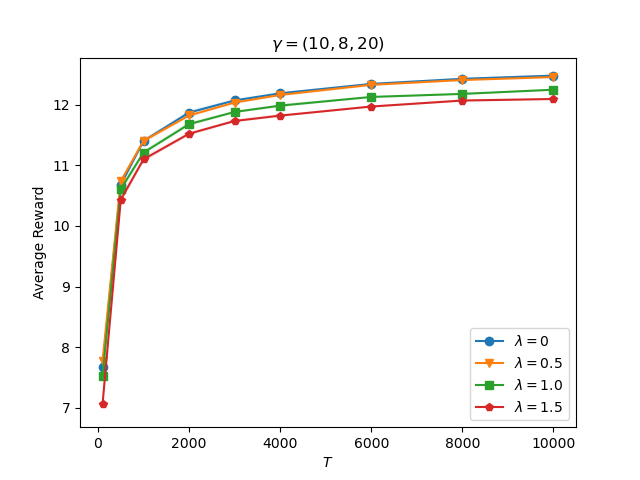}
\caption{ {The max-min fairness $\min_i\left(\frac{1}{T}\sum_{t=1}^T[\bm{A}\bm{d}_t]_i\right)$ and the average reward $\frac{1}{T}\sum_{t=1}^Tr(\bm{p}_t)$ of Algorithm~\ref{algucb}+$\EG^\pm$ at regularization levels $\lambda \in\{ 0, 0.5,1.0, 1.5\}$ under the initial inventory level $\bm{\gamma} =(10,8,20)$  on the classic NRM example \eqref{eq:linear demand}.  }}
\label{classicfig:fairnesslow}
\end{figure}
\begin{figure}[!h]
\centering
\includegraphics[width =0.4\textwidth]{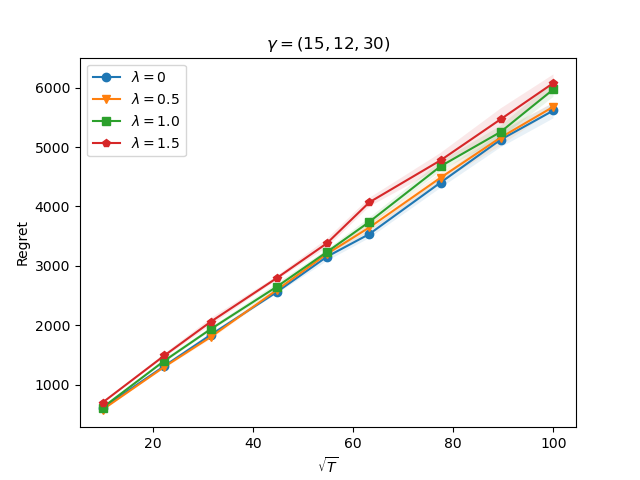}
\includegraphics[width =0.4\textwidth]{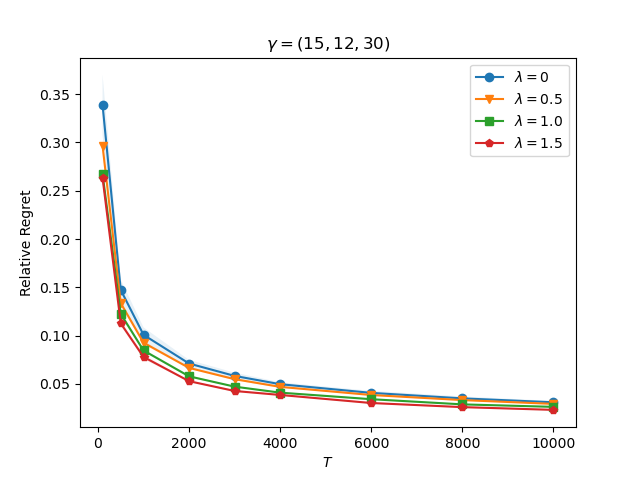}
\caption{ {The performance of Algorithm~\ref{algucb}+$\EG^\pm$ with $\bm{\gamma} = (15,12,30)$ and $\lambda \in\{ 0, 0.5,1.0, 1.5\}$ on the classic NRM example  \eqref{eq:linear demand}.  Here the $x$-axis of the left figure is the square root of the total time periods $T$ and the $y$-axis is the cumulative regret defined in Eq.~(\ref{regretform2}). The $x$-axis of the right figure is the total time periods $T$ and the $y$-axis is the relative regret defined in \eq{\ref{eq:relativeregret}}.}}
\label{classicfig:regret}
\end{figure}
\begin{figure}[!h]
\centering
\includegraphics[width =0.4\textwidth]{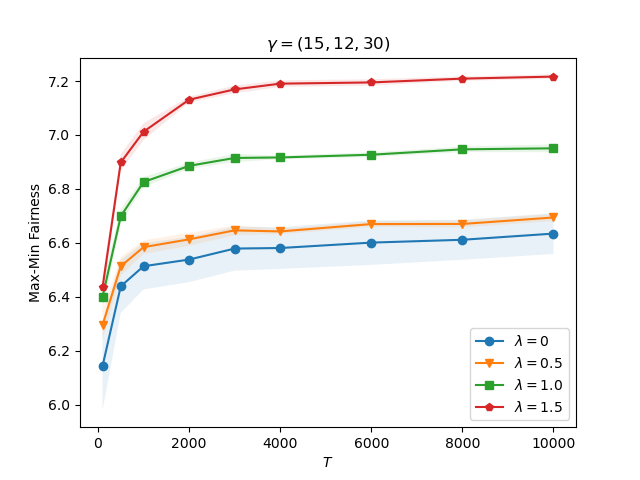}
\includegraphics[width =0.4\textwidth]{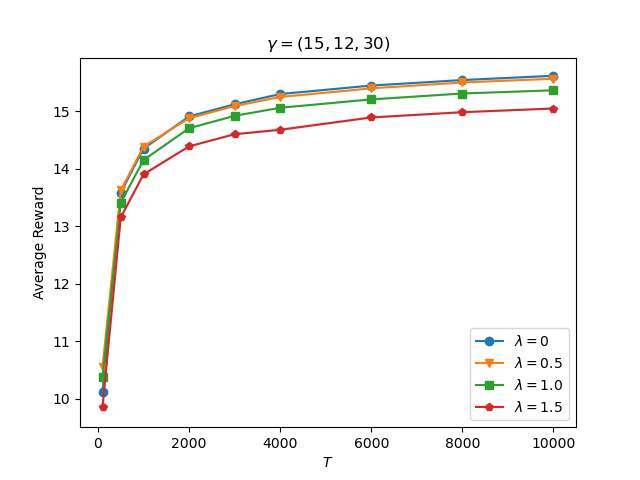}
\caption{ {The max-min fairness $\min_i\left(\frac{1}{T}\sum_{t=1}^T[\bm{A}\bm{d}_t]_i\right)$ and the average reward $\frac{1}{T}\sum_{t=1}^Tr(\bm{p}_t)$ of Algorithm~\ref{algucb}+$\EG^\pm$ at regularization levels $\lambda \in\{ 0, 0.5,1.0, 1.5\}$ under the initial inventory level $\bm{\gamma} =(15,12,30)$  on the classic NRM example \eqref{eq:linear demand}.}}
\label{classicfig:fairness}
\end{figure}




\end{document}